\documentclass{article} 
\usepackage{iclr2027_conference,times}

\usepackage{amsmath,amsfonts,bm}

\def\eqref#1{equation~\ref{#1}}

\def\1{\bm{1}}

\DeclareMathAlphabet{\mathsfit}{\encodingdefault}{\sfdefault}{m}{sl}
\SetMathAlphabet{\mathsfit}{bold}{\encodingdefault}{\sfdefault}{bx}{n}

\usepackage{hyperref}
\usepackage{url}

\usepackage{tikz}
\usepackage[most]{tcolorbox}
\usepackage{xcolor}
\usepackage{listings}
\usetikzlibrary{matrix, arrows.meta, calc}
\usepackage{booktabs}
\usepackage{graphicx}
\usepackage{subcaption}
\usepackage{tabularx}
\usepackage{multirow}
\usepackage{array}
\usepackage{cleveref}
\usepackage{arydshln}
\usepackage{wrapfig}

\definecolor{predicateblue}{HTML}{315F85}
\definecolor{actionbrown}{HTML}{965025}

\newcommand{\predconcept}[1]{\textcolor{predicateblue}{#1}}
\newcommand{\actconcept}[1]{\textcolor{actionbrown}{#1}}

\usepackage{etoolbox}

\newlength{\TitleAuthorAdjust}
\newlength{\AuthorRowStrut}
\newlength{\AfterAuthorsSkip}
\makeatletter

\patchcmd{\@maketitle}
  {\@title\par}
  {\@title\par
   \ificlrfinal
     \vskip\dimexpr
       24pt-\AuthorRowStrut+\TitleAuthorAdjust
     \relax
   \fi}
  {}{\PackageError{author-spacing}{Title patch failed}{}}

\newcommand{\PatchAuthorStrut}{%
  \patchcmd{\@maketitle}
    {\rule{\z@}{24pt}}
    {\rule{\z@}{\AuthorRowStrut}}
    {}
    {\PackageError{author-spacing}{Strut patch failed}{}}%
}

\PatchAuthorStrut
\PatchAuthorStrut
\PatchAuthorStrut

\patchcmd{\@maketitle}
  {\vskip 0.3in minus 0.1in}
  {\ificlrfinal
     \vskip\AfterAuthorsSkip
   \else
     \vskip 0.3in minus 0.1in
   \fi}
  {}
  {\PackageError{author-spacing}{Bottom patch failed}{}}

\makeatother

\title{ConceptARC at Scale: Studying Conceptual Understanding in Reasoning Models}

\author{Philipp Mondorf\thanks{%
  Correspondence to: {\protect\hypersetup{pdfborder={0 0 0}}%
   \href{mailto:p.mondorf@lmu.de}{\texttt{p.mondorf@lmu.de}}}.%
   \hspace{1mm}
   \textsuperscript{{\textdagger}}Equal supervision.
}\\
LMU Munich \& MCML
\And Adrien Deli\`ege \\
University of Li\`ege
\And Solim LeGris\\
New York University
\And Claas Beger\\
Santa Fe Institute
\And Ryan Yi\\
Santa Fe Institute
\And Barbara Plank\\
LMU Munich \& MCML
\And Brenden M. Lake\textsuperscript{\textdagger}\\
Princeton University
\And Todd M.~Gureckis\textsuperscript{\textdagger}\\
New York University
}

\iclrfinalcopy
\begin{document}

\maketitle

\begin{abstract}
Large reasoning models (LRMs) perform well on challenging benchmarks, yet it remains unclear whether they form generalizable abstractions across conceptually related reasoning tasks. The Abstraction and Reasoning Corpus (ARC) provides a controlled testbed for studying abstraction, requiring solvers to infer transformation rules over two-dimensional grids from a small set of examples. While ARC evaluates rule abstraction within individual tasks, different tasks may share higher-level structure despite requiring different transformations\textemdash{}for example, several tasks may involve reasoning about objects contained within other objects. Benchmarks such as ConceptARC capture this structure by grouping ARC-style tasks into recurring spatial and semantic concepts, but they often contain only a few manually constructed tasks per concept. We introduce \textsc{ConceptARC at Scale}, a generator for creating thousands of ARC-style tasks targeting a specific concept. Using this generator, we construct \textsc{ConceptARC-100}, a dataset of 1{,}600 new, manually reviewed tasks, with 100 tasks for each of the 16 concepts defined by ConceptARC\textemdash{}ten times the size of the original benchmark. We evaluate a range of recent LRMs and models trained from scratch on \textsc{ConceptARC-100} to characterize which concepts they solve reliably and where they struggle. We further analyze the models' internal representations and find clear semantic structure: task representations cluster according to the operations applied to objects, such as recoloring, removal, or translation, even when realized through substantially different rules and scenarios. Together, our results suggest that models form stable abstractions of transformation operations that generalize across diverse tasks.
\end{abstract}

\begin{figure}[h!]
\centering
\resizebox{0.965\textwidth}{!}{%
    \input{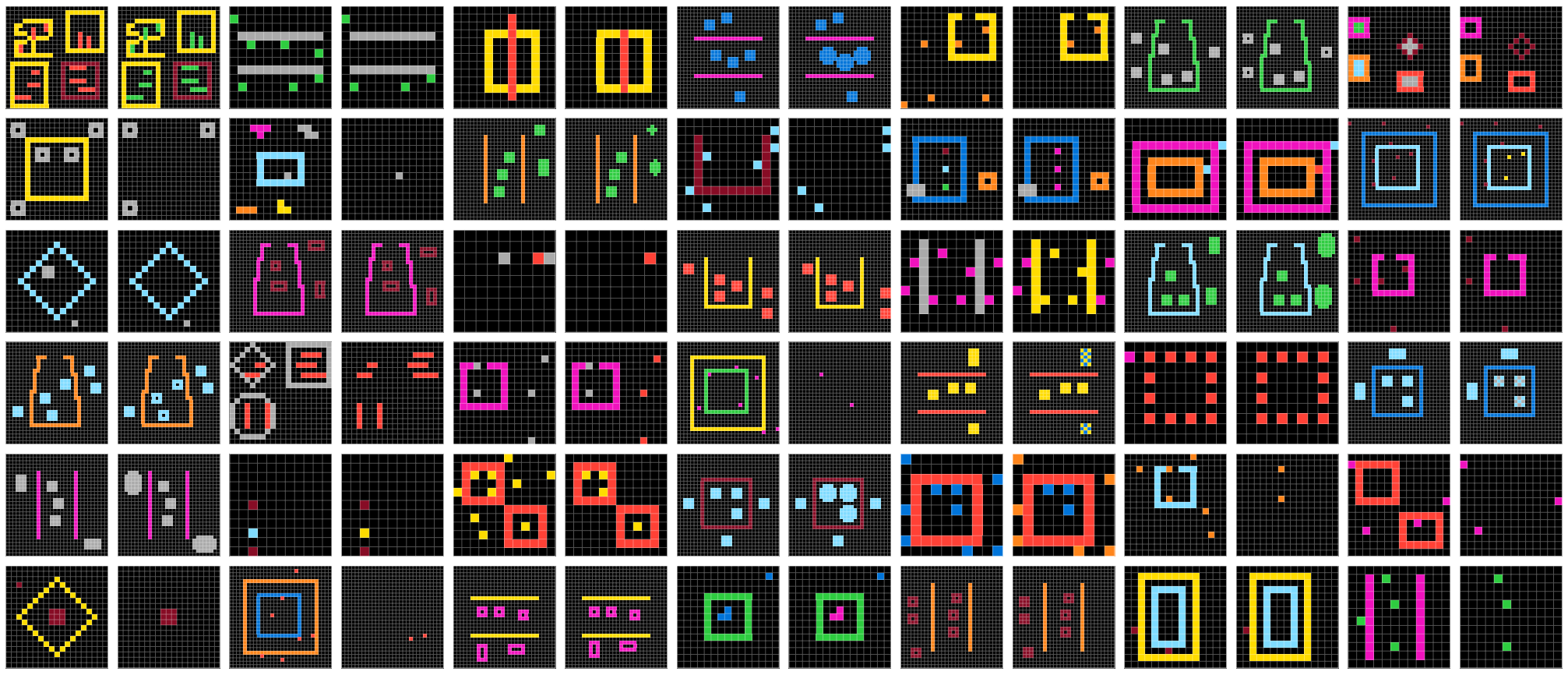}
}
\caption{\textbf{Examples from 42 different tasks generated for the ``\emph{Inside and Outside}'' concept.} Each example consists of an input grid and its corresponding output grid shown to its right, illustrating the task-specific transformation. To illustrate task diversity, we show only \emph{one} example from each of the 42 tasks. In \textsc{ConceptARC-100}, each task contains multiple examples (see Figure~\ref{fig:inside_outside_example}).}
\label{fig:opener_inside_outside_examples}
\end{figure}

\section{Introduction}\label{sec:introduction}
The ability to form abstractions and reason with higher-level concepts is fundamental to human problem-solving~\citep{gentner_1983, gentner_2017}. However, the extent to which large reasoning models rely on generalizable abstractions when solving complex reasoning problems remains unclear~\citep{beger2025ai, opielka2025analogical}. Answering this question is crucial for understanding how these models represent complex tasks and whether their representations support generalization to novel scenarios~\citep{hendel-etal-2023-context, ICLR2024_4ae163cb}.

The Abstraction and Reasoning Corpus (ARC)~\citep{chollet2019measure} provides a useful domain for studying this question. Each task presents a small number of input--output grid examples from which a solver must infer an underlying transformation rule and apply it to unseen inputs (see Figure~\ref{fig:inside_outside_example} in the Appendix for an example). ARC studies abstraction within a task: the solver must identify the structure shared across a few concrete examples and infer a rule that generalizes beyond them. Abstraction, however, can also extend across tasks. Different ARC tasks may share higher-level structure even when they require different transformation rules~\citep{NEURIPS2022_182aed03, hodel2024addressing}. For example, one task may require recoloring line objects inside a container, as shown in the upper-left input--output example in Figure~\ref{fig:opener_inside_outside_examples}, whereas another task may require removing colored pixels between two line objects, as shown in the example to its right. Although the transformations differ, both rely on the concept of spatial containment. This raises a broader question: do models abstract such shared concepts across tasks with different transformation rules? ConceptARC~\citep{moskvichev2023the} provides a basis for this type of analysis by organizing ARC-style tasks according to recurring spatial and semantic concepts. However, it contains only ten hand-crafted tasks for each of 16 concepts, with relatively few examples per task. Moreover, because its tasks have been public since 2023, training-data contamination cannot be ruled out for recent models.

To address these limitations, we introduce \textsc{ConceptARC at Scale}, a generator for automatically creating thousands of new ARC-style tasks organized around the concepts defined by ConceptARC. Each generated task is represented by a program that maps an input grid to an output grid, enabling the generation of many input--output grid examples per task. Figure~\ref{fig:opener_inside_outside_examples} shows one example from each of several generated tasks for the concept ``\emph{Inside and Outside}.'' Although their transformation rules differ, all require reasoning about whether elements lie \emph{inside or outside} a region.

We use our task generator to create \textsc{ConceptARC-100}, a dataset of 100 new, manually reviewed tasks for each of the 16 concepts defined by ConceptARC, totaling 1{,}600 tasks\textemdash{}ten times the size of the original benchmark. We evaluate several open-weight LRMs, as well as models trained from scratch on our generated data, and find comparatively strong performance on concepts such as ``\emph{Center}'', ``\emph{Above and Below}'', and ``\emph{Top and Bottom (2D)}'', whose transformations depend largely on spatial relations within the grid. In contrast, models perform worse on concepts such as ``\emph{Same / Different}'', ``\emph{Order}'', and ``\emph{Count}'', which require recognizing and comparing objects or patterns based on properties such as shape, numerosity, or color. Under a stricter task-level criterion that requires all 10 test inputs to be solved correctly, even the best-performing model succeeds on fewer than 24\% of tasks, indicating that models struggle to solve tasks consistently across multiple test inputs.

Behavioral performance alone, however, does not reveal which abstractions models have formed. We therefore analyze their internal representations of tasks and find highly structured clusters organized primarily around the operations applied to objects\textemdash{}such as recoloring, removal, or translation\textemdash{}rather than by ConceptARC's higher-level spatial and semantic concepts such as ``\emph{Inside and Outside}'' (Figure~\ref{fig:opener_inside_outside_examples}). For example, tasks involving object translations often cluster together even when they belong to different concepts and follow substantially different rules. These results suggest that models represent structured regularities across ARC-style tasks, but that this structure reflects similarities in the required transformation operations more than the higher-level concepts motivating the benchmark.

In summary, our contributions are as follows:

\begin{enumerate}
\item We introduce \textsc{ConceptARC at Scale}, a generator for automatically creating thousands of new ARC-style tasks organized around the 16 concepts defined by ConceptARC. Each task is represented by a program that supports the generation of many grid examples.
\item We publicly release \textsc{ConceptARC-100}, a dataset of 100 new, manually reviewed tasks for each concept\textemdash{}ten times the size of the original ConceptARC benchmark.
\item We evaluate a range of recent LRMs, as well as models trained from scratch on \textsc{ConceptARC-100}, characterizing performance across concepts and showing that even the best-performing model struggles to consistently solve all test inputs within a task.
\item We analyze the models' internal task representations and find clear semantic structure: tasks cluster according to the transformation operations applied to objects, even when those operations arise from substantially different rules, scenarios, and concepts.
\end{enumerate}

\section{Related work}\label{sec:related_work}

\paragraph{ARC and beyond.}
The Abstraction and Reasoning Corpus~\citep{chollet2019measure} was introduced to evaluate the ability to infer abstract transformation rules from a small number of examples and generalize them to novel inputs. Subsequent work has extended ARC or introduced related datasets, either handcrafted or procedurally generated. LARC~\citep{acquaviva2021larc} provides natural-language descriptions of ARC tasks, DARC~\citep{huang2023ANPL} records human interactions with a programmatic task-decomposition system, and H-ARC~\citep{LeGris2025} collects solution attempts from about 1,700 humans. While these datasets provide valuable insights into how humans approach ARC tasks, they do not introduce new tasks nor support large-scale task generation.

Handcrafted ARC-like benchmarks include ARC-AGI-2~\citep{chollet2026arcagi2newchallengefrontier}, designed to be substantially more challenging than its predecessor; Mini-ARC~\citep{Kim2022Playgrounds}, which contains 150 new tasks on smaller $5\times5$ grids and broad transformation labels; and 1D-ARC~\citep{xu2024llms}, with 50 variants of 18 rules on one-dimensional grids. AugARC~\citep{bikov2025augarc} instead augments existing tasks without introducing new ones. These benchmarks remain limited in scale.

More closely related to our work are ARC-like task generators. RE-ARC~\citep{hodel2024addressing} uses a custom domain-specific language (DSL) to generate input--output pairs for all 400 original ARC training tasks, although the generated instances may extend substantially beyond the diversity of the original tasks. ARC-GEN~\citep{Moffitt2025} follows a similar approach but more closely preserves the original task distributions. BARC~\citep{li2025combining} uses LLMs to synthesize tasks from descriptions of approximately 40\% of the original training tasks, although some generated tasks appear erroneous~\citep{li2025combining, Moffitt2025}. ARC-TGI~\citep{Lehmann2026ARCTGI} provides human-validated generators covering half of the original training tasks. Other benchmarks, including Compositional-ARC~\citep{mondorf2026compositionalarc} and Sort-of-ARC~\citep{assouel2022objectcentric}, target compositional reasoning.

A complementary line of work approaches ARC as a program synthesis problem, aiming to infer programs for existing ARC tasks that map input grids to their corresponding outputs~\citep{10.1007/978-3-030-93409-5_54, legris2026synthesizing}. Particularly relevant to our work is ARGA~\citep{Xu_Khalil_Sanner_2023}, which represents task programs as compositions of \emph{filters} that select objects and \emph{transformations} that modify them. ARGA abstracts grids into object-centric graphs and uses constraint-guided search to identify suitable programs. HySynth~\citep{NEURIPS2024_1c9c85ba} builds on ARGA's DSL, using LLM-generated program proposals to construct a probabilistic grammar for bottom-up search. NSA~\citep{batorski2025nsa} extends ARGA's transformation vocabulary and trains a transformer to guide symbolic search. Similar to our task generator (Section~\ref{sec:data_generator}), these approaches express ARC transformation rules through compositions of object-selection and transformation functions. However, while they primarily aim to infer programs that solve existing tasks, we compose layouts, selectors, and actions to generate new ARC-style tasks, using concept-specific constraints to ensure that each task program instantiates a targeted spatial or semantic concept.

\paragraph{ConceptARC.}
ConceptARC~\citep{moskvichev2023the} evaluates conceptual understanding and abstraction in the ARC domain. Each task provides two to four input--output grid examples from which a transformation must be inferred and applied to three test inputs. Unlike ARC, ConceptARC organizes its tasks around 16 \emph{concepts} capturing recurring spatial and semantic structures, with ten manually designed tasks per concept.

For our analysis, we broadly group these concepts into eight \emph{predicate} and eight \emph{action} concepts. Predicate concepts describe spatial or relational properties that determine which objects, regions, or positions are relevant to a transformation. Examples include ``\emph{Above and Below},'' ``\emph{Horizontal and Vertical},'' and ``\emph{Inside and Outside}''. Action concepts describe operations applied to objects or grid regions. Examples include ``\emph{Clean Up},'' ``\emph{Complete Shape},'' and ``\emph{Count}.'' Table~\ref{tab:conceptarc_concepts} in the Appendix provides detailed descriptions of all concepts and their classification as predicate or action concepts.

In the original study, \citet{moskvichev2023the} compare human participants with three LLM-based solvers and analyze performance across concepts. Subsequent work, such as \citet{beger2025ai}, studies more recent reasoning and multimodal models, analyzing their natural-language descriptions of inferred rules to assess whether successful solutions reflect the targeted concepts or surface-level shortcuts. To the best of our knowledge, however, no prior work has examined whether recent LRMs internally organize ARC tasks according to ConceptARC's spatial and semantic abstractions.

\section{ConceptARC at Scale}\label{sec:data_generator}

\definecolor{carcARCzero}{HTML}{000000}
\definecolor{carcARCone}{HTML}{0074D9}
\definecolor{carcARCtwo}{HTML}{FF4136}
\definecolor{carcARCthree}{HTML}{2ECC40}
\definecolor{carcARCfour}{HTML}{FFDC00}
\definecolor{carcARCfive}{HTML}{AAAAAA}
\definecolor{carcARCsix}{HTML}{F012BE}
\definecolor{carcARCseven}{HTML}{FF851B}
\definecolor{carcARCeight}{HTML}{7FDBFF}
\definecolor{carcARCnine}{HTML}{870C25}

\definecolor{carcCard}{HTML}{F8F9FA}
\definecolor{carcCardBorder}{HTML}{D1D5DB}
\definecolor{carcInk}{HTML}{374151}
\definecolor{carcMuted}{HTML}{9CA3AF}

\definecolor{carcCodeBg}{HTML}{EEF1F5}
\definecolor{carcCodeBorder}{HTML}{D8DDE5}

\definecolor{carcSelectedBlue}{HTML}{D9EEF8}
\definecolor{carcSelectedBlueStrong}{HTML}{5DA9D6}

\definecolor{carcGateCard}{HTML}{F3F4F6}
\definecolor{carcGateBorder}{HTML}{D1D5DB}
\definecolor{carcGateInk}{HTML}{31566B}
\definecolor{carcGateRule}{HTML}{FFF4B8}
\definecolor{carcGateAllowed}{HTML}{D8F0DF}
\definecolor{carcGateAllowedStrong}{HTML}{4F9D69}
\definecolor{carcGateDisallowed}{HTML}{F6DADA}
\definecolor{carcGateDisallowedStrong}{HTML}{C65D5D}

\definecolor{carcConceptInsideOutside}{HTML}{5DA9D6}
\definecolor{carcConceptAboveBelow}{HTML}{E3A34F}

\def\carcUnit{0.27cm}
\def\carcLayoutN{15}

\def\carcDepthX{0.78}
\def\carcDepthY{0.68}

\def\carcTitleGap{0.78}

\newcommand{\carcMainTitleFont}{%
  \bfseries\Huge
}
\newcommand{\carcConceptTitleFont}{%
  \bfseries\Huge
}
\newcommand{\carcCodeFont}{%
  \ttfamily\Large
}
\newcommand{\carcGateTextFont}{%
  \bfseries\Large
}
\newcommand{\carcIOFont}{%
  \bfseries\LARGE
}
\newcommand{\carcDotsFont}{%
  \bfseries\Huge
}

\newcommand{\carcLayoutLayer}[3]{%
  \begin{scope}[shift={(#1,#2)}]

    \fill[carcARCzero]
      (0,0) rectangle (\carcLayoutN,\carcLayoutN);

    #3

    \draw[
      step=1,
      draw=white,
      opacity=0.32,
      line width=0.06pt
    ]
      (0,0) grid (\carcLayoutN,\carcLayoutN);

    \draw[
      draw=white,
      opacity=0.55,
      line width=0.18pt
    ]
      (0,0) rectangle (\carcLayoutN,\carcLayoutN);

  \end{scope}%
}

\newcommand{\carcYellowContainer}{%
  \fill[carcARCfour] (4,4) rectangle (11,5);
  \fill[carcARCfour] (4,10) rectangle (11,11);
  \fill[carcARCfour] (4,5) rectangle (5,10);
  \fill[carcARCfour] (10,5) rectangle (11,10);
}

\newcommand{\carcInsideMarkers}[1]{%
  \fill[#1] (5,6) rectangle (6,7);
  \fill[#1] (7,6) rectangle (8,7);
  \fill[#1] (6,8) rectangle (7,9);
  \fill[#1] (8,9) rectangle (9,10);
  \fill[#1] (9,7) rectangle (10,8);
}

\newcommand{\carcAboveMarkers}[1]{%
  \fill[#1] (0,12) rectangle (1,13);
  \fill[#1] (14,11) rectangle (15,12);
  \fill[#1] (5,12) rectangle (6,13);
  \fill[#1] (9,13) rectangle (10,14);
  \fill[#1] (11,14) rectangle (12,15);
}

\newcommand{\carcOtherOutsideMarkers}[1]{%
  \fill[#1] (1,9) rectangle (2,10);
  \fill[#1] (2,4) rectangle (3,5);
  \fill[#1] (12,8) rectangle (13,9);
  \fill[#1] (13,3) rectangle (14,4);
  \fill[#1] (4,1) rectangle (5,2);
  \fill[#1] (8,2) rectangle (9,3);
  \fill[#1] (10,0) rectangle (11,1);
}

\newcommand{\carcInputLayout}{%
  \carcYellowContainer
  \carcInsideMarkers{carcARCtwo}
  \carcAboveMarkers{carcARCtwo}
  \carcOtherOutsideMarkers{carcARCtwo}
}

\newcommand{\carcInsideOutsideOutput}{%
  \carcYellowContainer
  \carcInsideMarkers{carcARCthree}
  \carcAboveMarkers{carcARCtwo}
  \carcOtherOutsideMarkers{carcARCtwo}
}

\newcommand{\carcAboveBelowOutput}{%
  \carcYellowContainer
  \carcInsideMarkers{carcARCtwo}
  \carcAboveMarkers{carcARCthree}
  \carcOtherOutsideMarkers{carcARCtwo}
}

\def\carcPrimitiveW{57.5}
\def\carcPrimitiveH{16.6}

\def\carcPrimitiveCodeX{0.85}
\def\carcPrimitiveCodeY{1.15}
\def\carcPrimitiveCodeW{41.5}
\def\carcPrimitiveCodeH{14.50}

\newcommand{\carcPrimitiveCard}[3]{%
  \begin{scope}[shift={(#1,#2)}]

    \fill[
      carcCard,
      rounded corners=6pt
    ]
      (0,0) rectangle
      (\carcPrimitiveW,\carcPrimitiveH);

    \draw[
      carcCardBorder,
      rounded corners=6pt,
      line width=0.95pt
    ]
      (0,0) rectangle
      (\carcPrimitiveW,\carcPrimitiveH);

    #3

  \end{scope}%
}

\newcommand{\carcPrimitiveCodeBox}[1]{%

  \fill[
    carcCodeBg,
    rounded corners=5pt
  ]
    (\carcPrimitiveCodeX,\carcPrimitiveCodeY)
    rectangle
    ({\carcPrimitiveCodeX+\carcPrimitiveCodeW},
     {\carcPrimitiveCodeY+\carcPrimitiveCodeH});

  \draw[
    carcCodeBorder,
    rounded corners=5pt,
    line width=0.75pt
  ]
    (\carcPrimitiveCodeX,\carcPrimitiveCodeY)
    rectangle
    ({\carcPrimitiveCodeX+\carcPrimitiveCodeW},
     {\carcPrimitiveCodeY+\carcPrimitiveCodeH});

  \node[
    anchor=north west,
    align=left,
    font=\carcCodeFont,
    text=carcInk,
    text width=11.05cm,
    inner sep=4pt
  ] at (
    {\carcPrimitiveCodeX+0.65},
    {\carcPrimitiveCodeY+\carcPrimitiveCodeH-0.82}
  ) {#1};
}

\newcommand{\carcSelectorViz}{%

  \fill[carcSelectedBlue]
    (1.55,1.20) rectangle (9.85,5.65);

  \draw[
    carcARCfour,
    line width=1.8pt
  ]
    (1.55,1.20) rectangle (9.85,5.65);

  \foreach \x/\y in {
    2.45/4.55,
    3.40/2.25,
    4.55/3.70,
    5.85/2.45,
    7.70/4.60,
    7.70/3.25,
    3.80/4.85
  }{
    \fill[carcSelectedBlueStrong]
      (\x,\y) circle[radius=0.24];
  }

  \foreach \x/\y in {
    0.35/4.50,
    0.60/2.15,
    0.30/1.10,
    10.85/4.55,
    11.15/2.45,
    10.75/1.20
  }{
    \fill[carcMuted]
      (\x,\y) circle[radius=0.21];
  }
}

\newcommand{\carcActionViz}{%

  \fill[carcARCtwo]
    (0.65,2.05) rectangle (2.70,4.10);

  \draw[
    carcInk,
    opacity=0.25,
    line width=0.55pt
  ]
    (0.65,2.05) rectangle (2.70,4.10);

  \draw[
    -{Stealth[length=2.5mm,width=1.8mm]},
    line width=1.45pt,
    draw=carcInk
  ]
    (3.55,3.075)
    --
    (6.05,3.075);

  \fill[carcARCthree]
    (6.90,2.05) rectangle (8.95,4.10);

  \draw[
    carcInk,
    opacity=0.25,
    line width=0.55pt
  ]
    (6.90,2.05) rectangle (8.95,4.10);
}

\def\carcTaskW{72.0}
\def\carcTaskH{20.6}

\def\carcTaskCodeX{0.90}
\def\carcTaskCodeY{1.30}
\def\carcTaskCodeW{42.0}
\def\carcTaskCodeH{18.00}

\def\carcTaskGridScale{0.74}

\newcommand{\carcTaskCard}[3]{%
  \begin{scope}[shift={(#1,#2)}]

    \fill[
      carcCard,
      rounded corners=7pt
    ]
      (0,0) rectangle
      (\carcTaskW,\carcTaskH);

    \draw[
      carcCardBorder,
      rounded corners=7pt,
      line width=0.95pt
    ]
      (0,0) rectangle
      (\carcTaskW,\carcTaskH);

    #3

  \end{scope}%
}

\newcommand{\carcTaskCodeBox}[1]{%

  \fill[
    carcCodeBg,
    rounded corners=5pt
  ]
    (\carcTaskCodeX,\carcTaskCodeY)
    rectangle
    ({\carcTaskCodeX+\carcTaskCodeW},
     {\carcTaskCodeY+\carcTaskCodeH});

  \draw[
    carcCodeBorder,
    rounded corners=5pt,
    line width=0.75pt
  ]
    (\carcTaskCodeX,\carcTaskCodeY)
    rectangle
    ({\carcTaskCodeX+\carcTaskCodeW},
     {\carcTaskCodeY+\carcTaskCodeH});

  \node[
    anchor=north west,
    align=left,
    font=\carcCodeFont,
    text=carcInk,
    text width=11.45cm,
    inner sep=4pt
  ] at (
    {\carcTaskCodeX+0.72},
    {\carcTaskCodeY+\carcTaskCodeH-0.90}
  ) {#1};
}

\newcommand{\carcSchematicOverview}{%
\begin{tikzpicture}[
  x=\carcUnit,
  y=\carcUnit,
  line cap=round,
  line join=round,
  carcFlow/.style={
    -{Stealth[length=3.0mm,width=2.2mm]},
    line width=1.35pt
  },
  carcStrongFlow/.style={
    -{Stealth[length=3.3mm,width=2.4mm]},
    line width=1.55pt
  },
  carcInnerFlow/.style={
    -{Stealth[length=2.8mm,width=2.0mm]},
    line width=1.55pt
  }
]


\def\carcCenterLineY{0.0}

\def\carcOuterRowOffset{23.0}

\def\carcTaskRowOffset{21.8}

\def\carcPrimitiveX{-10.0}
\def\carcLayoutX{\carcPrimitiveX}
\def\carcActionX{\carcPrimitiveX}
\def\carcSelectorX{\carcPrimitiveX}

\def\carcGateW{34.0}

\def\carcTaskX{111.4}

\pgfmathsetmacro{\carcPrimitiveStackW}{
  \carcPrimitiveW + 2*\carcDepthX
}

\pgfmathsetmacro{\carcPrimitiveStackH}{
  \carcPrimitiveH + 2*\carcDepthY
}

\pgfmathsetmacro{\carcTaskStackW}{
  \carcTaskW + 2*\carcDepthX
}

\pgfmathsetmacro{\carcTaskStackH}{
  \carcTaskH + 2*\carcDepthY
}

\pgfmathsetmacro{\carcGateH}{
  \carcPrimitiveStackH
}

\pgfmathsetmacro{\carcLayoutY}{
  \carcCenterLineY
  + \carcOuterRowOffset
  - 0.5*\carcPrimitiveStackH
}

\pgfmathsetmacro{\carcSelectorY}{
  \carcCenterLineY
  - 0.5*\carcPrimitiveStackH
}

\pgfmathsetmacro{\carcActionY}{
  \carcCenterLineY
  - \carcOuterRowOffset
  - 0.5*\carcPrimitiveStackH
}

\pgfmathsetmacro{\carcTaskTopY}{
  \carcCenterLineY
  + \carcTaskRowOffset
  - 0.5*\carcTaskStackH
}

\pgfmathsetmacro{\carcTaskBottomY}{
  \carcCenterLineY
  - \carcTaskRowOffset
  - 0.5*\carcTaskStackH
}

\pgfmathsetmacro{\carcPrimitiveStackRight}{
  \carcPrimitiveX + \carcPrimitiveStackW
}

\pgfmathsetmacro{\carcGateCenterTargetX}{
  0.5*(\carcPrimitiveStackRight + \carcTaskX)
}

\pgfmathsetmacro{\carcGateX}{
  \carcGateCenterTargetX - 0.5*\carcGateW
}

\pgfmathsetmacro{\carcGateY}{
  \carcCenterLineY - 0.5*\carcGateH
}

\pgfmathsetmacro{\carcGateRight}{
  \carcGateX + \carcGateW
}

\pgfmathsetmacro{\carcGateTop}{
  \carcGateY + \carcGateH
}

\pgfmathsetmacro{\carcGateCenterX}{
  \carcGateX + 0.5*\carcGateW
}

\pgfmathsetmacro{\carcGateCenterY}{
  \carcGateY + 0.5*\carcGateH
}

\pgfmathsetmacro{\carcLayoutCenterX}{
  \carcLayoutX + 0.5*\carcPrimitiveStackW
}

\pgfmathsetmacro{\carcLayoutTop}{
  \carcLayoutY + \carcPrimitiveStackH
}

\pgfmathsetmacro{\carcLayoutCenterY}{
  \carcLayoutY + 0.5*\carcPrimitiveStackH
}

\pgfmathsetmacro{\carcLayoutRight}{
  \carcLayoutX + \carcPrimitiveStackW
}

\node[
  font=\carcMainTitleFont,
  anchor=south
] at (
  \carcLayoutCenterX,
  \carcLayoutTop+\carcTitleGap
)
  {Layouts};

\carcPrimitiveCard
  {\carcLayoutX+2*\carcDepthX}
  {\carcLayoutY+2*\carcDepthY}
  {}

\carcPrimitiveCard
  {\carcLayoutX+\carcDepthX}
  {\carcLayoutY+\carcDepthY}
  {}

\carcPrimitiveCard
  {\carcLayoutX}
  {\carcLayoutY}
  {%

    \carcPrimitiveCodeBox{%
      \textbf{\detokenize{def rectangle_with_markers(...):}}\\[2pt]
      \quad\detokenize{grid = empty_grid(...)}\\
      \quad\detokenize{draw_rectangle(grid, ...)}\\
      \quad\detokenize{place_markers(grid, ...)}\\
      \quad\detokenize{return grid}
    }

    \begin{scope}[
      shift={(44.85,2.55)},
      scale=0.70
    ]
      \carcLayoutLayer
        {0}
        {0}
        {\carcInputLayout}
    \end{scope}
  }

\pgfmathsetmacro{\carcSelectorCenterX}{
  \carcSelectorX + 0.5*\carcPrimitiveStackW
}

\pgfmathsetmacro{\carcSelectorTop}{
  \carcSelectorY + \carcPrimitiveStackH
}

\pgfmathsetmacro{\carcSelectorCenterY}{
  \carcSelectorY + 0.5*\carcPrimitiveStackH
}

\pgfmathsetmacro{\carcSelectorRight}{
  \carcSelectorX + \carcPrimitiveStackW
}

\node[
  font=\carcMainTitleFont,
  anchor=south
] at (
  \carcSelectorCenterX,
  \carcSelectorTop+\carcTitleGap
)
  {Selectors};

\carcPrimitiveCard
  {\carcSelectorX+2*\carcDepthX}
  {\carcSelectorY+2*\carcDepthY}
  {}

\carcPrimitiveCard
  {\carcSelectorX+\carcDepthX}
  {\carcSelectorY+\carcDepthY}
  {}

\carcPrimitiveCard
  {\carcSelectorX}
  {\carcSelectorY}
  {%

    \carcPrimitiveCodeBox{%
      \textbf{\detokenize{def select_objects_inside(...):}}\\[2pt]
      \quad\detokenize{container = find_container(grid)}\\
      \quad\detokenize{objects = parse_objects(container)}\\
      \quad\detokenize{return objects}
    }

    \begin{scope}[
      shift={(45.15,5.36)},
      scale=0.99
    ]
      \carcSelectorViz
    \end{scope}
  }

\pgfmathsetmacro{\carcActionCenterX}{
  \carcActionX + 0.5*\carcPrimitiveStackW
}

\pgfmathsetmacro{\carcActionTop}{
  \carcActionY + \carcPrimitiveStackH
}

\pgfmathsetmacro{\carcActionCenterY}{
  \carcActionY + 0.5*\carcPrimitiveStackH
}

\pgfmathsetmacro{\carcActionRight}{
  \carcActionX + \carcPrimitiveStackW
}

\node[
  font=\carcMainTitleFont,
  anchor=south
] at (
  \carcActionCenterX,
  \carcActionTop+\carcTitleGap
)
  {Actions};

\carcPrimitiveCard
  {\carcActionX+2*\carcDepthX}
  {\carcActionY+2*\carcDepthY}
  {}

\carcPrimitiveCard
  {\carcActionX+\carcDepthX}
  {\carcActionY+\carcDepthY}
  {}

\carcPrimitiveCard
  {\carcActionX}
  {\carcActionY}
  {%

    \carcPrimitiveCodeBox{%
      \textbf{\detokenize{def recolor_objects_green(...):}}\\[2pt]
      \quad\detokenize{return recolor(obj, color=GREEN)}
    }

    \begin{scope}[
      shift={(44.25,4.33)},
      scale=1.18
    ]
      \carcActionViz
    \end{scope}
  }

\node[
  font=\carcConceptTitleFont,
  anchor=south
] at (
  \carcGateCenterX,
  \carcGateTop+\carcTitleGap
)
  {Concept Gate};

\pgfmathsetmacro{\carcGateRouteX}{
  \carcGateX - 3.2
}

\pgfmathsetmacro{\carcGateLayoutInputY}{
  \carcGateY + 0.78*\carcGateH
}

\pgfmathsetmacro{\carcGateActionInputY}{
  \carcGateY + 0.22*\carcGateH
}

\draw[
  carcFlow,
  draw=carcGateInk,
  rounded corners=4pt
]
  (\carcLayoutRight,\carcLayoutCenterY)
  --
  (\carcGateRouteX,\carcLayoutCenterY)
  --
  (\carcGateRouteX,\carcGateLayoutInputY)
  --
  (\carcGateX,\carcGateLayoutInputY);

\draw[
  carcFlow,
  draw=carcGateInk
]
  (\carcSelectorRight,\carcSelectorCenterY)
  --
  (\carcGateX,\carcSelectorCenterY);

\draw[
  carcFlow,
  draw=carcGateInk,
  rounded corners=4pt
]
  (\carcActionRight,\carcActionCenterY)
  --
  (\carcGateRouteX,\carcActionCenterY)
  --
  (\carcGateRouteX,\carcGateActionInputY)
  --
  (\carcGateX,\carcGateActionInputY);

\fill[
  carcGateCard,
  rounded corners=8pt
]
  (\carcGateX,\carcGateY)
  rectangle
  (\carcGateRight,\carcGateTop);

\draw[
  carcGateBorder,
  rounded corners=8pt,
  line width=1.25pt
]
  (\carcGateX,\carcGateY)
  rectangle
  (\carcGateRight,\carcGateTop);

\begin{scope}[shift={(\carcGateX,\carcGateY)}]

  \fill[
    carcGateRule,
    rounded corners=4pt
  ]
    (5.0,12.3) rectangle (29.0,15.5);

  \draw[
    carcARCfour,
    rounded corners=4pt,
    line width=1.0pt
  ]
    (5.0,12.3) rectangle (29.0,15.5);

  \node[
    font=\carcGateTextFont,
    text=carcGateInk,
    align=center,
    text width=5.2cm
  ] at (17.0,13.9)
    {Concept Rule};

  \fill[
    carcGateAllowed,
    rounded corners=5pt
  ]
    (1.2,2.2) rectangle (16.2,9.0);

  \draw[
    carcGateAllowedStrong,
    rounded corners=5pt,
    line width=1.0pt
  ]
    (1.2,2.2) rectangle (16.2,9.);

  \fill[white]
    (3.3,5.6) circle[radius=0.70];

  \draw[
    carcGateAllowedStrong,
    line width=1.0pt
  ]
    (3.3,5.6) circle[radius=0.70];

  \draw[
    carcGateAllowedStrong,
    line width=1.9pt
  ]
    (2.92,5.56)
    --
    (3.20,5.27)
    --
    (3.69,5.96);

  \node[
    font=\carcGateTextFont,
    text=carcGateInk,
    align=center,
    text width=3.15cm
  ] at (10.3,5.6)
    {Allowed\\combination};

  \fill[
    carcGateDisallowed,
    rounded corners=5pt
  ]
    (18.0,2.2) rectangle (32.9,9.0);

  \draw[
    carcGateDisallowedStrong,
    rounded corners=5pt,
    line width=1.0pt
  ]
    (18.0,2.2) rectangle (32.9,9.0);

  \fill[white]
    (20.0,5.6) circle[radius=0.70];

  \draw[
    carcGateDisallowedStrong,
    line width=1.0pt
  ]
    (20.0,5.6) circle[radius=0.70];

  \draw[
    carcGateDisallowedStrong,
    line width=1.8pt
  ]
    (19.63,5.23)
    --
    (20.37,5.97);

  \draw[
    carcGateDisallowedStrong,
    line width=1.8pt
  ]
    (20.37,5.23)
    --
    (19.63,5.97);

  \node[
    font=\carcGateTextFont,
    text=carcGateInk,
    align=center,
    text width=3.15cm
  ] at (27.0,5.6)
    {Disallowed\\combination};

\end{scope}

\pgfmathsetmacro{\carcTopTaskStackTop}{
  \carcTaskTopY + \carcTaskStackH
}

\pgfmathsetmacro{\carcTopTaskCenterX}{
  \carcTaskX + 0.5*\carcTaskStackW
}

\pgfmathsetmacro{\carcTopTaskCenterY}{
  \carcTaskTopY + 0.5*\carcTaskStackH
}

\node[
  font=\carcConceptTitleFont,
  anchor=south
] at (
  \carcTopTaskCenterX,
  \carcTopTaskStackTop+\carcTitleGap
)
  {Inside \& Outside};

\carcTaskCard
  {\carcTaskX+2*\carcDepthX}
  {\carcTaskTopY+2*\carcDepthY}
  {}

\carcTaskCard
  {\carcTaskX+\carcDepthX}
  {\carcTaskTopY+\carcDepthY}
  {}

\carcTaskCard
  {\carcTaskX}
  {\carcTaskTopY}
  {%

    \carcTaskCodeBox{%
      \textbf{\detokenize{def generate_inside_outside_0001():}}\\[2pt]
      \quad\detokenize{x = rectangle_with_markers(...)}\\
      \quad\detokenize{s = select_objects_inside(x)}\\
      \quad\detokenize{y = x.map(s, recolor_object_green)}\\
      \quad\detokenize{return Task(input=x, output=y)}
    }

    \node[
      font=\carcIOFont,
      text=carcInk
    ] at (50.55,16.65)
      {input};

    \begin{scope}[
      shift={(45.00,4.15)},
      scale=\carcTaskGridScale
    ]
      \carcLayoutLayer
        {0}
        {0}
        {\carcInputLayout}
    \end{scope}

    \draw[
      carcInnerFlow,
      draw=carcInk
    ]
      (56.45,9.70)
      --
      (57.85,9.70);

    \node[
      font=\carcIOFont,
      text=carcInk
    ] at (63.85,16.65)
      {output};

    \begin{scope}[
      shift={(58.30,4.15)},
      scale=\carcTaskGridScale
    ]
      \carcLayoutLayer
        {0}
        {0}
        {\carcInsideOutsideOutput}
    \end{scope}
  }

\pgfmathsetmacro{\carcBottomTaskStackTop}{
  \carcTaskBottomY + \carcTaskStackH
}

\pgfmathsetmacro{\carcBottomTaskCenterX}{
  \carcTaskX + 0.5*\carcTaskStackW
}

\pgfmathsetmacro{\carcBottomTaskCenterY}{
  \carcTaskBottomY + 0.5*\carcTaskStackH
}

\pgfmathsetmacro{\carcMoreConceptsY}{
  0.5*(\carcBottomTaskStackTop + \carcTaskTopY)
}

\node[
  font=\carcDotsFont,
  text=black,
  inner sep=0pt
] at (
  \carcTopTaskCenterX,
  \carcMoreConceptsY
)
  {$\cdots$};

\node[
  font=\carcConceptTitleFont,
  anchor=south
] at (
  \carcBottomTaskCenterX,
  \carcBottomTaskStackTop+\carcTitleGap
)
  {Above \& Below};

\carcTaskCard
  {\carcTaskX+2*\carcDepthX}
  {\carcTaskBottomY+2*\carcDepthY}
  {}

\carcTaskCard
  {\carcTaskX+\carcDepthX}
  {\carcTaskBottomY+\carcDepthY}
  {}

\carcTaskCard
  {\carcTaskX}
  {\carcTaskBottomY}
  {%

    \carcTaskCodeBox{%
      \textbf{\detokenize{def generate_above_below_0001():}}\\[2pt]
      \quad\detokenize{x = rectangle_with_markers(...)}\\
      \quad\detokenize{s = select_objects_above(x)}\\
      \quad\detokenize{y = x.map(s, recolor_object_green)}\\
      \quad\detokenize{return Task(input=x, output=y)}
    }

    \node[
      font=\carcIOFont,
      text=carcInk
    ] at (50.55,16.65)
      {input};

    \begin{scope}[
      shift={(45.00,4.15)},
      scale=\carcTaskGridScale
    ]
      \carcLayoutLayer
        {0}
        {0}
        {\carcInputLayout}
    \end{scope}

    \draw[
      carcInnerFlow,
      draw=carcInk
    ]
      (56.45,9.70)
      --
      (57.85,9.70);

    \node[
      font=\carcIOFont,
      text=carcInk
    ] at (63.85,16.65)
      {output};

    \begin{scope}[
      shift={(58.30,4.15)},
      scale=\carcTaskGridScale
    ]
      \carcLayoutLayer
        {0}
        {0}
        {\carcAboveBelowOutput}
    \end{scope}
  }

\pgfmathsetmacro{\carcConceptRouteX}{
  \carcGateRight + 5.0
}

\pgfmathsetmacro{\carcGateTopOutputY}{
  \carcGateCenterY + 0.19*\carcGateH
}

\pgfmathsetmacro{\carcGateBottomOutputY}{
  \carcGateCenterY - 0.19*\carcGateH
}

\draw[
  carcStrongFlow,
  draw=carcGateInk,
  rounded corners=4pt
]
  (\carcGateRight,\carcGateTopOutputY)
  --
  (\carcConceptRouteX,\carcGateTopOutputY)
  --
  (\carcConceptRouteX,\carcTopTaskCenterY)
  --
  (\carcTaskX,\carcTopTaskCenterY);

\draw[
  carcStrongFlow,
  draw=carcGateInk,
  rounded corners=4pt
]
  (\carcGateRight,\carcGateBottomOutputY)
  --
  (\carcConceptRouteX,\carcGateBottomOutputY)
  --
  (\carcConceptRouteX,\carcBottomTaskCenterY)
  --
  (\carcTaskX,\carcBottomTaskCenterY);

\end{tikzpicture}%
}
\begin{figure}[tbp]
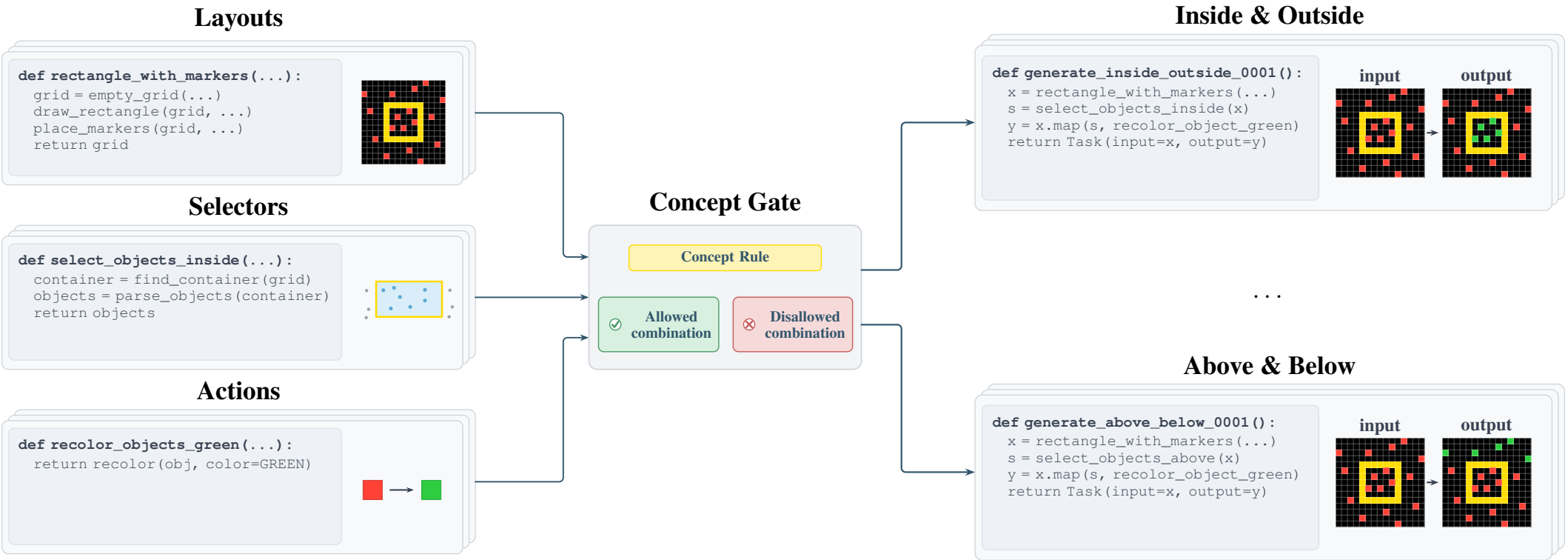

\centering
\resizebox{\linewidth}{!}{\carcSchematicOverview}%
\caption{\textbf{Schematic overview of our ConceptARC task generator}. A shared DSL provides primitive functions for grid layouts, object selection, and actions. Concepts such as ``\emph{Inside and Outside}'' or ``\emph{Above and Below}'' act as semantic gates that determine which primitive combinations are valid for a given concept. Each valid composition defines a task program that can generate multiple input--output examples illustrating the corresponding transformation.}
\label{fig:dataset_generator}
\end{figure}

In this work, we investigate whether recent reasoning models organize their representations of ARC tasks around stable semantic abstractions. ConceptARC provides a natural testbed by grouping ARC-style tasks into 16 spatial and semantic concepts, but its ten manually designed tasks per concept offer limited diversity for assessing representational stability. To scale ConceptARC without costly manual task design, we introduce a generator that produces thousands of new ConceptARC-style tasks automatically. Each task is represented as a program that maps input grids to output grids according to an underlying transformation rule, enabling various examples to be generated per task.

As illustrated in Figure~\ref{fig:dataset_generator}, the key idea behind our task generator is to construct ARC tasks from three basic components: (i) a \emph{grid layout}, which specifies the input scene\textemdash{}for instance, the types of objects present and their spatial arrangement; (ii) a \emph{selector}, which specifies the objects or grid regions to be transformed; and (iii) an \emph{action}, which specifies how the selected objects or grid regions are transformed. This compositional structure allows us to generate a large variety of tasks from a limited set of grid layouts, selectors, and actions~\citep{Xu_Khalil_Sanner_2023}. For example, the layout in Figure~\ref{fig:dataset_generator} contains a hollow yellow rectangle surrounded by red single-cell markers. Pairing it with a selector for objects inside the rectangle and an action that recolors them green produces the ``\emph{Inside and Outside}'' task at the upper right. Replacing the selector with one that selects objects above a reference object produces the ``\emph{Above and Below}'' task at the lower right.

Each concept acts as a semantic gate by constraining the allowable combinations of layouts, selectors, and actions so that generated tasks preserve its defining semantics. For \emph{predicate} concepts (cf. Section~\ref{sec:related_work}) such as ``\emph{Inside and Outside}'' and ``\emph{Above and Below},'' the defining constraint lies primarily in the selector, whereas for \emph{action} concepts such as ``\emph{Extend to Boundary}'' and \emph{Count},'' it lies in the action. Components further declare what they require and provide, ensuring compatibility. For example, a selector that operates on a container can only be paired with layouts that provide one, while an action that operates on a group of objects requires a selector that returns such a group.

Layouts, selectors, and actions are implemented as functions in a shared domain-specific language for constructing and manipulating grid scenes. A task program combines the chosen layout, selector, and action together with their parameters, as shown in Figure~\ref{fig:dataset_generator}. To generate an input--output example, the program constructs an input scene according to the layout and its permitted variability, applies the selector to identify the relevant entities, and uses the action to compute the output. Appendix~\ref{sec:task_generator_details} provides further details on the DSL and the different grid layouts, selectors, and actions.

Although our primary goal is to generate new tasks, the generator is expressive enough to represent all existing ConceptARC tasks. We successfully validate these programs by verifying that the corresponding programs map the original ConceptARC input grids to their manually designed outputs.

\subsection{ConceptARC-100}\label{subsec:conceptarc_100}
Using our compositional task generator described in Section~\ref{sec:data_generator}, we create \textsc{ConceptARC-100}, a dataset containing 100 new tasks for each of the 16 concepts defined by ConceptARC\textemdash{}ten times as many tasks as the original benchmark.\footnote{Link to dataset: \href{https://huggingface.co/datasets/mainlp/ConceptARC-100}{https://huggingface.co/datasets/mainlp/ConceptARC-100}.} Notably, each task is represented by a program that can generate many input--output examples following the same transformation rule.

Following the few-shot format of ARC, we generate three support examples for each task from which the rule must be inferred~\citep{chollet2019measure, chollet2026arcagi2newchallengefrontier}. However, unlike ConceptARC and other ARC benchmarks, we generate ten test examples per task, approximately three to four times as many as in ConceptARC. This larger test set provides a more reliable estimate of whether a model has inferred the underlying rule rather than exploiting incidental input properties such as object colors or arrangements. While it would be prohibitively labor-intensive to create so many examples manually, our programmatic task representations make this possible. In addition, because the task programs are composed of layouts, selectors, and actions, we can report metadata for each task specifying the components used. This metadata enables more fine-grained downstream analyses.

Figures~\ref{fig:layouts_concept_heatmap}--~\ref{fig:action_concept_heatmap} in the Appendix show the distributions of layout, selector, and action families across tasks within each concept, with descriptions provided in Tables~\ref{tab:layout-families}--~\ref{tab:action-families}. A task may involve multiple components. Importantly, concepts describe the semantic properties that tasks are intended to capture, whereas generator components specify how their transformations are implemented. Predicate concepts primarily constrain object selection, while action concepts primarily constrain the transformation. However, there is no one-to-one correspondence: the same component can be used across different concepts, and a concept can be realized through different combinations of components.

\paragraph{Dataset validation.} We manually review all generated tasks to verify (i) that the transformation rule is applied consistently across examples and (ii) that it can be inferred unambiguously from the support set. We find that $6.2\%$ of initially generated tasks have either inconsistent transformation rules or ambiguous support examples, and we subsequently correct these issues. We note that the concepts defined by the original ConceptARC benchmark are not necessarily mutually exclusive. For example, a task that counts objects inside a container may involve both ``\emph{Inside and Outside}'' and ``\emph{Count}.'' Such overlap is expected, as a task's assigned concept describes the primary abstraction it is designed to test, rather than all semantic properties involved in its solution. To assess whether the intended concepts are recognizable, we ask an independent annotator to assign one of the 16 concept labels to each generated task. The annotator identifies the intended concept in $96.75\%$ of cases (Figure~\ref{fig:annotator_concept_confusion} in the Appendix), indicating that concept assignments remain largely identifiable despite potential semantic overlap. We additionally ensure that no tasks are duplicated across concepts. Appendix~\ref{sec:task_generator_details} provides further details on the task generation, deduplication, and annotation process.

\section{Experiments}\label{sec:experiments}
We evaluate a broad range of recent open-weight LRMs on \textsc{ConceptARC-100}, including Gemma-4-31B~\citep{team2026gemma}, Qwen3.5-27B and Qwen3.5-9B~\citep{qwen3.5}, GPT-OSS-120B and GPT-OSS-20B~\citep{openai2025gptoss120bgptoss20bmodel}, Qwen3-32B and Qwen3-8B~\citep{qwen3technicalreport}, and Phi4-Reasoning-Plus~\citep{abdin2025phi}. We evaluate all models in reasoning/thinking mode, using high reasoning effort for GPT-OSS-120B and GPT-OSS-20B. For each model, we use the decoding parameters recommended for reasoning mode on its official HuggingFace page and evaluate the model across three random seeds to account for stochastic decoding. We deliberately focus on open-weight models to ensure that our results are reproducible. More broadly, this choice also reflects our concern about the increasing reliance on proprietary models owned by large companies.

We use the same prompt as ConceptARC and represent two-dimensional grids as sequences of bracketed rows (see Figure~\ref{fig:conceptarc_prompt} in the Appendix for the exact prompt). A response is counted as correct only if the predicted grid exactly matches the target grid in both dimensions and cell values. We parse predicted grids by selecting the last valid sequence of bracketed rows in each response, allowing spaces or commas between entries, and fall back to Python/JSON-style nested lists when necessary. Appendix~\ref{subsec:eval_lrms} provides the decoding parameters and further details for each model.

\subsection{Evaluating concept-level performance}\label{subsec:performance}
Table~\ref{tab:conceptarc_performance} reports accuracy across the 16 concepts. Performance varies substantially across model families and sizes. Qwen3.5-27B achieves the highest overall accuracy at $63.48\%$, followed by Gemma-4-31B at $58.0\%$ and GPT-OSS-120B at $45.41\%$. The Qwen3 models perform substantially worse than the newer Qwen3.5 models: Qwen3-32B, for example, achieves only $10.50\%$ despite having more parameters than Qwen3.5-27B. Within each model family, performance increases with model size. Qwen3.5-27B achieves nearly twice the accuracy of Qwen3.5-9B ($32.89\%$), while GPT-OSS-120B achieves nearly twice the accuracy of GPT-OSS-20B ($28.07\%$).

Table~\ref{tab:conceptarc_original_performance} in Appendix~\ref{subsec:original_conceptarc_performance} reports additional results on the original ConceptARC dataset. Accuracy is notably higher on ConceptARC than on \textsc{ConceptARC-100} (mean accuracy across all LRMs: $54.39\%$ vs. $31.18\%$, respectively). However, because ConceptARC has been publicly available since 2023, we cannot rule out training-data contamination. Despite this difference in accuracy, the relative ordering of model performance is similar across the two datasets.

\begin{table*}[t]
\centering
\small
\setlength{\tabcolsep}{5pt}
\renewcommand{\arraystretch}{1.15}
\caption{\textbf{Accuracy (\%) on \textsc{ConceptARC-100}, averaged across evaluated seeds for each concept.} Bold values indicate the highest reported result in each row. Concepts are grouped alphabetically by type (\predconcept{predicate} and \actconcept{action}).}
\label{tab:conceptarc_performance}
\resizebox{\textwidth}{!}{%
\begin{tabular}{l*{8}{r}:*{2}{r}}
\toprule
Concept
& Gemma4-31B
& Qwen3.5-27B
& Qwen3.5-9B
& \shortstack{GPT-OSS-120B}
& \shortstack{GPT-OSS-20B}
& Qwen3-32B
& Qwen3-8B
& Phi4-Reas.
& VARC
& TRM\\
\midrule
\predconcept{AboveBelow}
& \textbf{75.30} & 73.40 & 43.90 & 56.97 & 37.17 & 13.97 & 6.90 & 13.17 & 37.07 & 14.63 \\
\predconcept{Center}
& 80.03 & \textbf{83.27} & 61.70 & 79.93 & 60.77 & 24.80 & 5.77 & 19.17 & 16.73 & 3.97 \\
\predconcept{FilledNotFilled}
& 60.53 & \textbf{71.13} & 32.43 & 48.20 & 23.37 & 9.83 & 2.40 & 4.33 & 16.90 & 7.27 \\
\predconcept{HorizontalVertical}
& 64.23 & \textbf{73.17} & 34.87 & 47.83 & 27.50 & 11.77 & 5.20 & 7.10 & 9.33 & 10.57 \\
\predconcept{InsideOutside}
& \textbf{61.30} & 58.03 & 31.07 & 33.77 & 26.37 & 9.17 & 2.80 & 3.93 & 38.53 & 13.60 \\
\predconcept{SameDifferent}
& \textbf{44.10} & 36.27 & 10.97 & 17.70 & 10.17 & 5.23 & 3.27 & 2.70 & 12.60 & 7.13 \\
\predconcept{TopBottom2D}
& \textbf{73.63} & 66.23 & 23.77 & 35.60 & 17.60 & 17.37 & 5.20 & 10.00 & 25.43 & 15.27 \\
\predconcept{TopBottom3D}
& 55.40 & \textbf{66.83} & 33.63 & 52.73 & 31.67 & 9.83 & 4.10 & 7.30 & 9.53 & 4.53 \\
\actconcept{CleanUp}
& 49.83 & \textbf{59.43} & 28.37 & 37.30 & 22.57 & 5.63 & 1.80 & 4.90 & 13.37 & 4.77 \\
\actconcept{CompleteShape}
& 70.77 & \textbf{77.03} & 44.77 & 60.80 & 37.33 & 6.17 & 1.90 & 5.40 & 36.13 & 10.43 \\
\actconcept{Copy}
& 50.40 & \textbf{58.37} & 27.83 & 43.23 & 23.07 & 6.53 & 1.67 & 3.10 & 9.57 & 4.57 \\
\actconcept{Count}
& 27.53 & \textbf{33.30} & 16.00 & 16.90 & 9.80 & 4.33 & 2.53 & 2.97 & 5.27 & 5.93 \\
\actconcept{ExtendToBoundary}
& 76.87 & \textbf{80.73} & 51.00 & 69.57 & 46.90 & 17.27 & 7.47 & 14.27 & 32.97 & 10.70 \\
\actconcept{ExtractObjects}
& 53.13 & \textbf{69.70} & 35.93 & 53.80 & 30.67 & 9.20 & 4.20 & 4.57 & 6.97 & 2.80 \\
\actconcept{MoveToBoundary}
& 56.60 & \textbf{68.50} & 40.87 & 56.17 & 38.33 & 15.87 & 9.23 & 9.87 & 10.93 & 2.70 \\
\actconcept{Order}
& 28.30 & \textbf{40.27} & 9.07 & 16.13 & 5.87 & 1.07 & 0.17 & 0.17 & 0.10 & 0.03 \\
\midrule
\predconcept{\textbf{Avg. Predicate Concepts}}
& 64.32 & \textbf{66.04} & 34.04 & 46.59 & 29.33 & 12.75 & 4.45 & 8.46 & 20.77 & 9.62 \\
\actconcept{\textbf{Avg. Action Concepts}}
& 51.68 & \textbf{60.92} & 31.73 & 44.24 & 26.82 & 8.26 & 3.62 & 5.65 & 14.41 & 5.24 \\
\midrule
\midrule
\textbf{Average}
& 58.00 & \textbf{63.48} & 32.89 & 45.41 & 28.07 & 10.50 & 4.04 & 7.06 & 17.59 & 7.43 \\
\bottomrule
\end{tabular}%
}
\end{table*}

\paragraph{LRMs perform well on spatial concepts but struggle with comparison and counting.}
Performance across concepts reveals several consistent patterns. Stronger models achieve high accuracy on spatial concepts such as ``\emph{Above and Below}'', ``\emph{Center}'', and ``\emph{Top / Bottom (2D)}'', whose transformations depend primarily on spatial relations. Qwen3.5-27B, for example, achieves $73.40\%$ accuracy on ``\emph{Above and Below}'' and $83.27\%$ on ``\emph{Center}''. In contrast, models perform worse on concepts requiring comparisons between objects. Qwen3.5-27B achieves only $40.27\%$ on ``\emph{Order}'' and $36.27\%$ on ``\emph{Same / Different}''. ``\emph{Order}'' tasks often require objects to be compared and rearranged by properties such as height, width, marker position, or cell count, whereas ``\emph{Same / Different}'' tasks commonly require comparisons of shape, color, or internal structure (we invite readers to look at some examples using our \href{https://mainlp.github.io/conceptarc-100-project-page/data_viewer.html}{data viewer}). Performance is also low on ``\emph{Count}'', which requires counting objects, cells, or other selected features. One possible explanation is the task implementation: whereas ConceptARC typically implements counting as standalone bars, in our dataset, counts are encoded as colored bars or stripes overlaid on the input grid. This representation may be more complex and less common in the models' training data. Finally, LRMs perform slightly better on \emph{predicate} than on \emph{action} concepts (see Section~\ref{sec:related_work}). Qwen3.5-27B, for example, achieves $66.04\%$ average accuracy on predicate concepts and $60.92\%$ on action concepts.

\begin{figure}[tbp]
\centering
\includegraphics[width=0.855\textwidth]{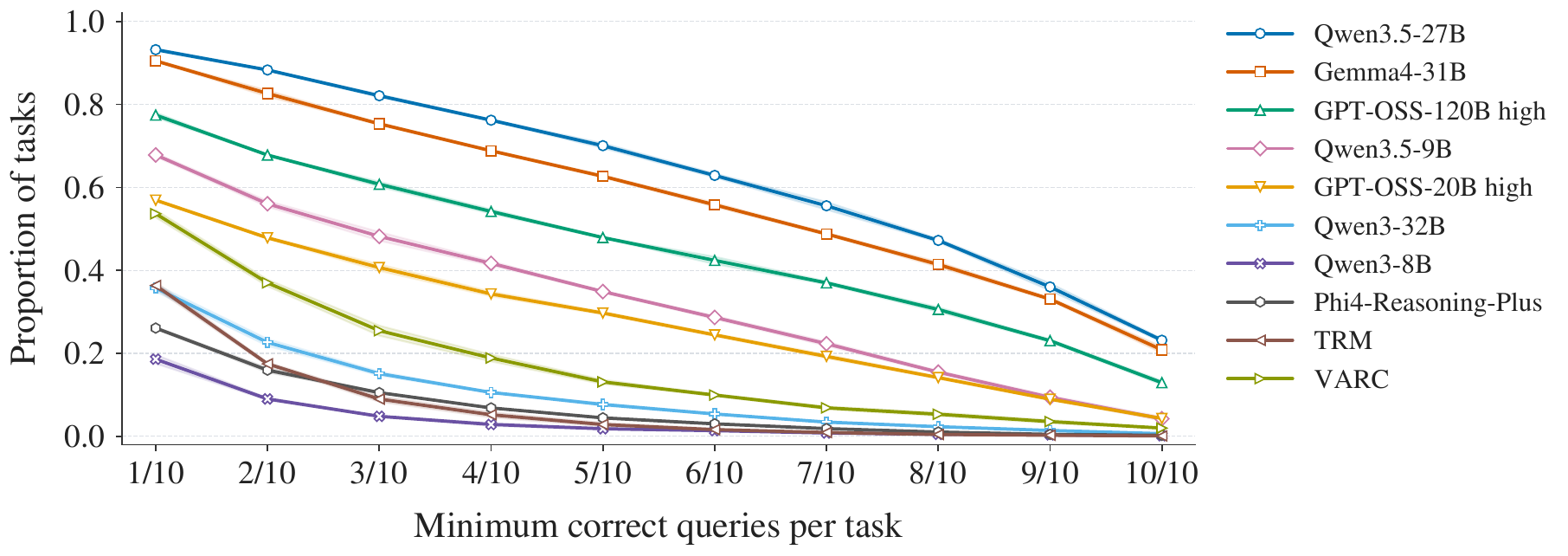}
\caption{\textbf{Large reasoning models do not solve tasks robustly}. Proportion of tasks for which a model correctly solves at least $\text{X}/10$ test queries. Performance decreases steadily as the required number of correct queries increases, indicating limited capacity to solve tasks robustly. Values are averaged over three random seeds; shaded regions indicate standard deviations.
}
\label{fig:lrm_robustness_eval}
\end{figure}

\paragraph{LRMs struggle to solve test queries consistently.} As described in Section~\ref{subsec:conceptarc_100}, each \textsc{ConceptARC-100} task contains three support examples and ten test queries. Figure~\ref{fig:lrm_robustness_eval} shows the proportion of tasks for which a model solves at least $X$ of the ten test queries, for different values of $X$. Across all models, this proportion decreases steadily as $X$ increases, indicating that models often fail to apply the inferred rule consistently across test queries. Qwen3.5-27B, for example, solves at least one query for $93.21\%$ of tasks but all ten queries for only $23.17\%$.

A potential concern is that some test queries may not be solvable from the corresponding support examples. We therefore manually review each task to verify that (i) the transformation rule is applied consistently across examples and (ii) it can be inferred unambiguously from the support set, as described in Section~\ref{subsec:conceptarc_100}. Because human error remains possible, we additionally compute the fraction of tasks for which all ten test queries are solved by at least one model. At $68.88\%$, this fraction is substantially higher than the $23.17\%$ of tasks fully solved by Qwen3.5-27B, providing additional evidence that the low performance primarily reflects model failures rather than unsolvable queries.

\subsubsection{Assessing models trained from scratch}\label{subsubsec:assessing_trained_from_scratch}
We complement our evaluation of pretrained LRMs with two models trained from scratch on \textsc{ConceptARC-100}: VARC-ViT-18M, following~\citet{hu2025arcvisionproblem}, and Tiny Recursive Model-7M, following~\citet{jolicoeurmartineau2025morerecursivereasoningtiny}, hereafter VARC and TRM. Both are substantially smaller than the LRMs evaluated above, with approximately 18M and 7M parameters, respectively. We use each model's default configuration; Appendix~\ref{subsec:training_details} provides further details for each model.

\paragraph{Training \& inference.} We train VARC for 1{,}000 epochs and TRM for 10{,}000 epochs on the three support examples of each of the 1{,}600 \textsc{ConceptARC-100} tasks. Both models learn a task-specific embedding shared across all examples of a task, intended to encode its underlying rule. VARC processes patch-based grid representations with a 10-layer Transformer, whereas TRM operates on grid-cell tokens and recursively reuses two Transformer layers with adaptive halting. Both models are trained end-to-end to predict the output grid; further architectural and training details are provided in Appendix~\ref{subsec:training_details}. At inference time, each test input is paired with its learned task embedding to test generalization to unseen inputs from the same task. We do not use test-time training, task augmentation, or multiple test-time predictions, as our goal is to assess how well these models learn the new \textsc{ConceptARC-100} tasks from their support examples, which concepts are most challenging, and how task information is represented in their embeddings.

\paragraph{VARC and TRM show similar concept-level trends to pretrained LRMs.} Table~\ref{tab:conceptarc_performance} reports VARC and TRM performance alongside the pretrained LRMs. VARC achieves $17.59\%$ average accuracy, outperforming Phi4-Reasoning-Plus and both Qwen3 models despite having far fewer parameters, while remaining below larger, more recent models such as Qwen3.5-27B and Gemma-4-31B. TRM achieves $7.43\%$ accuracy. Performance varies substantially across concepts but broadly follows the trends observed for pretrained LRMs. VARC performs best on ``\emph{Inside and Outside}'' and ``\emph{Above and Below}'', reaching approximately $38\%$, and worst on ``\emph{Order}'' and ``\emph{Count}'', at $0.10\%$ and $5.27\%$, respectively. ``\emph{Above and Below}'' is also TRM's second-best concept, after ``\emph{Top / Bottom (2D)}'', while ``\emph{Order}'' is again the most difficult, with $0.03\%$ accuracy. TRM outperforms VARC on two concepts, including ``\emph{Count},'' potentially indicating a benefit from its recursive computation. Across concepts, VARC and TRM performance is strongly correlated (Pearson $r=0.84$; Spearman $\rho=0.74$), suggesting similar relative concept difficulty for both models.

\paragraph{VARC and TRM also struggle to solve tasks consistently.} As with the LRMs, we measure the proportion of tasks for which VARC and TRM solve at least $X$ of the ten test queries. Only $2\%$ of tasks are fully solved by VARC and $0.1\%$ by TRM, and these fractions could decrease further with additional test queries. One possible explanation is that models struggle to learn generalizable rules from only three support, instead capturing partial rules or shortcuts. To test this hypothesis, we use the task programs to increase the support set eightfold to 24 examples per task and retrain VARC. On the same test queries, the resulting model achieves $77.4\%$ query accuracy and fully solves $58.3\%$ of tasks, outperforming all evaluated LRMs. This result suggests that the amount of task-specific training data is an important performance bottleneck.

\subsection{Understanding how models represent tasks internally}\label{subsec:repr_analysis}
The behavioral evaluation in Section~\ref{subsec:performance} measures how well models solve tasks from different concepts, but does not reveal how these tasks are represented internally or whether their representations capture abstractions that generalize across tasks. We therefore analyze the models' internal task representations and ask whether they are organized according to semantically meaningful properties.

\paragraph{Extracting task representations.} To examine how LRMs represent a task, we extract hidden states from the residual stream at different layers. Specifically, we use the token position at the end of the third support example, corresponding to the closing bracket after the final row of its output grid. At this point, the model has processed all three support examples and should have sufficient information to infer the underlying task rule. To obtain stable representations, we exploit the ability of our data generator to produce multiple support examples for each task. For each task $t$, we construct eight support sets, $s_1,\ldots,s_8$, containing different examples. From each set $s_v$, we extract the hidden state $h_{t,v}^{(l)}$ at layer $l$ and average across support sets: $h_t^{(l)} = \frac{1}{8}\sum_{v=1}^{8} h_{t,v}^{(l)}$. We then center these representations by subtracting the mean representation across all tasks at each layer.

\paragraph{LRMs organize tasks by layout in early layers and actions in later layers.} To characterize how models organize tasks in their representation space, we use the task metadata from Section~\ref{sec:data_generator}, which specifies the \emph{layout}, \emph{selector}, and \emph{action}, together with each task's \emph{concept} category. For each of these four label sets, we measure whether tasks sharing a label are closer in representation space than tasks with different labels. For example, let $h_{t_i}^{(l)}$ denote the representation of task $t_i$ with action label $a_i$. We measure the probability that a task $t_j$ with the same action label is closer to $t_i$ than a task $t_k$ with a different action label: $d_{\mathrm{cos}}\!\left(h_{t_i}^{(l)}, h_{t_j}^{(l)}\right) < d_{\mathrm{cos}}\!\left(h_{t_i}^{(l)}, h_{t_k}^{(l)}\right)$. We evaluate all eligible comparisons, excluding self-comparisons and assigning half credit to ties. Scores are first averaged across tasks within each label and then across labels. A score of $0.5$ corresponds to chance-level separation; higher values indicate stronger organization according to the corresponding property.

\begin{figure}[tpb]
    \centering
    \begin{subfigure}[t]{0.40\linewidth}
        \centering
        \includegraphics[width=\linewidth]{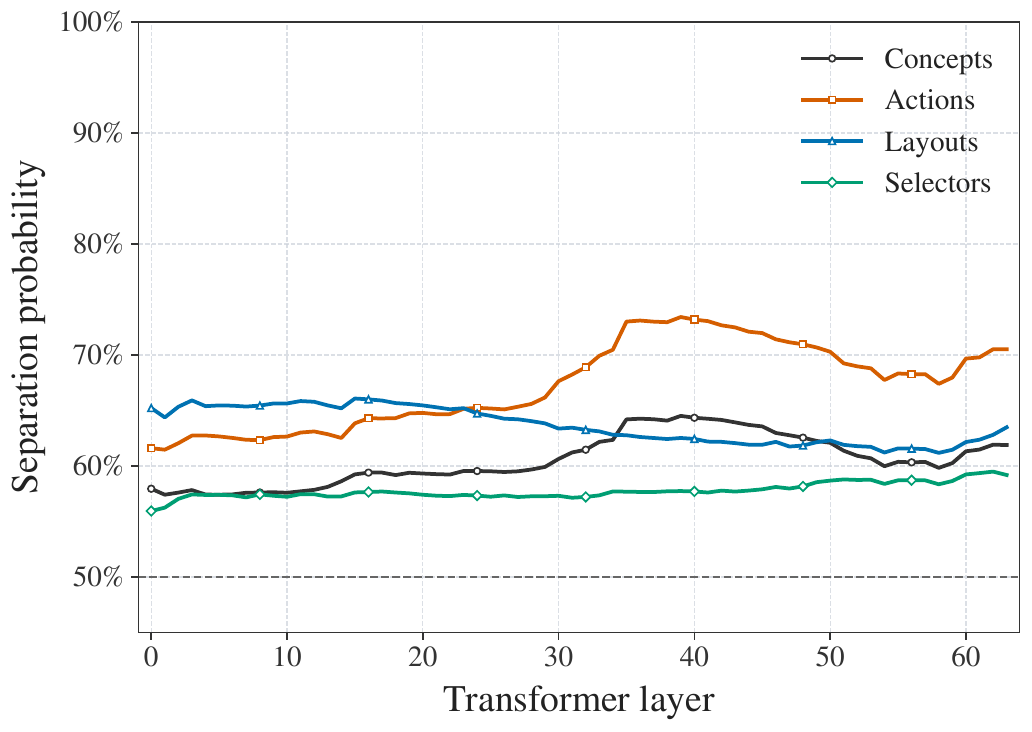}
        \caption{Label set comparison}
        \label{fig:qwen35_27b_separation_probability}
    \end{subfigure}
    \hspace{5mm}
    \begin{subfigure}[t]{0.40\linewidth}
        \centering
        \includegraphics[width=\linewidth]{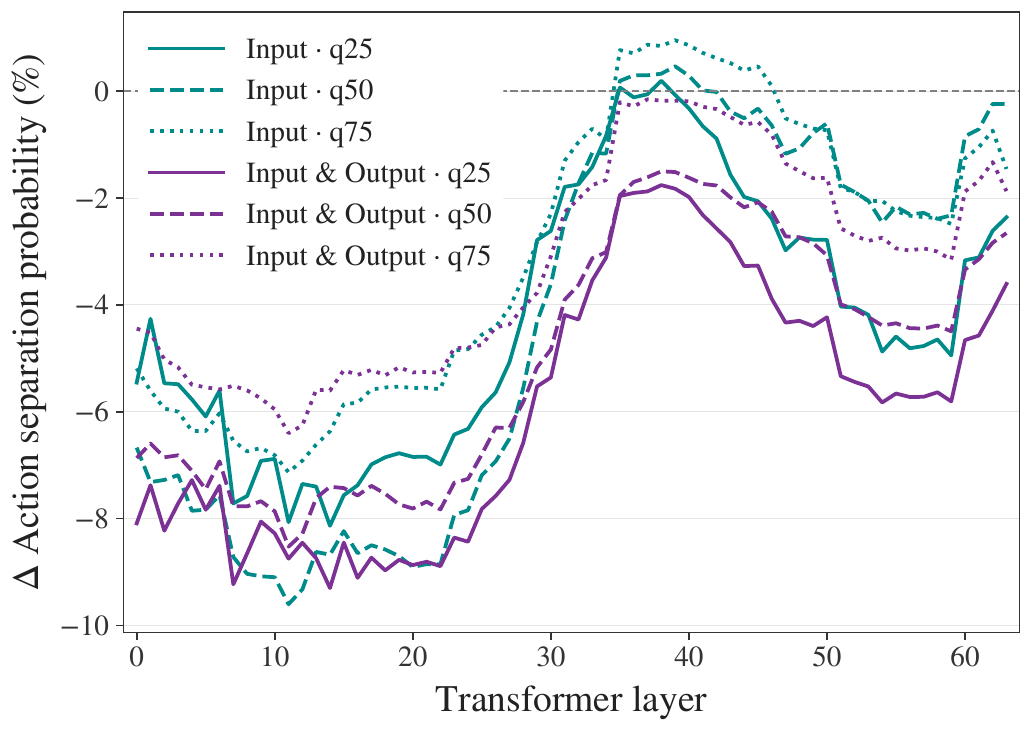}
        \caption{Controlled setup}
        \label{fig:qwen35_27b_separation_probability_controlled}
    \end{subfigure}
    \caption{\textbf{Qwen3.5-27B representations shift from layout- to action-based organization across layers.}
(a) Early layers organize tasks primarily by grid layout, while action-based organization becomes stronger in later layers, suggesting a shift from spatial properties toward the transformations required to solve each task.
(b) Change in action separation probability relative to the uncontrolled baseline under six visual matching conditions. Controls restrict comparisons of tasks by visual similarity of input grids alone or both input and output grids. Visual matching substantially reduces action-based separation in early and late layers but has a smaller effect in intermediate layers.}
    \label{fig:qwen35_separation_probability_uncontrolled_vs_controlled}
\end{figure}

Figure~\ref{fig:qwen35_27b_separation_probability} shows the separation probability of Qwen3.5-27B by \emph{layout}, \emph{selector}, \emph{action}, and \emph{concept} across layers. In early layers, tasks are organized primarily by grid layout. As information propagates through the model, this organization shifts: separation by layout gradually decreases, while separation by action rises from $61.6\%$ at layer~0 to a peak of $73.4\%$ at layer~39. At that layer, separation reaches $64.5\%$ for concept, $62.5\%$ for layout, and $57.7\%$ for selector. Action-based separation subsequently declines but remains the strongest through the final layer, where it reaches $70.5\%$. Toward the final layers, separation probabilities increase slightly across all four label categories, with layout-based separation partially recovering toward its initial level. Overall, action-based separation exceeds concept-based separation across all layers, whereas selector-based separation remains comparatively weak. We observe similar trends for other models, including Gemma4-31B, GPT-OSS-120B, and Qwen3.5-9B (Figures~\ref{fig:gemma4_31b_separation_probability}--\ref{fig:separation_probability_other_models} in the Appendix).

\paragraph{Action separation remains high when controlling for visual grid features.}
We further analyze whether the strong action-based separation in intermediate and later layers can be explained by similarities in visual grid features that tasks with the same action label tend to share, but tasks with different action labels do not. To this end, we recompute the action separation probability for Qwen3.5-27B across layers while controlling for layout family, action frequency, and visual grid properties, including grid dimensions, color histograms, and changed-cell counts. For each input--output support example, we compute 30 visual features (see Appendix~\ref{subsubsec:controlled_action_sep} for details) and concatenate the feature vectors of the three support examples into a single vector $f_{\mathrm{vis}} \in \mathbb{R}^{90}$.

When computing action separation for a reference task $t_i$ with action label $a_i$, we restrict comparisons to pairs of tasks $t_j$ and $t_k$, whose visual feature vectors $f_{\mathrm{vis},j}$ and $f_{\mathrm{vis},k}$ are similar. In all control conditions, tasks $t_j$ and $t_k$ must belong to the same layout family, and action frequencies are balanced across comparisons. To control for visual similarity, we compute the Euclidean distances between the feature vectors of all eligible task pairs and retain only pairs whose distance falls below the 25th, 50th, or 75th percentile of this distribution (q25/q50/q75), with q25 imposing the strictest matching. We further distinguish between visual features computed from input grids alone and those computed from both input and output grids, yielding six matching conditions. Figures~\ref{fig:visual_distance_corner_removal} and~\ref{fig:visual_distance_mark_center} in the Appendix illustrate how stricter matching yields more visually similar task pairs.

Figure~\ref{fig:qwen35_27b_separation_probability_controlled} compares the resulting action separation probabilities across these six conditions with the uncontrolled baseline from Figure~\ref{fig:qwen35_27b_separation_probability}. Visual matching substantially reduces action-based separation in early layers. Near its peak in intermediate layers, however, separation decreases by only $1.8\%$, from $73.4\%$ to $71.6\%$, even under the strictest matching condition. For comparison, we also compute action separation directly from visual feature vectors rather than the model's task representations, using 132 features. Under the same matching condition, this visual-feature baseline achieves only $59.96\%$ separation (see Appendix~\ref{subsubsec:controlled_action_sep} for details). These results suggest that the observed action-based organization is not solely driven by the measured visual properties of the support examples.

\paragraph{VARC's task embeddings show similar organization by action.}
We perform the same separation analysis on VARC, trained on \textsc{ConceptARC-100}. As described in Section~\ref{subsubsec:assessing_trained_from_scratch}, VARC learns task-embedding vectors during training and uses them at inference time to provide information about the underlying task rules. We treat these embeddings as task representations and compute the corresponding separation probabilities. As with Qwen3.5-27B, VARC's representations are organized predominantly by action, with separation probabilities of $57.1\%$ for layout, $54.8\%$ for selector, $86.2\%$ for action, and $70.8\%$ for concept. Action-based separation remains similarly high ($86.4\%$) under the strictest visual matching condition (\emph{input \& output $\cdot$ q25}).

\begin{figure}[tbp]
\centering
\resizebox{0.9575\textwidth}{!}{%
    \input{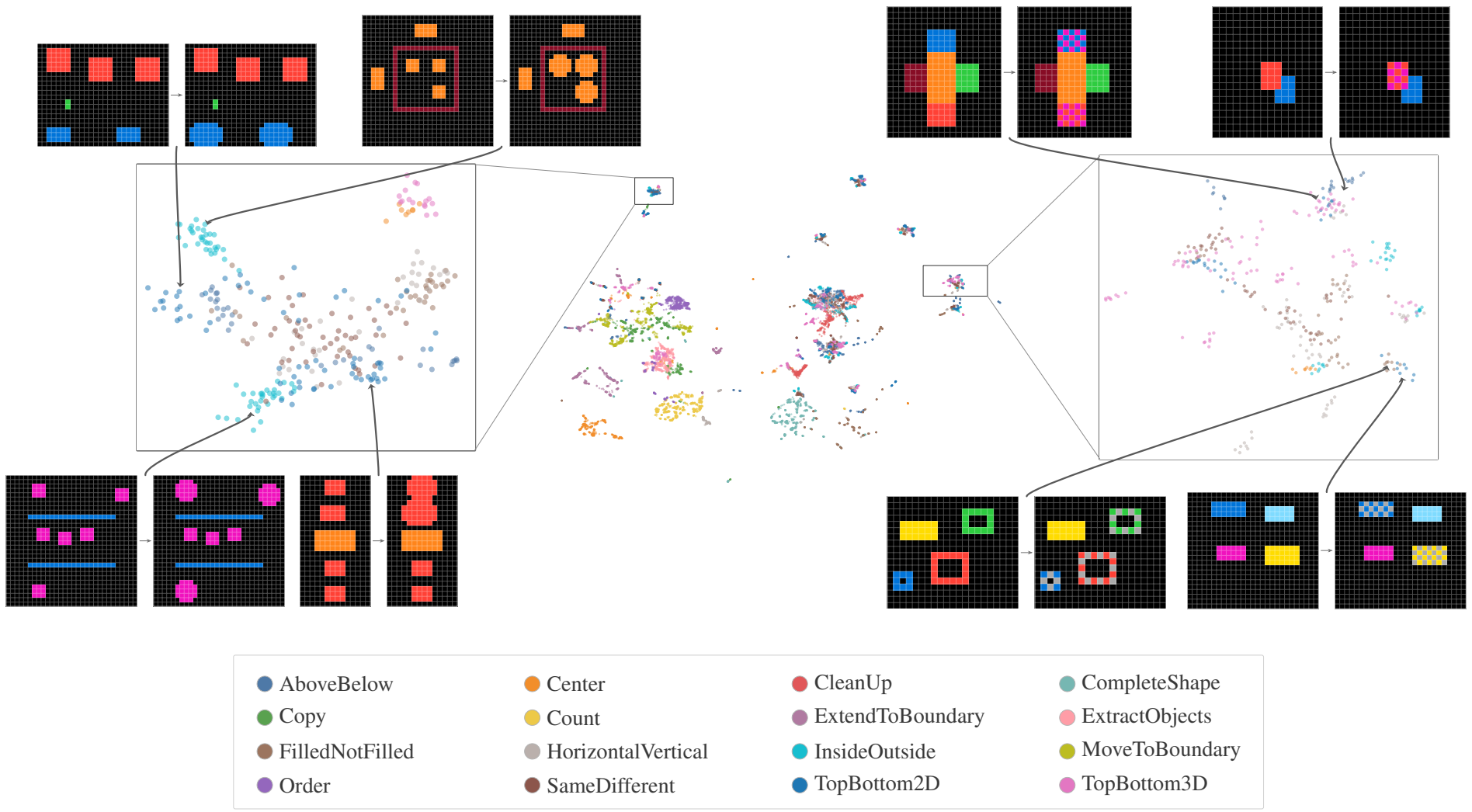}
}
\caption{\textbf{VARC learns semantically structured task embeddings.} UMAP projection of task embeddings learned by VARC. Tasks requiring similar transformation operations form distinct clusters, including \emph{extension of object-boundary cells} (left) and \emph{checkerboard coloring of objects} (right). Enlarged views reveal finer semantic structure: boundary-expanding tasks separate according to spatial relations, such as \emph{outside} versus \emph{inside} a region and \emph{above} versus \emph{below} a reference object, while checkerboard-coloring tasks further separate by color, including \emph{pink} and \emph{gray} patterns.}
\label{fig:varc_task_embeddings}
\end{figure}

Figure~\ref{fig:varc_task_embeddings} visualizes VARC's task-embeddings using a two-dimensional UMAP projection of pairwise cosine distances, with 30 neighbors and a minimum-distance of $0.1$. Each point represents a task, colored by concept. We observe surprisingly interpretable clusters corresponding to specific actions, even when those transformations arise from substantially different rules and scenarios across tasks. For instance, the cluster enlarged on the left contains tasks from different concepts that all expand object-boundary cells. Its subclusters exhibit additional semantic structure: tasks expanding boundary cells from objects \emph{outside} a region separate from those operating on objects \emph{inside} a container, while tasks acting on objects \emph{above} a reference separate from those acting \emph{below} it.

The cluster enlarged on the right contains tasks from different concepts that color selected objects with a checkerboard pattern. Within this cluster, tasks are further organized by checkerboard color, with distinct subclusters for, e.g., pink and gray patterns. Readers can further explore this representation space using our \href{https://mainlp.github.io/conceptarc-100-project-page/index.html#embeddings}{interactive embedding space viewer}. A UMAP visualization of Qwen3.5-27B's task representations at layer~41 is provided in Figure~\ref{fig:qwen_layer41_embeddings} in the Appendix.

Together, these results suggest that models exhibit structured representations of regularities shared across ARC tasks. This organization does not simply reproduce the benchmark's concepts; instead, much of the structure reflects similarities in the transformation operations applied to the input grids.

 \section{Conclusion}\label{sec:conclusion}
We introduced \textsc{ConceptARC at Scale}, a task generator for creating thousands of ARC-style tasks based on the 16 concepts from ConceptARC. Using this generator, we constructed \textsc{ConceptARC-100}, a dataset of 1{,}600 novel, manually reviewed tasks. We evaluated LRMs and models trained from scratch on this dataset, characterized their performance across concepts, and found that even the best-performing model struggles to solve all test inputs of a task consistently. We further analyzed models' internal task representations and found semantic structure organized around transformation operations, even across tasks with diverse rules and layouts. This suggests that models form shared abstractions of transformations that generalize across diverse tasks.

\paragraph{Limitations \& Future Directions.}
A key limitation of our analysis is that the relationship between representation structure and model behavior remains correlational. Although the representation spaces exhibit semantic structure aligned with transformation operations, we have not established that models rely on these representations during inference. Future work could intervene on representations associated with identified clusters and measure the resulting behavioral changes. For example, shifting a task representation toward a specific cluster could test whether these representations causally influence the model toward the corresponding transformation.

\subsection*{AI use statement}
We acknowledge the use of generative AI tools for minor language revisions, such as grammar checks and reformulations. In addition, we used generative AI tools such as Codex to assist with code implementation, particularly in developing the DSL described in Section~\ref{sec:data_generator}. We reviewed all code generated by Codex and tested it for correctness. We take responsibility for the final content of this work, including text, claims, and artifacts produced with the aid of generative AI.

\subsection*{Reproducibility statement}
To ensure the reproducibility of our experiments, we make all code for evaluating models on \textsc{ConceptARC-100} publicly available at \href{https://github.com/mainlp/conceptarc-100-eval}{https://github.com/mainlp/conceptarc-100-eval}. Details of our data generation procedure is described in Section~\ref{sec:data_generator} and Appendix~\ref{sec:task_generator_details}. A description of how we evaluate models is presented in Section~\ref{sec:experiments}, with further details about the exact model parameters given in Appendix~\ref{subsec:eval_lrms}. A description of the models we train on \textsc{ConceptARC-100} can be found in Section~\ref{subsubsec:assessing_trained_from_scratch}, with further details of the training procedure given in Appendix~\ref{subsec:training_details}. Finally, we make \textsc{ConceptARC-100} publicly available and provide a dataset viewer, as well as a interactive tool to explore VARC's task-embedding space on our project page: \href{https://mainlp.github.io/conceptarc-100-project-page}{https://mainlp.github.io/conceptarc-100-project-page}.


\subsubsection*{Acknowledgments}
This project resulted from a research visit by PM to NYU and Princeton University. We would like to thank the LMU–NYU Cooperation Program for funding this visit and making this collaboration possible. Furthermore, we would like to thank Jasmin Orth and Xinyuan Cheng for their valuable work and support
in data annotation. We would also like to thank the anonymous reviewers for their constructive feedback. AD is an F.R.S.-FNRS postdoctoral researcher; his research contributions to this project benefited from the computational resources of Lucia, the Tier-1 supercomputer of the Walloon Region, funded under grant No. 1910247. We further acknowledge the support provided to RY by the BANYAN Institute, funded by Sandia National Laboratories' Laboratory Directed Research and Development program. Finally, we acknowledge the support provided to BP through the ERC Consolidator Grant DIALECT 101043235.

\bibliography{iclr2027_conference}

@article{gentner_1983,
author = {Gentner, Dedre},
title = {Structure-Mapping: A Theoretical Framework for Analogy},
journal = {Cognitive Science},
volume = {7},
number = {2},
pages = {155-170},
doi = {https://doi.org/10.1207/s15516709cog0702\_3},
url = {https://onlinelibrary.wiley.com/doi/abs/10.1207/s15516709cog0702_3},
eprint = {https://onlinelibrary.wiley.com/doi/pdf/10.1207/s15516709cog0702_3},
year = {1983}
}

@article{gentner_2017,
author = {Gentner, Dedre and Hoyos, Christian},
title = {Analogy and Abstraction},
journal = {Topics in Cognitive Science},
volume = {9},
number = {3},
pages = {672-693},
doi = {https://doi.org/10.1111/tops.12278},
url = {https://onlinelibrary.wiley.com/doi/abs/10.1111/tops.12278},
eprint = {https://onlinelibrary.wiley.com/doi/pdf/10.1111/tops.12278},
year = {2017}
}

@article{beger2025ai,
  title={Do AI Models Perform Human-like Abstract Reasoning Across Modalities?},
  author={Beger, Claas and Yi, Ryan and Fu, Shuhao and Denton, Kaleda and Moskvichev, Arseny and Tsai, Sarah W and Rajamanickam, Sivasankaran and Mitchell, Melanie},
  journal={arXiv preprint arXiv:2510.02125},
  year={2025}
}

@article{opielka2025analogical,
  title={Analogical reasoning inside large language models: Concept vectors and the limits of abstraction},
  author={Opie{\l}ka, Gustaw and Rosenbusch, Hannes and Stevenson, Claire E},
  journal={arXiv preprint arXiv:2503.03666},
  year={2025}
}

@inproceedings{hendel-etal-2023-context,
    title = "In-Context Learning Creates Task Vectors",
    author = "Hendel, Roee  and
      Geva, Mor  and
      Globerson, Amir",
    editor = "Bouamor, Houda  and
      Pino, Juan  and
      Bali, Kalika",
    booktitle = "Findings of the Association for Computational Linguistics: EMNLP 2023",
    month = dec,
    year = "2023",
    address = "Singapore",
    publisher = "Association for Computational Linguistics",
    url = "https://aclanthology.org/2023.findings-emnlp.624/",
    doi = "10.18653/v1/2023.findings-emnlp.624",
    pages = "9318--9333"
}

@inproceedings{ICLR2024_4ae163cb,
 author = {Todd, Eric and Li, Millicent and Sen Sharma, Arnab and Mueller, Aaron and Wallace, Byron and Bau, David},
 booktitle = {International Conference on Learning Representations},
 editor = {B. Kim and Y. Yue and S. Chaudhuri and K. Fragkiadaki and M. Khan and Y. Sun},
 pages = {17282--17333},
 title = {Function Vectors in Large Language Models},
 url = {https://proceedings.iclr.cc/paper_files/paper/2024/file/4ae163cb8788970e53b4fd9578141139-Paper-Conference.pdf},
 volume = {2024},
 year = {2024}
}

@inproceedings{NEURIPS2022_182aed03,
 author = {Acquaviva, Sam and Pu, Yewen and Kryven, Marta and Sechopoulos, Theodoros and Wong, Catherine and Ecanow, Gabrielle and Nye, Maxwell and Tessler, Michael and Tenenbaum, Josh},
 booktitle = {Advances in Neural Information Processing Systems},
 doi = {10.52202/068431-0270},
 editor = {S. Koyejo and S. Mohamed and A. Agarwal and D. Belgrave and K. Cho and A. Oh},
 pages = {3731--3743},
 publisher = {Curran Associates, Inc.},
 title = {Communicating Natural Programs to Humans and Machines},
 url = {https://proceedings.neurips.cc/paper_files/paper/2022/file/182aed0379591ebd1d655b2bdc152075-Paper-Datasets_and_Benchmarks.pdf},
 volume = {35},
 year = {2022}
}

@article{hodel2024addressing,
  title={Addressing the abstraction and reasoning corpus via procedural example generation},
  author={Hodel, Michael},
  journal={arXiv preprint arXiv:2404.07353},
  year={2024}
}

@article{
moskvichev2023the,
title={The Concept{ARC} Benchmark: Evaluating Understanding and Generalization in the {ARC} Domain},
author={Arsenii Kirillovich Moskvichev and Victor Vikram Odouard and Melanie Mitchell},
journal={Transactions on Machine Learning Research},
issn={2835-8856},
year={2023},
url={https://openreview.net/forum?id=8ykyGbtt2q},
note={}
}

@article{chollet2019measure,
  title={On the measure of intelligence},
  author={Chollet, Fran{\c{c}}ois},
  journal={arXiv preprint arXiv:1911.01547},
  year={2019}
}

@inproceedings{hu2025arcvisionproblem,
  title = {{ARC Is a Vision Problem!}},
  author = {Hu, Keya and Cy, Ali and Qiu, Linlu and Ding, Xiaoman Delores and Wang, Runqian and Zhu, Yeyin Eva and Andreas, Jacob and He, Kaiming},
  booktitle = {Proceedings of the IEEE/CVF Conference on Computer Vision and Pattern Recognition},
  pages = {2537--2546},
  month = {June},
  year = {2026},
  url = {https://openaccess.thecvf.com/content/CVPR2026/html/Hu_ARC_Is_a_Vision_Problem_CVPR_2026_paper.html}
}

@inproceedings{
dosovitskiy2021animageisworth,
title={{An Image is Worth 16x16 Words: Transformers for Image Recognition at Scale}},
author={Alexey Dosovitskiy and Lucas Beyer and Alexander Kolesnikov and Dirk Weissenborn and Xiaohua Zhai and Thomas Unterthiner and Mostafa Dehghani and Matthias Minderer and Georg Heigold and Sylvain Gelly and Jakob Uszkoreit and Neil Houlsby},
booktitle={International Conference on Learning Representations},
year={2021},
url={https://openreview.net/forum?id=YicbFdNTTy}
}

@misc{shu2026loopvit,
  title = {{LoopViT: Scaling Visual {ARC} with Looped Transformers}},
  author = {Shu, Wen-Jie and Qiu, Xuerui and Zhu, Rui-Jie and Chen, Harold Haodong and Liu, Yexin and Yang, Harry},
  year = {2026},
  eprint = {2602.02156},
  archivePrefix = {arXiv},
  url = {https://arxiv.org/abs/2602.02156}
}

@misc{deliege2026implicitruleinduction,
      title={{Implicit Rule Induction with Test-Time Task Embeddings in ARC-like Tasks}}, 
      author={Adrien Deliège and Claas Beger and Marc Van Droogenbroeck and Melanie Mitchell},
      year={2026},
      eprint={2609.21181},
      archivePrefix={arXiv},
      primaryClass={cs.AI},
      url={https://arxiv.org/abs/2609.21181}, 
}

@misc{jolicoeurmartineau2025morerecursivereasoningtiny,
  title = {{Less is More: Recursive Reasoning with Tiny Networks}},
  author = {Jolicoeur-Martineau, Alexia},
  year = {2025},
  eprint = {2510.04871},
  archivePrefix = {arXiv},
  primaryClass = {cs.LG},
  url = {https://arxiv.org/abs/2510.04871}
}

@misc{wang2025hrm,
  title = {{Hierarchical Reasoning Model}},
  author = {Wang, Guan and Li, Jin and Sun, Yuhao and Chen, Xing and Liu, Changling and Wu, Yue and Lu, Meng and Song, Sen and Yadkori, Yasin Abbasi},
  year = {2025},
  eprint = {2506.21734},
  archivePrefix = {arXiv},
  url = {https://arxiv.org/abs/2506.21734}
}

@inproceedings{movahedi2026fixedpointreasonersstableadaptive,
title={{Fixed-Point Reasoners: Stable and Adaptive Deep Looped Transformers}},
author={Sajad Movahedi and Vera Milovanovi{\'c} and Shlomo Libo Feigin and Alexander Theus and Thomas Hofmann and Valentina Boeva and T. Konstantin Rusch and Antonio Orvieto},
booktitle={ICML 2026 Workshop on High-dimensional Learning Dynamics},
year={2026},
url={https://openreview.net/forum?id=4wOixCRrge}
}

@inproceedings{baek2026generativerecursivereasoning,
      title={{Generative Recursive Reasoning}}, 
      author={Junyeob Baek and Mingyu Jo and Minsu Kim and Yoshua Bengio and Sungjin Ahn},
      booktitle={ICLR 2026 Workshop on AI with Recursive Self-Improvement},
      year={2026},
      url={https://openreview.net/pdf?id=Vxu6kcIjwV},
}

@misc{chollet2026arcagi2newchallengefrontier,
      title={{ARC-AGI-2: A New Challenge for Frontier AI Reasoning Systems}}, 
      author={Francois Chollet and Mike Knoop and Gregory Kamradt and Bryan Landers and Henry Pinkard},
      year={2026},
      eprint={2505.11831},
      archivePrefix={arXiv},
      primaryClass={cs.AI},
      url={https://arxiv.org/abs/2505.11831}, 
}

@inproceedings{Kim2022Playgrounds,
        title = {{Playgrounds for Abstraction and Reasoning}},
        author = {Kim, Subin and Phunyaphibarn, Prin and Ahn, Donghyun and Kim, Sundong},
        booktitle = {NeurIPS 2022 Workshop on Neuro Causal and Symbolic AI (nCSI)},
        pages = {1-6},
        month = Nov,
        year = {2022},
	d2b-version = {1.0},
        url = {https://nips.cc/virtual/2022/57706}
}

@article{
xu2024llms,
title={{LLMs and the Abstraction and Reasoning Corpus: Successes, Failures, and the Importance of Object-based Representations}},
author={Yudong Xu and Wenhao Li and Pashootan Vaezipoor and Scott Sanner and Elias Boutros Khalil},
journal={Transactions on Machine Learning Research},
issn={2835-8856},
year={2024},
url={https://openreview.net/forum?id=E8m8oySvPJ},
note={}
}

@inproceedings{acquaviva2021larc,
  title     = {{LARC: Language annotated Abstraction and Reasoning Corpus}},
  author    = {Acquaviva, Sam and Pu, Yewen and Nye, Maxwell and Wong, Catherine and Tessler, Michael Henry and Tenenbaum, Josh},
  booktitle = {Proceedings of the Annual Meeting of the Cognitive Science Society},
  volume    = {43},
  year      = {2021},
  url       = {https://escholarship.org/uc/item/3r6807ct}
}

@inproceedings{huang2023ANPL,
author = {Huang, Di and Nan, Ziyuan and Hu, Xing and Jin, Pengwei and Peng, Shaohui and Wen, Yuanbo and Zhang, Rui and Du, Zidong and Guo, Qi and Pu, Yewen and Chen, Yunji},
title = {{ANPL: towards natural programming with interactive decomposition}},
year = {2023},
publisher = {Curran Associates Inc.},
address = {Red Hook, NY, USA},
booktitle = {Proceedings of the 37th International Conference on Neural Information Processing Systems},
articleno = {3040},
numpages = {37},
location = {New Orleans, LA, USA},
series = {NIPS '23}
}

@Article{LeGris2025,
author={LeGris, Solim
and Vong, Wai Keen
and Lake, Brenden M.
and Gureckis, Todd M.},
title={{A Comprehensive Behavioral Dataset for the Abstraction and Reasoning Corpus}},
journal={Scientific Data},
year={2025},
month={Aug},
day={07},
volume={12},
number={1},
pages={1380},
issn={2052-4463},
doi={10.1038/s41597-025-05687-1},
url={https://doi.org/10.1038/s41597-025-05687-1}
}

@inproceedings{
bikov2025augarc,
title={{Aug{ARC}: Augmented Abstraction and Reasoning Benchmark for Large Language Models}},
author={Kiril Bikov and Mikel Bober-Irizar and Soumya Banerjee},
booktitle={Workshop on Preparing Good Data for Generative AI: Challenges and Approaches},
year={2025},
url={https://openreview.net/forum?id=cgUTWzgvCj}
}

@misc{Moffitt2025,
  title={{ARC-GEN: A Mimetic Procedural Benchmark Generator for the Abstraction and Reasoning Corpus}}, 
  author={Michael D. Moffitt},
  year={2025},
  eprint={2511.00162},
  archivePrefix={arXiv},
  primaryClass={cs.AI},
  url={https://arxiv.org/abs/2511.00162}, 
}

@inproceedings{
li2025combining,
title={{Combining Induction and Transduction for Abstract Reasoning}},
author={Wen-Ding Li and Keya Hu and Carter Larsen and Yuqing Wu and Simon Alford and Caleb Woo and Spencer M. Dunn and Hao Tang and Wei-Long Zheng and Yewen Pu and Kevin Ellis},
booktitle={The Thirteenth International Conference on Learning Representations},
year={2025},
url={https://openreview.net/forum?id=UmdotAAVDe}
}

@inproceedings{Lehmann2026ARCTGI,
author = {Lehmann, Jens and Khushbakht, Syeda and Salehfard, Nikoo and Nishat, Nur A Zarin and Bhandiwad, Dhananjay and Aioanei, Andrei and Vahdati, Sahar},
title = {{ARC-TGI: Human-Validated Task Generators with Reasoning Chain Templates for ARC-AGI}},
year = {2026},
isbn = {9798400722592},
publisher = {Association for Computing Machinery},
address = {New York, NY, USA},
url = {https://doi.org/10.1145/3770855.3817596},
doi = {10.1145/3770855.3817596},
booktitle = {Proceedings of the 32nd ACM SIGKDD Conference on Knowledge Discovery and Data Mining V.2},
pages = {9290–9301},
numpages = {12},
location = {Republic of Korea},
series = {KDD '26}
}

@inproceedings{
mondorf2026compositionalarc,
title={{Compositional-{ARC}: Assessing Systematic Generalization in Abstract Spatial Reasoning}},
author={Philipp Mondorf and Shijia Zhou and Monica Riedler and Barbara Plank},
booktitle={The Fourteenth International Conference on Learning Representations},
year={2026},
url={https://openreview.net/forum?id=h497VpgFKd}
}

@inproceedings{
assouel2022objectcentric,
title={{Object-centric Compositional Imagination for Visual Abstract Reasoning}},
author={Rim Assouel and Pau Rodriguez and Perouz Taslakian and David Vazquez and Yoshua Bengio},
booktitle={ICLR2022 Workshop on the Elements of Reasoning: Objects, Structure and Causality},
year={2022},
url={https://openreview.net/forum?id=rCzfIruU5x5}
}

@article{team2026gemma,
  title={Gemma 4 technical report},
  author={Team, Gemma and Abd, Sherif El and Aggarwal, Vaibhav and Algayres, Robin and Andreev, Alek and Bachem, Olivier and Ballantyne, Ian and Brick, Cormac and C{\u{a}}rbune, Victor and Casbon, Michelle and others},
  journal={arXiv preprint arXiv:2607.02770},
  year={2026}
}

@misc{qwen3.5,
    title  = {{Qwen3.5}: Towards Native Multimodal Agents},
    author = {{Qwen Team}},
    month  = {February},
    year   = {2026},
    url    = {https://qwen.ai/blog?id=qwen3.5}
}

@misc{openai2025gptoss120bgptoss20bmodel,
      title={gpt-oss-120b \& gpt-oss-20b Model Card}, 
      author={OpenAI},
      year={2025},
      eprint={2508.10925},
      archivePrefix={arXiv},
      primaryClass={cs.CL},
      url={https://arxiv.org/abs/2508.10925}, 
}

@misc{qwen3technicalreport,
      title={Qwen3 Technical Report}, 
      author={Qwen Team},
      year={2025},
      eprint={2505.09388},
      archivePrefix={arXiv},
      primaryClass={cs.CL},
      url={https://arxiv.org/abs/2505.09388}, 
}

@article{abdin2025phi,
  title={Phi-4-reasoning technical report},
  author={Abdin, Marah and Agarwal, Sahaj and Awadallah, Ahmed and Balachandran, Vidhisha and Behl, Harkirat and Chen, Lingjiao and de Rosa, Gustavo and Gunasekar, Suriya and Javaheripi, Mojan and Joshi, Neel and others},
  journal={arXiv preprint arXiv:2504.21318},
  year={2025}
}

@article{camposampiero2025raven,
  title={I-raven-x: Benchmarking generalization and robustness of analogical and mathematical reasoning in large language and reasoning models},
  author={Camposampiero, Giacomo and Hersche, Michael and Wattenhofer, Roger and Sebastian, Abu and Rahimi, Abbas},
  journal={arXiv preprint arXiv:2510.17496},
  year={2025}
}

@article{kharlapenko2026fluid,
  title={Fluid Representations in Reasoning Models},
  author={Kharlapenko, Dmitrii and Stolfo, Alessandro and Conmy, Arthur and Sachan, Mrinmaya and Jin, Zhijing},
  journal={arXiv preprint arXiv:2602.04843},
  year={2026}
}

@misc{beger2026potarcin,
  title         = {{PotARCin}: Multi-Dimensional Evaluation of Skill Acquisition in Abstract Reasoning Tasks},
  author        = {Beger, Claas and Yi, Ryan and Mitchell, Melanie},
  year          = {2026},
  eprint        = {2609.27288v1},
  archivePrefix = {arXiv},
  primaryClass  = {cs.AI},
  doi           = {10.48550/arXiv.2609.27288},
  url           = {https://arxiv.org/abs/2609.27288v1}
}

@article{Xu_Khalil_Sanner_2023, title={Graphs, Constraints, and Search for the Abstraction and Reasoning Corpus}, volume={37}, url={https://ojs.aaai.org/index.php/AAAI/article/view/25527}, DOI={10.1609/aaai.v37i4.25527}, abstractNote={The Abstraction and Reasoning Corpus (ARC) aims at benchmarking the performance of general artificial intelligence algorithms. The ARC’s focus on broad generalization and few-shot learning has made it difficult to solve using pure machine learning. A more promising approach has been to perform program synthesis within an appropriately designed Domain Specific Language (DSL). However, these too have seen limited success. We propose Abstract Reasoning with Graph Abstractions (ARGA), a new object-centric framework that first represents images using graphs and then performs a search for a correct program in a DSL that is based on the abstracted graph space. The complexity of this combinatorial search is tamed through the use of constraint acquisition, state hashing, and Tabu search. An extensive set of experiments demonstrates the promise of ARGA in tackling some of the complicated object-centric tasks of the ARC rather efficiently, producing programs that are correct and easy to understand.}, number={4}, journal={Proceedings of the AAAI Conference on Artificial Intelligence}, author={Xu, Yudong and Khalil, Elias B. and Sanner, Scott}, year={2023}, month={Jun.}, pages={4115–4122} }

@InProceedings{10.1007/978-3-030-93409-5_54,
author="Alford, Simon
and Gandhi, Anshula
and Rangamani, Akshay
and Banburski, Andrzej
and Wang, Tony
and Dandekar, Sylee
and Chin, John
and Poggio, Tomaso
and Chin, Peter",
editor="Benito, Rosa Maria
and Cherifi, Chantal
and Cherifi, Hocine
and Moro, Esteban
and Rocha, Luis M.
and Sales-Pardo, Marta",
title="Neural-Guided, Bidirectional Program Search for Abstraction and Reasoning",
booktitle="Complex Networks {\&} Their Applications X",
year="2022",
publisher="Springer International Publishing",
address="Cham",
pages="657--668",
isbn="978-3-030-93409-5"
}

@article{legris2026synthesizing,
  title={Synthesizing symbolic models of human and machine reasoning at scale},
  author={LeGris, Solim and Mondorf, Philipp and Jin, Jinran and Jagadish, Akshay K and Lake, Brenden M and Gureckis, Todd M},
  journal={arXiv preprint arXiv:2611.07088},
  year={2026}
}

@inproceedings{NEURIPS2024_1c9c85ba,
 author = {Barke, Shraddha and Anaya Gonzalez, Emmanuel and Kasibatla, Saketh Ram and Berg-Kirkpatrick, Taylor and Polikarpova, Nadia},
 booktitle = {Advances in Neural Information Processing Systems},
 doi = {10.52202/079017-0499},
 editor = {A. Globerson and L. Mackey and D. Belgrave and A. Fan and U. Paquet and J. Tomczak and C. Zhang},
 pages = {15612--15645},
 publisher = {Curran Associates, Inc.},
 title = {HYSYNTH: Context-Free LLM Approximation for Guiding Program Synthesis},
 url = {https://proceedings.neurips.cc/paper_files/paper/2024/file/1c9c85bae6161d52182d0fe2f3640512-Paper-Conference.pdf},
 volume = {37},
 year = {2024}
}

@article{batorski2025nsa,
  title={Nsa: Neuro-symbolic arc challenge},
  author={Batorski, Pawe{\l} and Brinkmann, Jannik and Swoboda, Paul},
  journal={arXiv preprint arXiv:2501.04424},
  year={2025}
}
\bibliographystyle{iclr2027_conference}

\appendix
\section{Extended literature review}\label{sec:extended_lit_review}
\paragraph{Conceptual understanding in LRMs.} Evaluating conceptual understanding in LRMs requires distinguishing successful task execution from the formation of reusable abstractions. On symbolic Raven-style problems,~\citet{camposampiero2025raven} find that LRMs generalize better than conventional LLMs to longer reasoning relations and larger attribute ranges, but remain vulnerable to simulated perceptual uncertainty. Complementing such generalization tests,~\citet{beger2026potarcin} show that models can correctly formalize an ARC rule yet respond inconsistently when using it for classification, generation, or inversion. At the representational level, earlier work on LLMs identifies compact task vectors that summarize input--output demonstrations and guide predictions on new inputs~\citep{hendel-etal-2023-context}. Using representational similarity analysis,~\citet{opielka2025analogical} further identify format-invariant representations of some verbal concepts, which can be present even when the model produces an incorrect answer. More directly examining reasoning models,~\citet{kharlapenko2026fluid} find that QwQ-32B refines action and predicate representations during reasoning on planning problems with obfuscated terminology, developing encodings that align across different wordings. Their intervention experiments provide causal evidence that these refined representations contribute to successful problem solving. Our work complements these studies by examining representational organization across ARC-style tasks with distinct transformation rules. Specifically, we investigate whether this organization reflects the benchmark's spatial and semantic concepts or the transformation operations shared across tasks.

\section{A task generator for ConceptARC}\label{sec:task_generator_details}
In this section, we provide additional details about our compositional task generator introduced in Section~\ref{sec:data_generator}. Our DSL represents each task program as a structured specification containing a layout, a selector, and an action. Each component specifies an operation and its parameters, which are interpreted by shared execution functions. Components may contain nested expressions; for example, a selector can first identify a reference object and then select other objects relative to it. Some parameters are fixed when the task program is created, while others specify permitted variation across examples, such as ranges of object counts or dimensions. In addition to checking compatibility between components, concept-specific predicates inspect the entire program specification and its parameters to determine whether it expresses the intended concept. Acceptance checks also use trial task generations to assess whether candidates produce valid, non-trivial examples.

To encourage novelty, we exclude programs matching official ConceptARC task signatures and filter duplicate families using structural signatures supplemented by concept-specific equivalence checks. Program selection favors underrepresented components and rule types to promote diversity across the generated collection. Accepted programs are saved with their task identifier, concept, and component specifications. To generate examples, we sample input scenes under the program's constraints and apply its transformation rule to compute the outputs. Unsuitable scenes are rejected and resampled, and duplicate examples are filtered when assembling a task. We additionally impose program-dependent constraints on the support set, requiring variation in relevant properties such as object geometry, orientation, or count. Where specified, configurations appearing in test examples must also be demonstrated in the support set. These constraints provide evidence for the intended rule and reduce known ambiguities. The transformation in the task program can also be applied directly to an existing compatible input grid, as done when validating the official ConceptARC mappings.

As discussed in Section~\ref{subsec:conceptarc_100}, the concepts defined by the original ConceptARC benchmark are not necessarily semantically mutually exclusive. For example, a task that counts objects inside a container may involve both ``\emph{Inside and Outside}'' and ``\emph{Count}.'' Such overlap is expected, as a task's assigned concept describes the primary abstraction it is designed to test, rather than all semantic properties involved in its solution. To identify duplicate tasks across concepts, we first verify that no two tasks share an identical support set. However, different support sets may still represent the same transformation rule. We therefore compare task programs across concepts by applying each program to the other task's input grids. We consider two tasks duplicates if both programs reproduce each other's corresponding outputs on all tested inputs. Importantly, any disagreement on a single input establishes that the two programs are not functionally equivalent; thus, this procedure may conservatively identify distinct programs as duplicates, but cannot fail to identify functionally identical programs. This procedure identifies only five pairs of duplicate tasks across concepts (10 out of 1,600 tasks, or $0.625\%$), which we subsequently replace.

\begin{table*}[t]
\centering
\footnotesize
\caption{Decoding parameters and token budgets for \textsc{ConceptARC-100}. A top-$k$ value of $0$ disables top-$k$ filtering.}
\label{tab:decoding_parameters}
\begin{tabular}{lrrrrrr}
\toprule
Model
& Temperature
& Top-$p$
& Top-$k$
& \shortstack{Presence\\penalty}
& \shortstack{Total output\\budget}
& \shortstack{Reasoning\\budget} \\
\midrule
Gemma4-31B
& 1.0 & 0.95 & 64 & 0.0 & 65,536 & 49,152 \\
Qwen3.5-27B / Qwen3.5-9B
& 1.0 & 0.95 & 20 & 1.5 & 65,536 & 49,152 \\
GPT-OSS-120B / GPT-OSS-20B
& 1.0 & 1.00 & 0 & 0.0 & 65,536 & 49,152 \\
Qwen3-32B / Qwen3-8B
& 0.6 & 0.95 & 20 & 1.5 & 32,768 & 20,480 \\
Phi4-Reasoning-Plus
& 0.8 & 0.95 & 50 & 0.0 & 18,432 & 15,360 \\
\bottomrule
\end{tabular}
\end{table*}

\paragraph{Manual task review.} To ensure that generated tasks are valid and of high quality, we perform multiple rounds of manual review throughout the development and generation process. During DSL implementation, we regularly inspect generated tasks to verify that they represent valid transformation rules, are unambiguously inferable from the three support examples, and faithfully capture their intended concepts. These reviews informed iterative refinements of the DSL and the addition of sanity checks for automatic task validation, as described above.

After the generator consistently produced valid tasks, we generated \textsc{ConceptARC-100} with 100 tasks per concept, as described in Section~\ref{subsec:conceptarc_100}. During the generation process, we manually reviewed every generated task and adjusted the DSL when necessary. We then conducted an independent validation round with a research assistant, who evaluated each task through an interface displaying the three input--output support examples and ten test examples. A task was considered valid if and only if (i) the transformation rule was applied consistently across examples and (ii) the rule could be inferred unambiguously from the support set. Invalid tasks were annotated with a textual explanation and subsequently corrected. This evaluation identified 6.2\% of tasks as invalid, all of which were correct before further processing. To strengthen benchmark validity, we conducted a second, independent post-correction audit of all 1,600 generated tasks in ConceptARC-100. In this audit, 1,596 of the 1,600 tasks (99.75\%) were judged valid. For tasks initially judged valid and therefore left unchanged after the first audit, the two independent annotators agreed in 98.03\% of cases. We subsequently revise the four tasks (0.25\%) judged invalid in the second audit.

\begin{figure}[tbp]
\centering
\includegraphics[width=\linewidth]{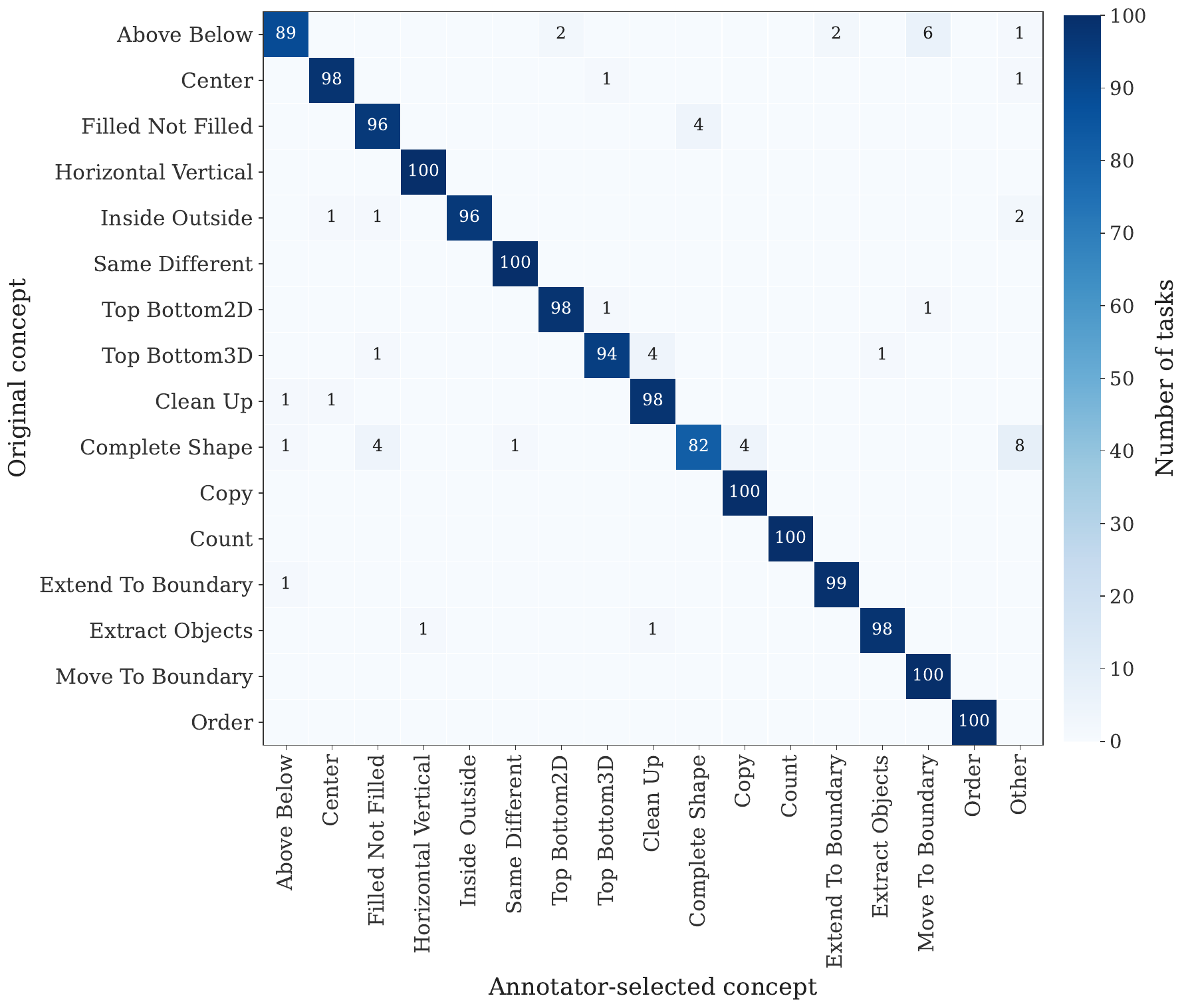}
\caption{\textbf{Confusion matrix of concept annotations.} Rows represent the intended concept labels, while columns show the labels assigned by an independent annotator without knowledge of the intended concepts. Each cell indicates the number of tasks, with 100 tasks per concept. The ``\emph{Other}'' category indicates that none of the 16 concepts was considered appropriate. The annotator identifies the intended concept in $96.75\%$ of all tasks.}
\label{fig:annotator_concept_confusion}
\end{figure}

To assess whether tasks clearly express their intended concepts, we additionally ask the annotator to assign the most appropriate label among the 16 ConceptARC concepts, without revealing the intended concept label. The annotator could also select ``\emph{Other}'' if none of the concepts clearly applied. The annotator identifies the intended concept in $96.75\%$ of cases ($96.38\%$ for predicate concepts and $97.13\%$ for action concepts), suggesting that the intended concepts are generally recognizable despite potential semantic overlap. Figure~\ref{fig:annotator_concept_confusion} shows the corresponding confusion matrix. Most disagreements involve ``\emph{Complete Shape},'' for which ``\emph{Other}'' is the most frequent alternative label ($8\%$ of tasks within this concept).

\section{Experimental Details}\label{sec:experimental_details}
In this section, we provide further details about our experiment setup. For instance, we report information about parameters when evaluating LRMs in Section~\ref{subsec:eval_lrms} and details regarding training models from scratch in Section~\ref{subsec:training_details}. 

\subsection{Evaluating LRMs}\label{subsec:eval_lrms}
We run inference using vLLM, generating one response per test query and evaluated seed. We enable thinking for Gemma4 and the Qwen models and set GPT-OSS reasoning effort to \texttt{high}. For Phi4-Reasoning-Plus, whose native chat template supplies its own system message, we preserve that message and prepend the ConceptARC system instructions to the user message, separated by a blank line. Table~\ref{tab:decoding_parameters} lists the decoding parameters and token budgets. All models use $\texttt{min\_p}=0$ and a repetition penalty of $1.0$. The total output budget includes all generated tokens, including reasoning, transition markers, and the final answer, but excludes prompt tokens. We additionally impose a separate reasoning-token budget. If this budget is exhausted before reasoning ends, decoding forces the model-specific closing sequence: \texttt{</think>} with surrounding newlines for Qwen and Phi4, and \texttt{<channel|>} for Gemma4. For GPT-OSS, a custom logits processor forces a complete Harmony transition from the \texttt{analysis} channel to the \texttt{final} channel. Generation then continues within the remaining total output budget. We use smaller output and reasoning budgets for Qwen3 and Phi4-Reasoning-Plus to accommodate their shorter context windows, which must contain both the task prompt and the generated response. For every model, we set the reasoning budget below the total output budget to leave room for final-answer generation. Responses that terminate without a normal stop, including those truncated by the output limit, and responses without a parseable grid are counted as incorrect. Otherwise, correctness requires an exact match to the target grid, as described in Section~\ref{sec:experiments}. Table~\ref{tab:conceptarc_performance_stdevs} provides an overview of the models' performance across the different concepts. Values represent means across three random seeds, along with their standard deviations.

\begin{table*}[t]
\centering
\footnotesize
\setlength{\tabcolsep}{5pt}
\renewcommand{\arraystretch}{1.15}
\caption{\textbf{Accuracy (\%) on \textsc{ConceptARC-100} (1,600 tasks and 16,000 test queries per seed).} Results are reported as mean$_{\pm\mathrm{SD}}$ across three random seeds. Sample standard deviations are computed across seeds after any concept averaging. GPT-OSS models use high reasoning effort. The highest mean in each row is bold.}
\label{tab:conceptarc_performance_stdevs}
\resizebox{\textwidth}{!}{%
\begin{tabular}{l*{8}{r}:*{2}{r}}
\toprule
Concept
& \shortstack{Gemma4-31B}
& \shortstack{Qwen3.5-27B}
& \shortstack{Qwen3.5-9B}
& \shortstack{GPT-OSS-120B}
& \shortstack{GPT-OSS-20B}
& \shortstack{Qwen3-32B}
& \shortstack{Qwen3-8B}
& \shortstack{Phi4-Reas.+}
& VARC
& TRM\\
\midrule
\predconcept{AboveBelow}
& $\mathbf{75.30}_{\pm 0.70}$ & $73.40_{\pm 0.26}$ & $43.90_{\pm 1.04}$ & $56.97_{\pm 0.59}$ & $37.17_{\pm 0.25}$ & $13.97_{\pm 0.35}$ & $6.90_{\pm 0.26}$ & $13.17_{\pm 0.67}$ & $37.07_{\pm 0.79}$ & $14.63_{\pm 0.87}$ \\
\predconcept{Center}
& $80.03_{\pm 1.04}$ & $\mathbf{83.27}_{\pm 0.51}$ & $61.70_{\pm 1.40}$ & $79.93_{\pm 0.59}$ & $60.77_{\pm 0.83}$ & $24.80_{\pm 1.45}$ & $5.77_{\pm 1.04}$ & $19.17_{\pm 0.90}$ & $16.73_{\pm 0.31}$ & $3.97_{\pm 0.45}$ \\
\predconcept{FilledNotFilled}
& $60.53_{\pm 1.70}$ & $\mathbf{71.13}_{\pm 1.01}$ & $32.43_{\pm 0.49}$ & $48.20_{\pm 0.10}$ & $23.37_{\pm 0.81}$ & $9.83_{\pm 0.15}$ & $2.40_{\pm 0.36}$ & $4.33_{\pm 0.51}$ & $16.90_{\pm 0.80}$ & $7.27_{\pm 0.66}$ \\
\predconcept{HorizontalVertical}
& $64.23_{\pm 1.12}$ & $\mathbf{73.17}_{\pm 0.45}$ & $34.87_{\pm 0.25}$ & $47.83_{\pm 1.20}$ & $27.50_{\pm 0.70}$ & $11.77_{\pm 0.64}$ & $5.20_{\pm 0.60}$ & $7.10_{\pm 0.95}$ & $9.33_{\pm 0.68}$ & $10.57_{\pm 0.19}$ \\
\predconcept{InsideOutside}
& $\mathbf{61.30}_{\pm 1.13}$ & $58.03_{\pm 0.87}$ & $31.07_{\pm 1.25}$ & $33.77_{\pm 0.21}$ & $26.37_{\pm 0.23}$ & $9.17_{\pm 0.64}$ & $2.80_{\pm 0.56}$ & $3.93_{\pm 0.25}$ & $38.53_{\pm 2.22}$ & $13.60_{\pm 0.54}$ \\
\predconcept{SameDifferent}
& $\mathbf{44.10}_{\pm 1.32}$ & $36.27_{\pm 0.58}$ & $10.97_{\pm 0.60}$ & $17.70_{\pm 0.69}$ & $10.17_{\pm 0.50}$ & $5.23_{\pm 0.21}$ & $3.27_{\pm 0.06}$ & $2.70_{\pm 0.46}$ & $12.60_{\pm 0.59}$ & $7.13_{\pm 0.78}$ \\
\predconcept{TopBottom2D}
& $\mathbf{73.63}_{\pm 1.10}$ & $66.23_{\pm 0.35}$ & $23.77_{\pm 1.59}$ & $35.60_{\pm 1.51}$ & $17.60_{\pm 0.80}$ & $17.37_{\pm 1.18}$ & $5.20_{\pm 0.26}$ & $10.00_{\pm 0.56}$ & $25.43_{\pm 0.54}$ & $15.27_{\pm 0.66}$ \\
\predconcept{TopBottom3D}
& $55.40_{\pm 1.40}$ & $\mathbf{66.83}_{\pm 1.07}$ & $33.63_{\pm 0.81}$ & $52.73_{\pm 1.72}$ & $31.67_{\pm 0.74}$ & $9.83_{\pm 0.83}$ & $4.10_{\pm 0.46}$ & $7.30_{\pm 0.26}$ & $9.53_{\pm 0.17}$ & $4.53_{\pm 0.33}$ \\
\actconcept{CleanUp}
& $49.83_{\pm 1.44}$ & $\mathbf{59.43}_{\pm 1.81}$ & $28.37_{\pm 1.36}$ & $37.30_{\pm 0.53}$ & $22.57_{\pm 1.69}$ & $5.63_{\pm 0.06}$ & $1.80_{\pm 0.26}$ & $4.90_{\pm 0.56}$ & $13.37_{\pm 1.02}$ & $4.77_{\pm 0.52}$ \\
\actconcept{CompleteShape}
& $70.77_{\pm 1.72}$ & $\mathbf{77.03}_{\pm 1.19}$ & $44.77_{\pm 2.18}$ & $60.80_{\pm 0.46}$ & $37.33_{\pm 1.12}$ & $6.17_{\pm 0.58}$ & $1.90_{\pm 0.10}$ & $5.40_{\pm 0.61}$ & $36.13_{\pm 1.22}$ & $10.43_{\pm 2.58}$ \\
\actconcept{Copy}
& $50.40_{\pm 0.40}$ & $\mathbf{58.37}_{\pm 2.25}$ & $27.83_{\pm 0.93}$ & $43.23_{\pm 1.19}$ & $23.07_{\pm 1.74}$ & $6.53_{\pm 0.40}$ & $1.67_{\pm 0.15}$ & $3.10_{\pm 0.70}$ & $9.57_{\pm 0.54}$ & $4.57_{\pm 0.21}$ \\
\actconcept{Count}
& $27.53_{\pm 0.83}$ & $\mathbf{33.30}_{\pm 0.40}$ & $16.00_{\pm 0.44}$ & $16.90_{\pm 1.15}$ & $9.80_{\pm 1.11}$ & $4.33_{\pm 0.85}$ & $2.53_{\pm 0.45}$ & $2.97_{\pm 0.25}$ & $5.27_{\pm 0.41}$ & $5.93_{\pm 0.94}$ \\
\actconcept{ExtendToBoundary}
& $76.87_{\pm 0.51}$ & $\mathbf{80.73}_{\pm 0.72}$ & $51.00_{\pm 0.53}$ & $69.57_{\pm 1.07}$ & $46.90_{\pm 0.10}$ & $17.27_{\pm 0.51}$ & $7.47_{\pm 0.76}$ & $14.27_{\pm 0.75}$ & $32.97_{\pm 0.12}$ & $10.70_{\pm 1.48}$ \\
\actconcept{ExtractObjects}
& $53.13_{\pm 1.88}$ & $\mathbf{69.70}_{\pm 0.85}$ & $35.93_{\pm 0.74}$ & $53.80_{\pm 1.71}$ & $30.67_{\pm 0.55}$ & $9.20_{\pm 0.61}$ & $4.20_{\pm 0.72}$ & $4.57_{\pm 0.15}$ & $6.97_{\pm 0.69}$ & $2.80_{\pm 0.50}$ \\
\actconcept{MoveToBoundary}
& $56.60_{\pm 0.70}$ & $\mathbf{68.50}_{\pm 0.70}$ & $40.87_{\pm 1.02}$ & $56.17_{\pm 1.12}$ & $38.33_{\pm 0.64}$ & $15.87_{\pm 1.10}$ & $9.23_{\pm 0.83}$ & $9.87_{\pm 0.38}$ & $10.93_{\pm 0.05}$ & $2.70_{\pm 0.29}$ \\
\actconcept{Order}
& $28.30_{\pm 0.61}$ & $\mathbf{40.27}_{\pm 0.93}$ & $9.07_{\pm 0.00}$ & $16.13_{\pm 0.59}$ & $5.87_{\pm 0.45}$ & $1.07_{\pm 0.25}$ & $0.17_{\pm 0.15}$ & $0.17_{\pm 0.12}$ & $0.10_{\pm 0.08}$ & $0.03_{\pm 0.05}$ \\
\midrule
\predconcept{\textbf{Avg. Predicate Concepts}}
& $64.32_{\pm 0.05}$ & $\mathbf{66.04}_{\pm 0.16}$ & $34.04_{\pm 0.36}$ & $46.59_{\pm 0.18}$ & $29.33_{\pm 0.41}$ & $12.75_{\pm 0.31}$ & $4.45_{\pm 0.24}$ & $8.46_{\pm 0.20}$ & $20.77_{\pm 0.66}$ & $9.62_{\pm 0.26}$ \\
\actconcept{\textbf{Avg. Action Concepts}}
& $51.68_{\pm 0.38}$ & $\mathbf{60.92}_{\pm 0.39}$ & $31.73_{\pm 0.35}$ & $44.24_{\pm 0.23}$ & $26.82_{\pm 0.26}$ & $8.26_{\pm 0.06}$ & $3.62_{\pm 0.15}$ & $5.65_{\pm 0.13}$ & $14.41_{\pm 0.28}$ & $5.24_{\pm 0.41}$ \\
\midrule
\textbf{Average}
& $58.00_{\pm 0.20}$ & $\mathbf{63.48}_{\pm 0.11}$ & $32.89_{\pm 0.22}$ & $45.41_{\pm 0.03}$ & $28.07_{\pm 0.28}$ & $10.50_{\pm 0.18}$ & $4.04_{\pm 0.16}$ & $7.06_{\pm 0.14}$ & $17.59_{\pm 0.46}$ & $7.43_{\pm 0.14}$ \\
\bottomrule
\end{tabular}%
}
\end{table*}

\subsection{Training details}\label{subsec:training_details}
In order to benchmark ConceptARC-100 with models trained from scratch, in complement to pretrained LRMs, we train a VARC-ViT-18M model, following~\citet{hu2025arcvisionproblem}, and a Tiny Recursive Model-7M following~\citet{jolicoeurmartineau2025morerecursivereasoningtiny}, respectively named VARC and TRM hereafter. We choose these models because, among models already tested on ARC-AGI-1 and trainable from scratch, they have a reusable code that can run with reasonable academic resources, and they serve as foundations for two distinct emergent lines of research on abstract visual reasoning. VARC supports the vision-inspired neural models for ARC-AGI-1, e.g. with Embed-TTT~\citep{deliege2026implicitruleinduction} and LoopViT~\citep{shu2026loopvit} as recent follow-ups building upon it. TRM, which simplifies its predecessor HRM~\citep{wang2025hrm}, supports the recursive approaches that leverage the compounding effect of nested loops to train models with a large effective depth without suffering of vanishing gradients, and has FPRM~\citep{movahedi2026fixedpointreasonersstableadaptive} and GRAM~\citep{baek2026generativerecursivereasoning} as follow-ups.

For training, VARC and TRM take in an input grid, where each cell is tokenized with a 512 dimensional vector according to its color index, VARC randomly scales it or not by a factor 2, and both finally place the resulting grid randomly on a larger background canvas with special boundary tokens, of size 64x64 for VARC and 30x30 for TRM. Then, for VARC, following the common ViT practice~\citep{dosovitskiy2021animageisworth}, the canvas is split into non-overlapping 2 $\times$ 2 patches, projected in the 512 dimensional latent space by a learnable linear transform, and flattened as a sequence of 1024 tokens. TRM directly flattens its canvas into a sequence of 900 tokens. A learnable task-specific token, which is shared across all inputs of a same task, is appended to each sequence, with TRM also appending 15 additional ``scratchpad'' slots (usable by the model during its computations but not containing any learnable parameter). Given that input grids can be identical in different tasks, the role of the task embeddings is to encode the the rule used to map the inputs to the outputs of the tasks. Then the 1025 or 916 token-long sequence is fed to the model backbone, made of 10 transformer layers for VARC, and of 2 transformer layers reused $N\times 21$ times for TRM, with $1\leq N \leq 16$ depending on a halting mechanism. Finally, the task token and extra slots are discarded, the token list is reshaped and upscaled back to the canvas size. The prediction head projects each cell token to a logits vector across the various color, background, and border classes. The part of the grid contained within the predicted boundary tokens is compared to the ground truth grid through a cross-entropy loss, allowing backpropagation all along.

\section{Additional Results}\label{sec:additional_results}

\begin{table*}[t]
\centering
\footnotesize
\setlength{\tabcolsep}{5pt}
\renewcommand{\arraystretch}{1.15}
\caption{\textbf{Original ConceptARC accuracy (\%; 160 tasks and 480 test queries per seed), reported as mean$_{\pm\mathrm{SD}}$ across seeds.} Predicate and action averages weight their respective eight concepts equally; the final row averages all 16 concepts equally. Sample standard deviations are computed across seeds after any concept averaging. GPT-OSS models use high reasoning effort. The highest mean in each row is bold.}
\label{tab:conceptarc_original_performance}
\resizebox{\textwidth}{!}{%
\begin{tabular}{l*{8}{r}}
\toprule
Concept
& \shortstack{Gemma4-31B}
& \shortstack{Qwen3.5-27B}
& \shortstack{Qwen3.5-9B}
& \shortstack{GPT-OSS-120B}
& \shortstack{GPT-OSS-20B}
& \shortstack{Qwen3-32B}
& \shortstack{Qwen3-8B}
& \shortstack{Phi4-Reas.+} \\
\midrule
\predconcept{AboveBelow}
& $91.11_{\pm 3.85}$ & $\mathbf{93.33}_{\pm 5.77}$ & $74.44_{\pm 1.92}$ & $86.67_{\pm 3.33}$ & $72.22_{\pm 1.92}$ & $31.11_{\pm 11.71}$ & $14.44_{\pm 1.92}$ & $27.78_{\pm 5.09}$ \\
\predconcept{Center}
& $95.56_{\pm 1.92}$ & $\mathbf{97.78}_{\pm 3.85}$ & $87.78_{\pm 1.92}$ & $90.00_{\pm 5.77}$ & $82.22_{\pm 1.92}$ & $40.00_{\pm 3.33}$ & $24.44_{\pm 3.85}$ & $36.67_{\pm 3.33}$ \\
\predconcept{FilledNotFilled}
& $78.89_{\pm 5.09}$ & $\mathbf{93.33}_{\pm 0.00}$ & $75.56_{\pm 5.09}$ & $83.33_{\pm 3.33}$ & $66.67_{\pm 0.00}$ & $27.78_{\pm 1.92}$ & $20.00_{\pm 6.67}$ & $18.89_{\pm 5.09}$ \\
\predconcept{HorizontalVertical}
& $78.89_{\pm 6.94}$ & $\mathbf{81.11}_{\pm 6.94}$ & $64.44_{\pm 10.18}$ & $75.56_{\pm 7.70}$ & $68.89_{\pm 5.09}$ & $27.78_{\pm 1.92}$ & $11.11_{\pm 5.09}$ & $20.00_{\pm 3.33}$ \\
\predconcept{InsideOutside}
& $\mathbf{84.44}_{\pm 1.92}$ & $80.00_{\pm 3.33}$ & $70.00_{\pm 6.67}$ & $60.00_{\pm 14.53}$ & $54.44_{\pm 1.92}$ & $18.89_{\pm 1.92}$ & $6.67_{\pm 0.00}$ & $8.89_{\pm 5.09}$ \\
\predconcept{SameDifferent}
& $84.44_{\pm 1.92}$ & $\mathbf{95.56}_{\pm 1.92}$ & $73.33_{\pm 3.33}$ & $75.56_{\pm 3.85}$ & $58.89_{\pm 1.92}$ & $23.33_{\pm 3.33}$ & $13.33_{\pm 6.67}$ & $13.33_{\pm 6.67}$ \\
\predconcept{TopBottom2D}
& $91.11_{\pm 6.94}$ & $\mathbf{94.44}_{\pm 3.85}$ & $86.67_{\pm 8.82}$ & $82.22_{\pm 5.09}$ & $76.67_{\pm 3.33}$ & $34.44_{\pm 10.72}$ & $15.56_{\pm 3.85}$ & $62.22_{\pm 8.39}$ \\
\predconcept{TopBottom3D}
& $57.78_{\pm 1.92}$ & $\mathbf{78.89}_{\pm 5.09}$ & $54.44_{\pm 8.39}$ & $61.11_{\pm 5.09}$ & $52.22_{\pm 10.18}$ & $16.67_{\pm 3.33}$ & $6.67_{\pm 3.33}$ & $13.33_{\pm 3.33}$ \\
\actconcept{CleanUp}
& $64.44_{\pm 1.92}$ & $\mathbf{74.44}_{\pm 6.94}$ & $64.44_{\pm 3.85}$ & $50.00_{\pm 3.33}$ & $37.78_{\pm 5.09}$ & $24.44_{\pm 3.85}$ & $21.11_{\pm 6.94}$ & $20.00_{\pm 6.67}$ \\
\actconcept{CompleteShape}
& $70.00_{\pm 0.00}$ & $\mathbf{78.89}_{\pm 3.85}$ & $62.22_{\pm 3.85}$ & $74.44_{\pm 8.39}$ & $58.89_{\pm 3.85}$ & $25.56_{\pm 1.92}$ & $10.00_{\pm 3.33}$ & $32.22_{\pm 3.85}$ \\
\actconcept{Copy}
& $83.33_{\pm 0.00}$ & $\mathbf{85.56}_{\pm 3.85}$ & $64.44_{\pm 6.94}$ & $74.44_{\pm 5.09}$ & $50.00_{\pm 3.33}$ & $22.22_{\pm 5.09}$ & $17.78_{\pm 1.92}$ & $18.89_{\pm 1.92}$ \\
\actconcept{Count}
& $\mathbf{92.22}_{\pm 3.85}$ & $91.11_{\pm 6.94}$ & $88.89_{\pm 5.09}$ & $86.67_{\pm 3.33}$ & $85.56_{\pm 1.92}$ & $50.00_{\pm 3.33}$ & $20.00_{\pm 5.77}$ & $44.44_{\pm 6.94}$ \\
\actconcept{ExtendToBoundary}
& $93.33_{\pm 3.33}$ & $\mathbf{94.44}_{\pm 1.92}$ & $76.67_{\pm 3.33}$ & $81.11_{\pm 6.94}$ & $72.22_{\pm 10.18}$ & $23.33_{\pm 3.33}$ & $10.00_{\pm 0.00}$ & $10.00_{\pm 6.67}$ \\
\actconcept{ExtractObjects}
& $75.56_{\pm 8.39}$ & $\mathbf{83.33}_{\pm 3.33}$ & $71.11_{\pm 3.85}$ & $76.67_{\pm 5.77}$ & $58.89_{\pm 10.18}$ & $4.44_{\pm 1.92}$ & $2.22_{\pm 1.92}$ & $8.89_{\pm 1.92}$ \\
\actconcept{MoveToBoundary}
& $66.67_{\pm 0.00}$ & $\mathbf{91.11}_{\pm 1.92}$ & $67.78_{\pm 13.47}$ & $86.67_{\pm 3.33}$ & $60.00_{\pm 11.55}$ & $21.11_{\pm 7.70}$ & $13.33_{\pm 3.33}$ & $11.11_{\pm 1.92}$ \\
\actconcept{Order}
& $56.67_{\pm 3.33}$ & $\mathbf{73.33}_{\pm 5.77}$ & $46.67_{\pm 3.33}$ & $53.33_{\pm 8.82}$ & $36.67_{\pm 6.67}$ & $17.78_{\pm 1.92}$ & $15.56_{\pm 1.92}$ & $14.44_{\pm 1.92}$ \\
\midrule
\predconcept{\textbf{Avg. Predicate Concepts}}
& $82.78_{\pm 1.27}$ & $\mathbf{89.31}_{\pm 1.68}$ & $73.33_{\pm 2.89}$ & $76.81_{\pm 3.87}$ & $66.53_{\pm 0.87}$ & $27.50_{\pm 2.32}$ & $14.03_{\pm 2.51}$ & $25.14_{\pm 1.46}$ \\
\actconcept{\textbf{Avg. Action Concepts}}
& $75.28_{\pm 1.27}$ & $\mathbf{84.03}_{\pm 0.24}$ & $67.78_{\pm 2.29}$ & $72.92_{\pm 2.60}$ & $57.50_{\pm 0.72}$ & $23.61_{\pm 0.48}$ & $13.75_{\pm 0.72}$ & $20.00_{\pm 3.33}$ \\
\midrule
\textbf{Average}
& $79.03_{\pm 0.64}$ & $\mathbf{86.67}_{\pm 0.95}$ & $70.56_{\pm 2.59}$ & $74.86_{\pm 3.13}$ & $62.01_{\pm 0.64}$ & $25.56_{\pm 1.39}$ & $13.89_{\pm 1.34}$ & $22.57_{\pm 1.39}$ \\
\bottomrule
\end{tabular}%
}
\end{table*}

\subsection{Model performance on ConceptARC}\label{subsec:original_conceptarc_performance}
Table~\ref{tab:conceptarc_original_performance} presents an overview of LRM performance on the original ConceptARC~\citep{moskvichev2023the} benchmark. Values denote averages over the three random seeds used for evaluation, together with the corresponding standard deviations. We find that overall model performance is notably higher on the original ConceptARC benchmark than on our new \textsc{ConceptARC-100} dataset. This is surprising because, based on our manual review of the tasks, as described in Section~\ref{sec:task_generator_details}, we judge the tasks in \textsc{ConceptARC-100} to be easier or at least not more difficult than the original ConceptARC tasks. This might further support the hypothesis that some recent LRMs were trained on, or otherwise exposed to, the original ConceptARC benchmark.

When examining the relative ordering of model performance, we observe trends similar to those found in our evaluation on \textsc{ConceptARC-100}: more recent models such as Gemma4-31B ($79.03\%$ accuracy) and Qwen3.5-27B ($86.67\%$ accuracy) outperform older LRMs such as Qwen3-32B ($25.56\%$ accuracy) by a wide margin. Within model families, model size again appears to be an influential factor, with larger models generally outperforming their smaller counterparts. For instance, Qwen3.5-27B, which achieves $86.67\%$ accuracy, outperforms Qwen3.5-9B, which achieves $70.56\%$ accuracy.

We also find that models perform well on relational concepts, such as ``\emph{Above and Below}'' and ``\emph{Top / Bottom (2d)}'', in the original ConceptARC benchmark. For instance, Qwen3.5-27B achieves $93.33\%$ accuracy on tasks from ``\emph{Above and Below}'' and $94.44\%$ accuracy on tasks from ``\emph{Top / Bottom (2D)}''. We further find that models perform comparatively less well on concepts requiring comparisons between objects. Qwen3.5-27B, for example, achieves only $73.33\%$ accuracy on tasks from ``\emph{Order}''.

Overall, we find that models tend to perform better on predicate concepts than on action concepts. One difference compared to our results on \textsc{ConceptARC-100} is that most models appear to perform well on tasks involving the concept ``\emph{Count}''. This could stem from differences in how counting is represented in the two datasets, as described in Section~\ref{subsec:performance}: whereas ConceptARC typically represents counting bars in an isolated manner, our dataset encodes counts as colored bars or stripes overlaid on the input grid, which might be a less intuitive and less common.

\subsection{Analyzing how LRMs represent tasks internally}\label{subsec:lrm_task_representation}
In Section~\ref{subsec:repr_analysis}, we analyze how models represent tasks internally and ask whether they are organized according to semantically meaningful properties. Figure~\ref{fig:gemma4_31b_separation_probability} shows the separation probability of Gemma4-31B by \emph{layout}, \emph{selector}, \emph{action}, and \emph{concept}, as described in Section~\ref{subsec:repr_analysis}. Similar to Qwen3.5-27B, we find that Gemma4-31B organizes tasks primarily by grid layout in early layers. This organization changes as information propagates through the model:
\begin{wrapfigure}{r}{0.45\columnwidth}
    \centering
    \vspace{-0.1em}
    \includegraphics[width=\linewidth]{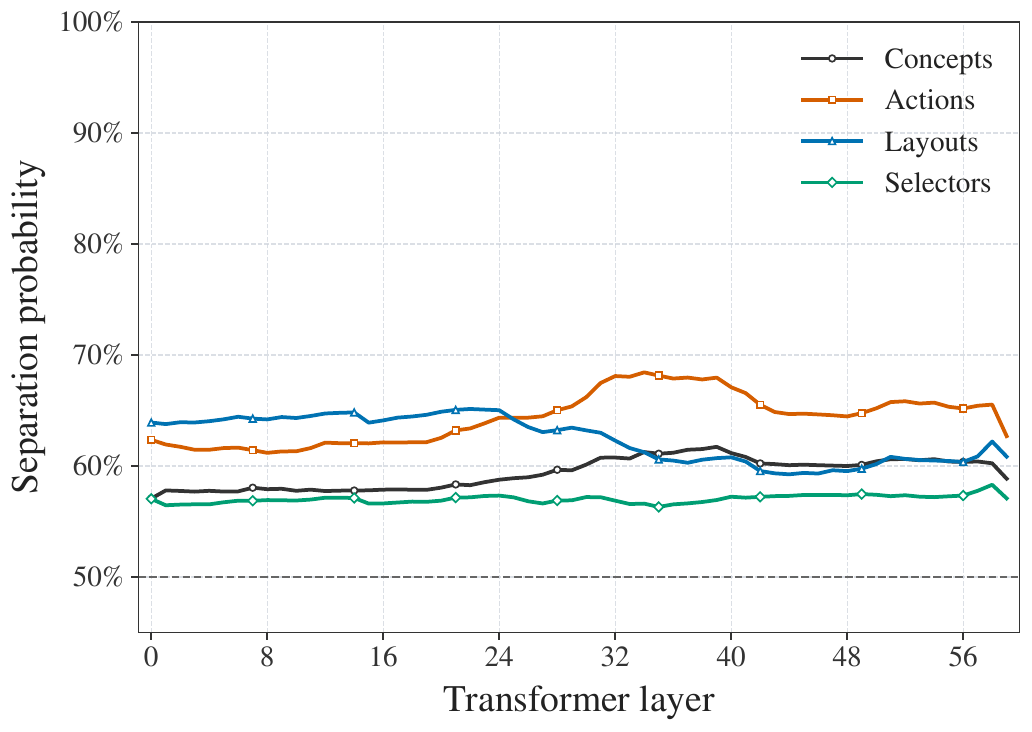}
    \caption{\textbf{Gemma4-31B representations also shift from layout- to action-based organization across layers.} Early layers primarily organize tasks by grid layout, while action-based organization becomes stronger in later layers, suggesting a shift from spatial properties toward the transformation operations required to solve each task.}
    \label{fig:gemma4_31b_separation_probability}
    \vspace{-1.2em}
\end{wrapfigure}
separation by layout decreases, while separation by action increases from $62.35\%$ at layer 0 to a peak of $68.43\%$ at layer 34. 
Similar to Qwen3.5-27B, selectors appear to have comparatively little influence in how tasks are grouped in the model's representation space.

In Section~\ref{subsec:repr_analysis}, we analyze how models represent tasks internally and ask whether these representations are organized according to semantically meaningful properties. Figure~\ref{fig:gemma4_31b_separation_probability} shows the separation probability of Gemma4-31B by \emph{layout}, \emph{selector}, \emph{action}, and \emph{concept}, as described in Section~\ref{subsec:repr_analysis}.
Similar to Qwen3.5-27B, we find that Gemma4-31B organizes tasks primarily according to grid layout in early layers. This organization changes as information propagates through the model: separation by layout decreases, while separation by action increases from $62.35\%$ at layer 0 to a peak of $68.43\%$ at layer 34. Similar to Qwen3.5-27B, selectors appear to have comparatively little influence on how tasks are grouped in the model's representation space.

\begin{figure}[tpb]
    \centering
    \begin{subfigure}[t]{0.44\linewidth}
        \centering
        \includegraphics[width=\linewidth]{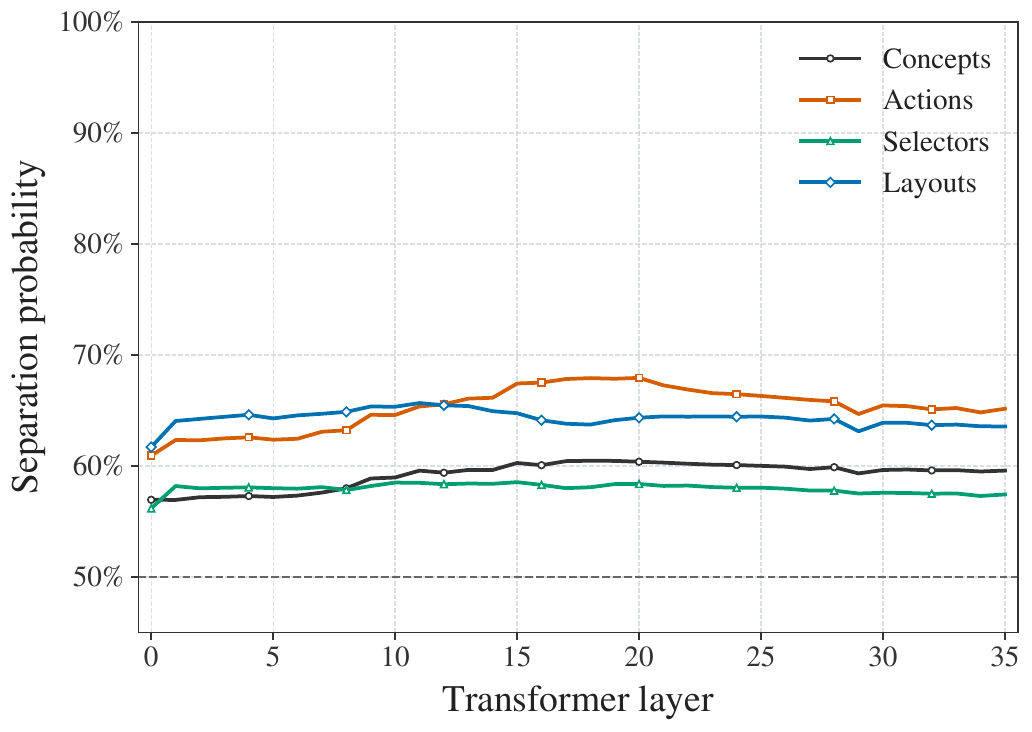}
        \caption{GPT-OSS-120B}
        \label{fig:gpt_oss_120b_separation_probability}
    \end{subfigure}
    \hfill
    \begin{subfigure}[t]{0.44\linewidth}
        \centering
        \includegraphics[width=\linewidth]{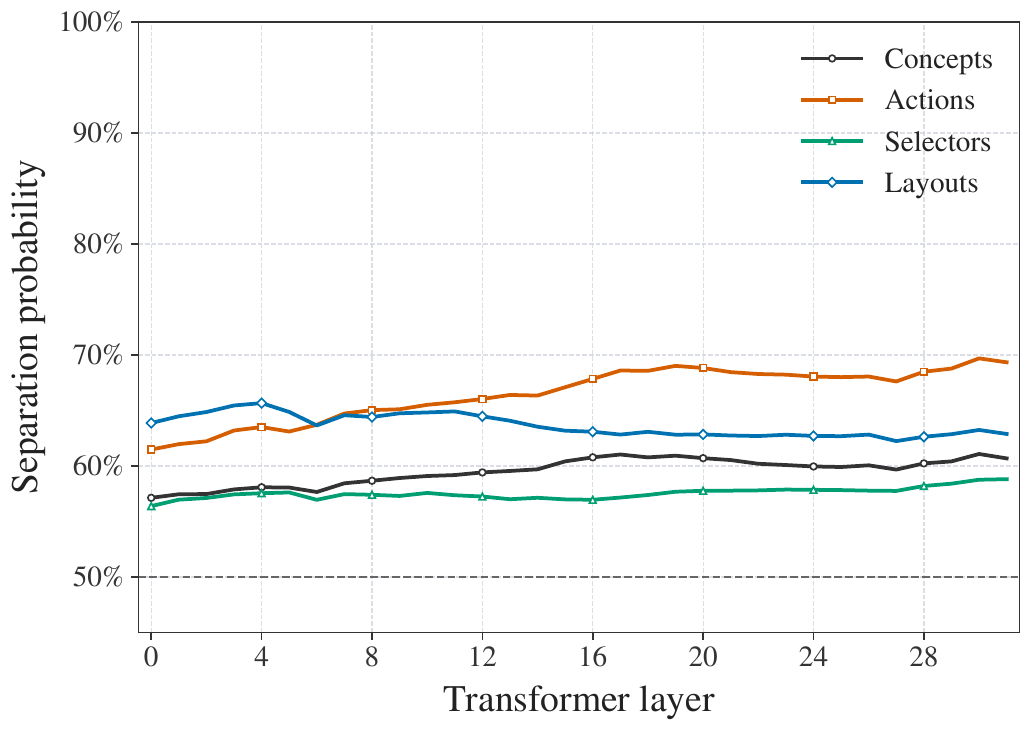}
        \caption{Qwen3.5-9B}
        \label{fig:qwen35_9b_separation_probability}
    \end{subfigure}
    \caption{\textbf{Layer-wise separation probabilities for GPT-OSS-120B and Qwen3.5-9B.}
    (a) For GPT-OSS-120B, layout-based separation is strongest in early layers, while action-based separation increases to a peak of approximately 68\% in intermediate layers before declining slightly toward the final layers. (b) For Qwen3.5-9B, layout-based separation is also strongest initially, but action-based separation overtakes it in early layers and generally increases toward the final layers, reaching approximately 70\%. In both models, concept- and selector-based separation remain comparatively weaker. The dashed horizontal line indicates chance-level separation (50\%).}
    \label{fig:separation_probability_other_models}
\end{figure}

Figure~\ref{fig:separation_probability_other_models} shows similar trends for GPT-OSS-120B and Qwen3.5-9B. In both models, layout-based separation is strongest in early layers, while action-based separation becomes dominant in later layers. For GPT-OSS-120B (Figure~\ref{fig:gpt_oss_120b_separation_probability}), action separation increases from 60.92\% at layer 0 to a peak of 67.93\% at layer 20, before declining slightly to 65.15\% in the final layer. For Qwen3.5-9B (Figure~\ref{fig:qwen35_9b_separation_probability}), action separation overtakes layout separation earlier and increases more consistently, from 61.47\% at layer 0 to a peak of 69.69\% at layer 30, remaining high at 69.33\% in the final layer. In both models, concept- and selector-based separation remain comparatively weaker.

\begin{figure}[tpb]
    \centering
    \begin{subfigure}[t]{0.44\linewidth}
        \centering
        \includegraphics[width=\linewidth]{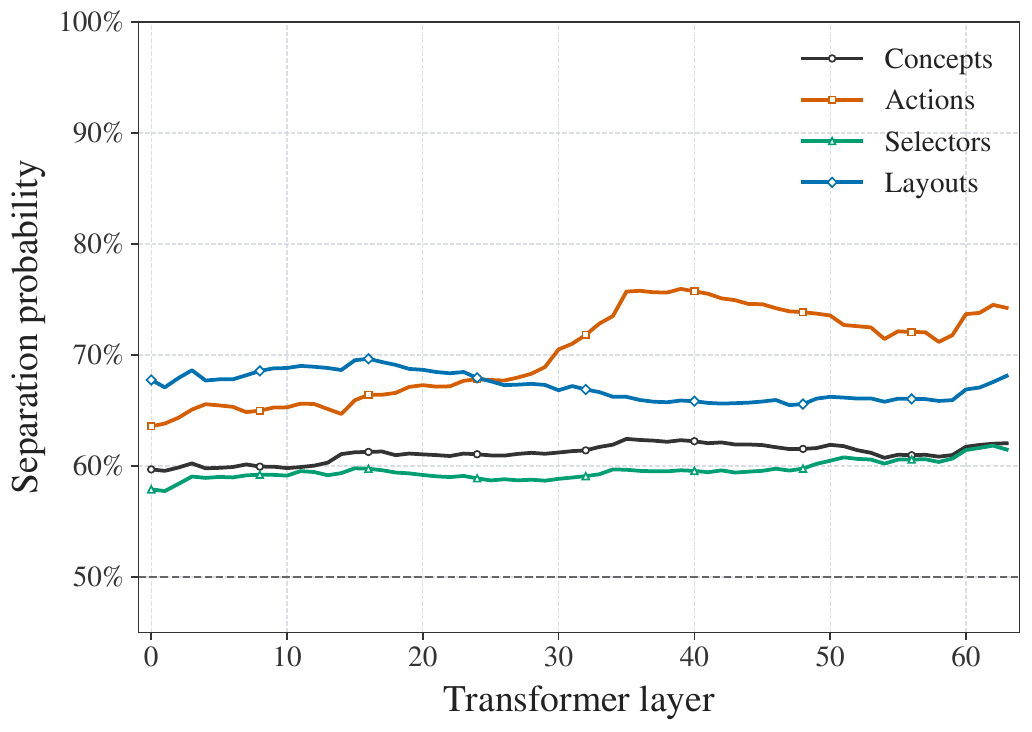}
        \caption{Predicate concepts}
        \label{fig:qwen35_27b_separation_probability_predicate_concepts}
    \end{subfigure}
    \hfill
    \begin{subfigure}[t]{0.44\linewidth}
        \centering
        \includegraphics[width=\linewidth]{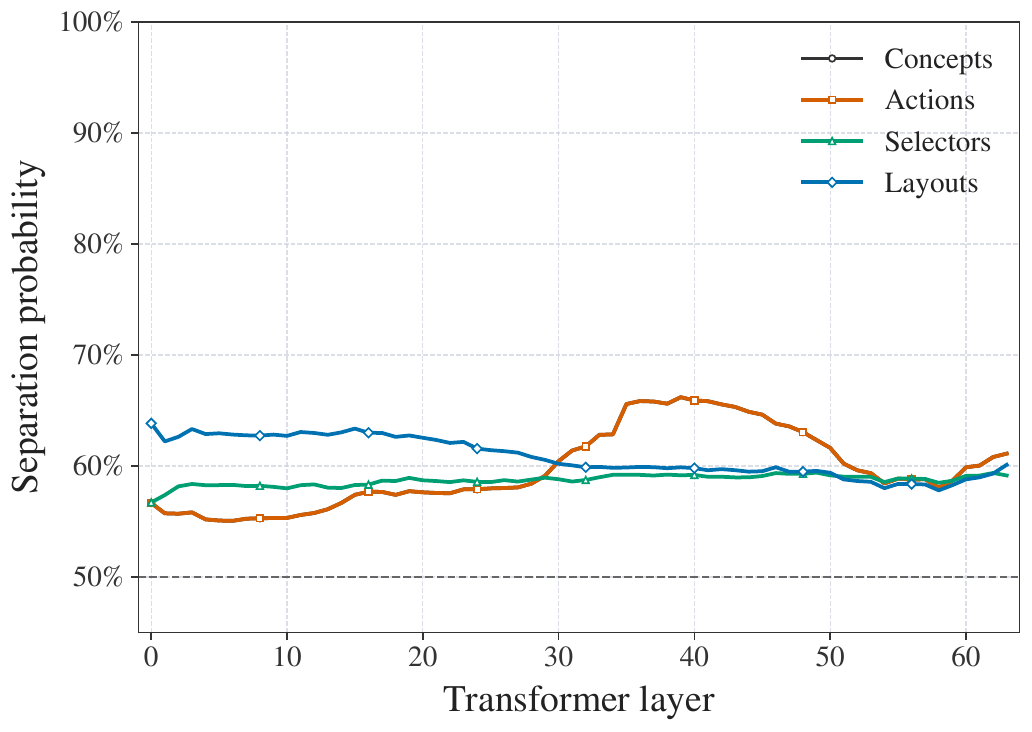}
        \caption{Action concepts}
        \label{fig:qwen35_27b_separation_probability_action_concepts}
    \end{subfigure}
    \caption{\textbf{Layer-wise separation probabilities for Qwen3.5-27B, evaluated separately on predicate and action concepts.}
    (a) For predicate concepts, layout-based separation is strongest in early layers, while action-based separation increases substantially in intermediate layers, reaching approximately 76\% and remaining strongest toward the final layers. (b) For action concepts, layout-based separation is also strongest initially, but gradually decreases, while action-based separation peaks at approximately 66\% in intermediate layers before declining toward the final layers, where all four separation probabilities converge to around 60\%. Concept- and action-based separation are aligned for action concepts, reflecting the close correspondence between concept labels and the actions applied in these tasks. The dashed horizontal line indicates chance-level separation (50\%).}
    \label{fig:qwen35_separation_probability_by_concept_type}
\end{figure}

\begin{figure}[tpb]
    \centering
    \begin{subfigure}[t]{0.44\linewidth}
        \centering
        \includegraphics[width=\linewidth]{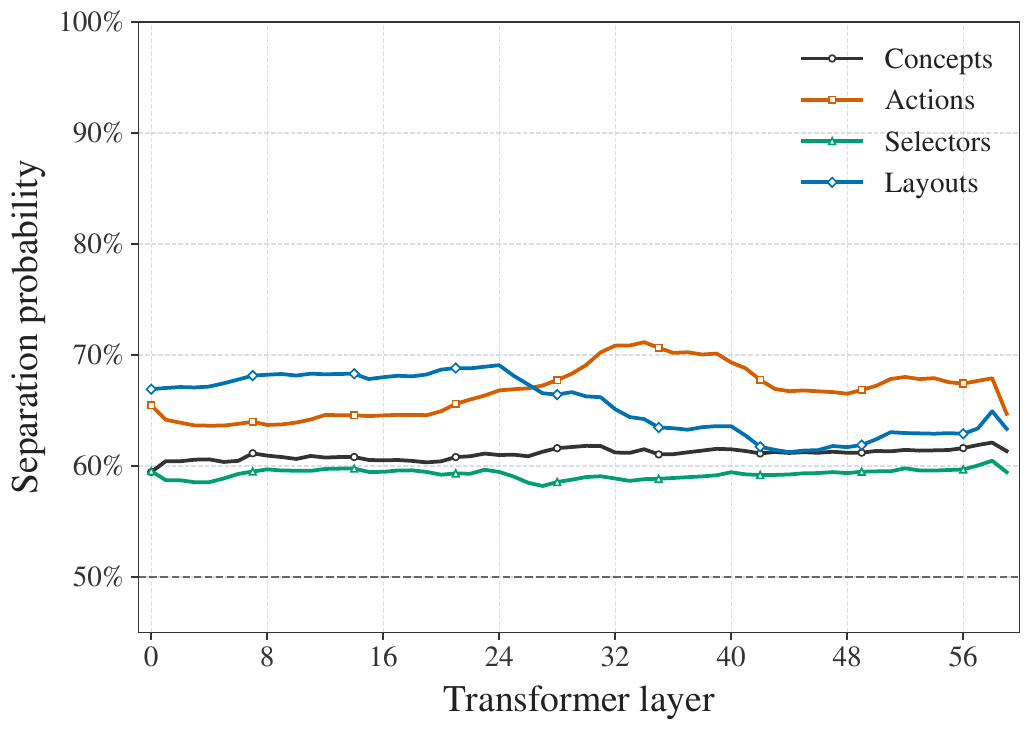}
        \caption{Predicate concepts}
        \label{fig:gemma4_separation_probability_predicate_concepts}
    \end{subfigure}
    \hfill
    \begin{subfigure}[t]{0.44\linewidth}
        \centering
        \includegraphics[width=\linewidth]{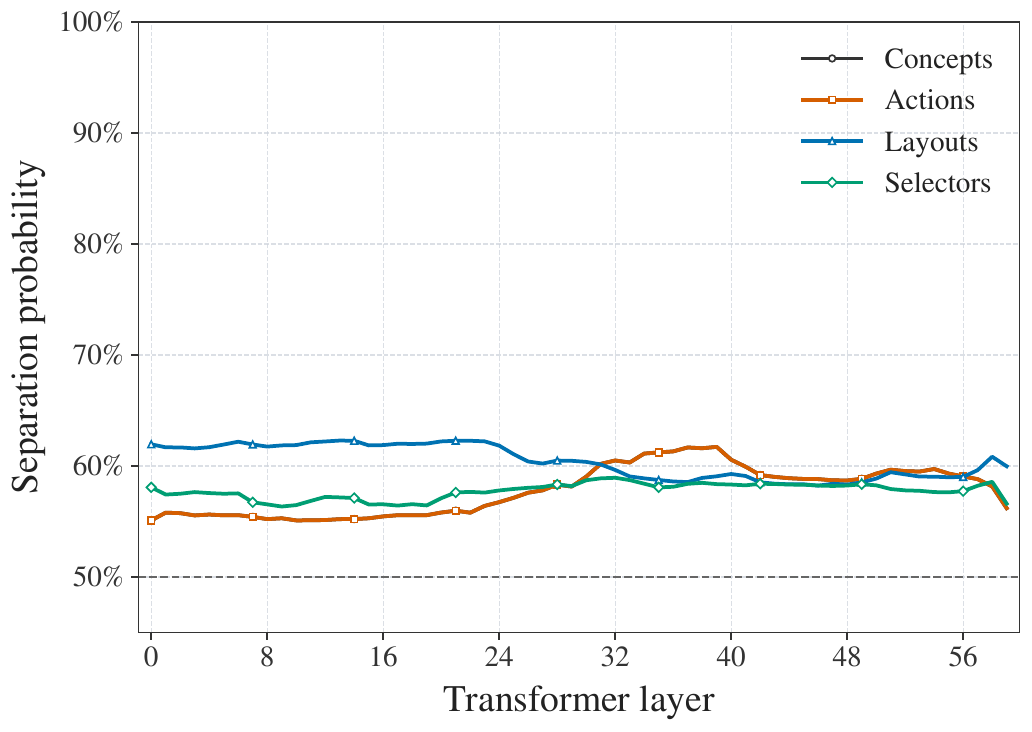}
        \caption{Action concepts}
        \label{fig:gemma4_separation_probability_action_concepts}
    \end{subfigure}
    \caption{\textbf{Layer-wise separation probabilities for Gemma4-31B, evaluated separately on predicate and action concepts.}
    For predicate concepts, layout-based separation is strongest in early layers, while action-based separation increases toward intermediate layers, reaching approximately 71\% before declining slightly but remaining strongest toward the final layers. (b) For action concepts, layout-based separation is also strongest initially but decreases in intermediate layers, while action-based separation increases modestly to a peak of approximately 62\%. All four separation probabilities remain relatively close in later layers, with concept- and action-based separation exhibiting similar values, reflecting the close correspondence between concept labels and actions for these tasks. The dashed horizontal line indicates chance-level separation (50\%).}
    \label{fig:gemma4_separation_probability_by_concept_type}
\end{figure}

To investigate whether the observed shift toward action-based organization depends on the type of concept, we repeat the separation-probability analysis separately for predicate concepts and action concepts (Figures~\ref{fig:qwen35_separation_probability_by_concept_type} and~\ref{fig:gemma4_separation_probability_by_concept_type}). As discussed in Section~\ref{sec:data_generator}, predicate concepts primarily define spatial or relational conditions that determine which objects are selected, whereas action concepts primarily specify the transformations to be performed. This distinction allows us to examine whether action-based organization also emerges for tasks whose defining concept is not itself an action.

For Qwen3.5-27B (Figure~\ref{fig:qwen35_separation_probability_by_concept_type}), the two subsets exhibit distinct trends. For predicate concepts (Figure~\ref{fig:qwen35_27b_separation_probability_predicate_concepts}), layout-based separation is strongest initially, but action-based separation increases substantially from 63.57\% at layer 0 to a peak of 75.95\% at layer 39, remaining high at 74.24\% in the final layer. For action concepts (Figure~\ref{fig:qwen35_27b_separation_probability_action_concepts}), layout separation is again strongest initially, while action separation reaches a lower peak of 66.19\% at layer 39 before declining to 61.10\% in the final layer, where all four separation probabilities converge to approximately 60\%. Notably, concept- and action-based separation align for action concepts, reflecting the close correspondence between their concept labels and associated actions.

Gemma4-31B (Figure~\ref{fig:gemma4_separation_probability_action_concepts}) exhibits a similar pattern, although the shift toward action-based separation is less pronounced. For predicate concepts (Figure~\ref{fig:gemma4_separation_probability_predicate_concepts}), layout separation dominates early layers, while action separation increases to a peak of 71.14\% at layer 34 and remains the strongest in the final layers. For action concepts (Figure~\ref{fig:gemma4_separation_probability_action_concepts}), action separation increases more modestly, reaching 61.71\% at layer 39 before declining toward the final layers, where layout separation again becomes strongest. As with Qwen3.5-27B, concept- and action-based separation align for action concepts. Together, these results suggest that action-based organization is not limited to concepts explicitly defined by transformations, but also emerges for predicate concepts.

\begin{figure}[tbp]
\centering
\resizebox{\textwidth}{!}{%
    \input{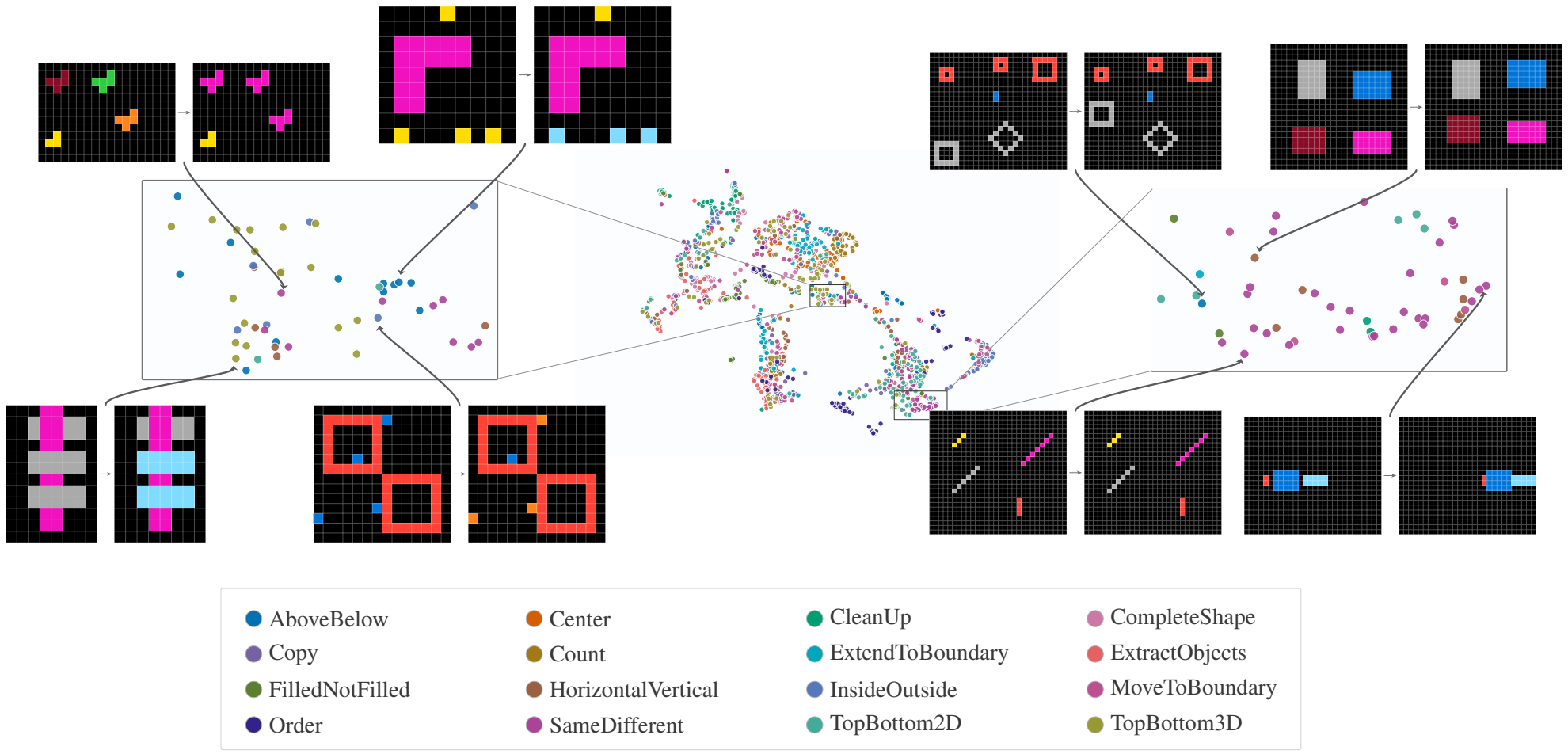}
}
\caption{\textbf{Qwen3.5-27B organizes task representations around semantically meaningful action clusters.} UMAP projection of Qwen3.5-27B's task representations at layer 41. Tasks requiring similar actions, i.e. transformation operations, form distinct clusters. For instance, in the cluster enlarged on the left, the model represents tasks in which objects are \emph{colored}. Notably, the model clusters these tasks even though they belong to different concepts and follow substantially different rules. In the cluster enlarged on the right, Qwen3.5-27B similarly groups tasks involving \emph{translation}. These tasks are represented closely in the model's hidden state space even though they are characterized by different grid layouts and belong to different concepts.}
\label{fig:qwen_layer41_embeddings}
\end{figure}

Figure~\ref{fig:qwen_layer41_embeddings} visualizes the task representations of Qwen3.5-27B at layer 41 using a two-dimensional UMAP projection of pairwise cosine distances, with 15 neighbors and a minimum-distance parameter of 0.1, as described in Section~\ref{subsec:repr_analysis}. Similar to our findings for VARC, we observe semantic structure in the model's representation space: Qwen3.5-27B represents tasks that share specific actions close together in clusters, supporting the quantitative results shown in Figure~\ref{fig:qwen35_27b_separation_probability}. For instance, the enlarged cluster on the left contains tasks from different concepts that all involve coloring specific objects. In comparison, the enlarged cluster on the right contains tasks from different concepts that involve translating certain objects. Notably, the model represents tasks with the same action similarly in its representation space, even when these tasks arise from different rules, scenarios, and concepts. This suggests that the model forms shared abstractions of transformation operations that generalize across diverse tasks.

\subsubsection{Action separation under controlled visual grid properties}
\label{subsubsec:controlled_action_sep}
As described in Section~\ref{subsec:repr_analysis}, we analyze whether Qwen3.5-27B's strong action-based separation (Figure~\ref{fig:qwen35_27b_separation_probability}) could reflect visual properties that tasks with the same action label tend to share, but tasks with different action labels do not. We therefore recompute action separation across layers while controlling for action frequency, layout, and visual grid properties.

For each input--output support example, we extract four geometry features and ten color proportions from each grid, along with the number and fraction of cells whose color changes between input and output (see Table~\ref{tab:visual_matching_features} for details). This yields 14 features for the input-only profile and 30 for the input--output profile. Concatenating the three support examples produces 42- and 90-dimensional feature vectors, respectively, which we average across each task's eight support-set variants. We remove coordinates that are constant across the 1{,}600 parent tasks, standardize the remaining coordinates to zero mean and unit variance, and divide each feature block (geometry, color, or pixel changes) by the square root of its retained coordinate count. Visual matching uses Euclidean distances between the resulting vectors.

When computing action separation for a reference task $t_i$ with action label $a_i$, we compare tasks $t_j$ and $t_k$ with $a_j = a_i$ and $a_k \neq a_i$, respectively. We evaluate eight control conditions against the original analysis: action-frequency balancing alone, balancing combined with layout matching, and six conditions additionally controlling for visual similarity. Layout matching requires $t_j$ and $t_k$ to share the same exact layout family, while leaving the reference task $t_i$ unrestricted in its layout and visual properties. For visual matching, we compute Euclidean distances between all eligible task pairs, separately for the input-only and input--output feature profiles. For each action label with at least ten parent tasks, we keep pairs whose distance falls below the 25th, 50th, or 75th percentile of the corresponding distance distribution (q25/q50/q75), yielding six visual matching conditions. The comparison sets are determined independently of the model representations and remain fixed across layers. For each anchor and target action family, we compute separation separately within each available negative action family, assigning half credit to ties as in the original analysis.

\begin{figure}[tbp]
\centering
\includegraphics[width=0.855\textwidth]{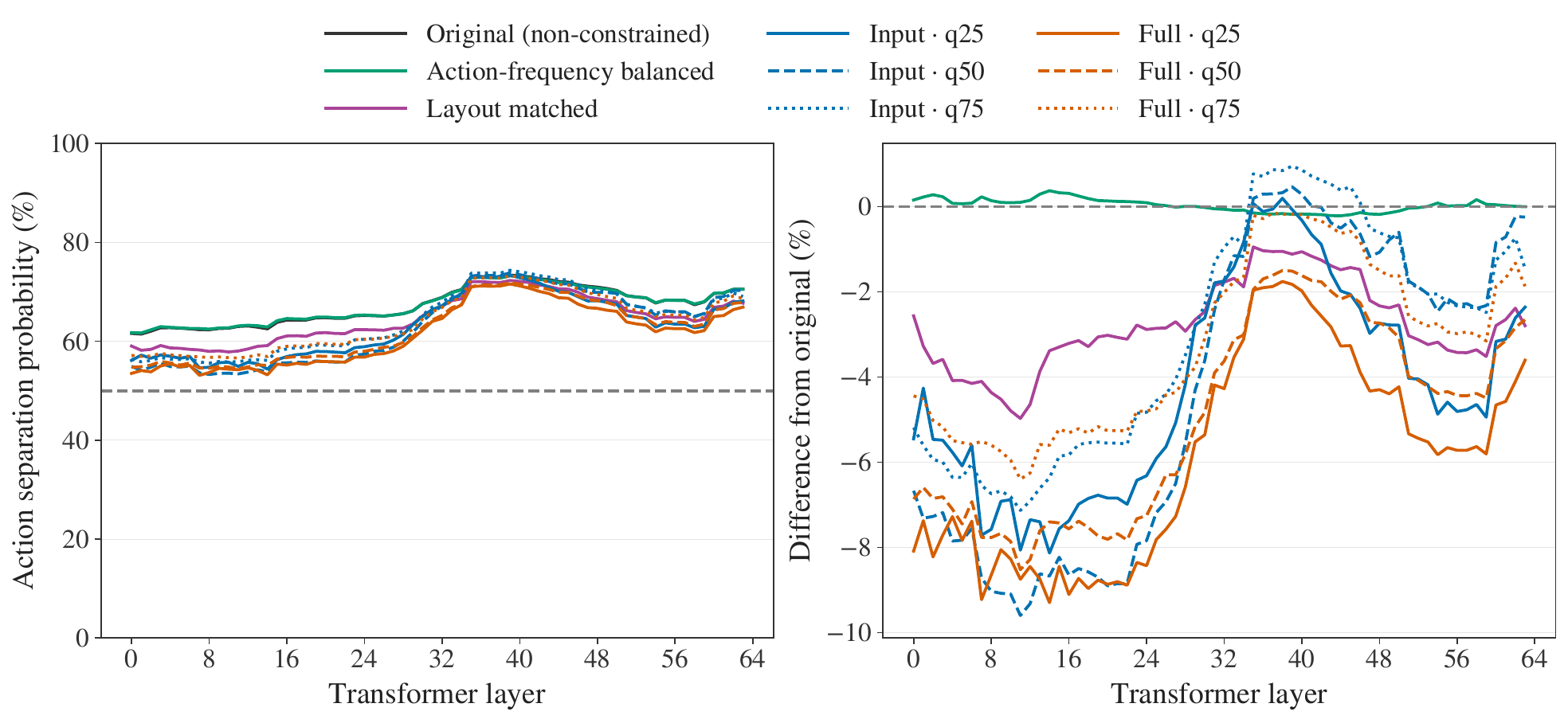}
\caption{\textbf{Action-based separation remains high under visual grid controls.}
(Left) action separation probability across Qwen3.5-27B layers, and (right) difference from the uncontrolled baseline. Eight control conditions progressively balance action frequencies, match grid layouts, and restrict comparisons by visual similarity of input grids alone or both input and output grids at three distance thresholds (q25/q50/q75). Visual matching substantially reduces separation in early and late layers but has a smaller effect near its intermediate-layer peak, where separation decreases by only $1.8\%$ under the strictest condition.}
\label{fig:qwen35_27b_sep_cos_tie_adjusted_controlled_full}
\end{figure}

Figure~\ref{fig:qwen35_27b_sep_cos_tie_adjusted_controlled_full} shows the action separation probabilities under each control condition (left) and their differences from the original analysis (right). Visual matching substantially reduces action-based separation in early and late layers. Near its peak in intermediate layers, however, separation decreases by only 1.8 percentage points, from $73.4\%$ to $71.6\%$, even under the strictest matching condition (\emph{input \& output $\cdot$ q25}).

\paragraph{A visual baseline for action separation.}
We also evaluate action separation using visual feature vectors instead of the model's task representations. Each task is represented by a 132-dimensional vector capturing geometry, color, spatial, and object features from the input and output grids of its three support examples (see Table~\ref{tab:visual_baseline_features} for details). We average these vectors across eight support-set variants and apply the same standardization and feature-block weighting as above. We then compute action separation using cosine similarity between visual vectors, assigning half credit to ties and reusing the same comparison sets, action-frequency balancing, and family eligibility criteria as in the controlled model analysis. Under the strictest input--output matching condition (q25), the visual baseline achieves $59.96\%$ action separation, compared with $71.59\%$ for Qwen3.5-27B at its peak layer, a difference of $11.63\%$.

\subsection{Extended VARC embedding representations}\label{subsec:extended_varc_embeddings}
\begin{figure}[tbp]
    \centering

    \begin{subfigure}[t]{0.75\linewidth}
        \centering
        \includegraphics[width=\linewidth]{
            figures/representation\_analysis/varc\_task\_embeddings/Embed-Full.pdf}
        \caption{All embeddings.}
        \label{fig:embed-full}
    \end{subfigure}

    \vspace{0.8em}

    \begin{subfigure}[t]{0.48\linewidth}
        \centering
        \includegraphics[width=\linewidth]{
            figures/representation\_analysis/varc\_task\_embeddings/Embed-Actions.pdf}
        \caption{Actions only are displayed.}
        \label{fig:embed-actions}
    \end{subfigure}
    \hfill
    \begin{subfigure}[t]{0.48\linewidth}
        \centering
        \includegraphics[width=\linewidth]{
            figures/representation\_analysis/varc\_task\_embeddings/Embed-Predicates.pdf}
        \caption{Predicates only are displayed.}
        \label{fig:embed-predicates}
    \end{subfigure}

    \vspace{1.6em}

    \begin{subfigure}[t]{0.98\linewidth}
        \centering
        \includegraphics[width=\linewidth]{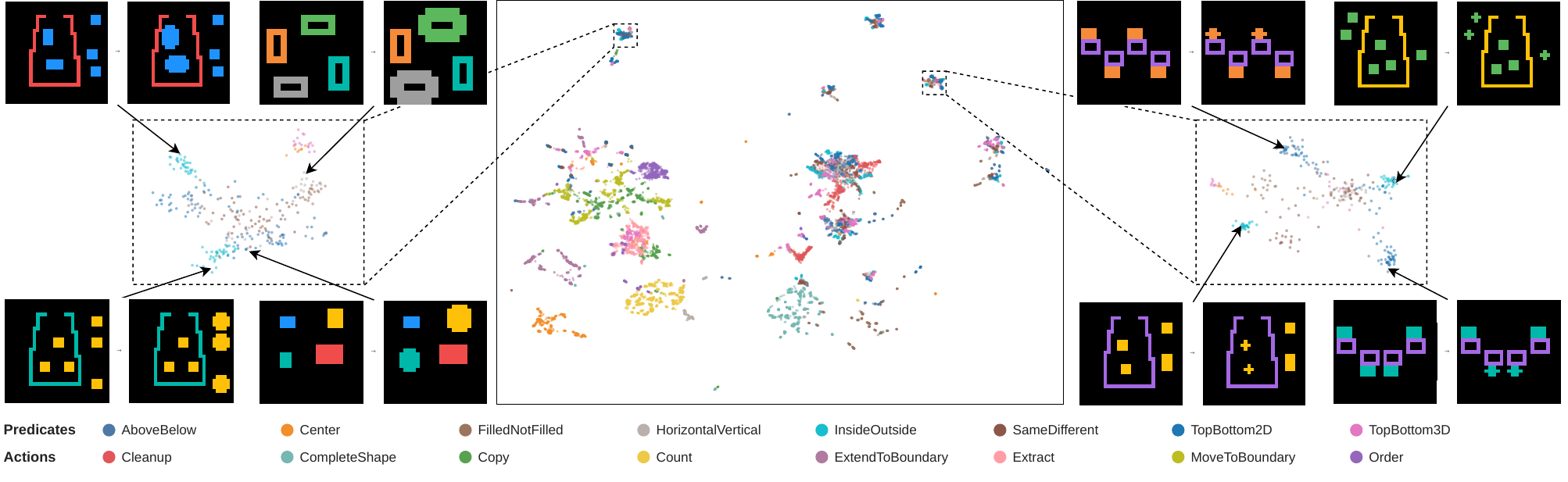}
        \caption{Zoom on two seemingly mixed clusters.}
        \label{fig:embed-zoom}
    \end{subfigure}

    \caption{UMAP projection of our 12,800 VARC embeddings (a), colored by concepts. Then, only the action concepts (b) are shown, separated from the predicate concepts (c). Action concepts are seemingly much better naturally clustered than predicate concepts. However, zooming (d) on some predicates clusters reveal an interesting structure, that is, clusters still represent some ``actions'' operating on the grids. Within a cluster, opposite actions are located far apart, further distinguishing what exactly needs to be done to solve the tasks.}
    \label{fig:embedding-comparison}
\end{figure}

We further examine the VARC embeddings obtained when training the model on 1600 ConceptARC-100 tasks, with 8 task instances each, giving 12,800 task embeddings. Figure~\ref{fig:embed-full} shows their UMAP 2D-projection, each embedding being colored according to its concept. As a first observation, action concepts seem better clustered than predicate concepts, with small clusters appearing then and there without any apparent reason. This difference of behavior between the two groups is explicitly visualized on~\Cref{fig:embed-actions,fig:embed-predicates}. Interestingly, in the predicates panel, the ``\emph{Center}'' concept seems clustered apart. It is actually split into one major cluster at the bottom left, corresponding to tasks more akin to "the action of centering something", and scattered tasks close to other concept tasks. This further reinforces the observation that tasks corresponding to an actual \emph{action} tend to be better clustered together. We then further investigate the small clusters appearing in the predicates panel, in Figure~\ref{fig:embed-zoom}. It appears that each cluster is consistent with some kind of action to perform, shared across various concepts. On the left, the action is akin to a dilation, based on a concept-related condition (inside/outside, horizontal/vertical, above/below, top/bottom mainly). Moreover, within that cluster, for a given concept, opposite conditions of a concept yield opposite sub-clusters: tasks related to dilating something \emph{outside} are opposed to those related to dilating something \emph{inside} an object. The same observation can be made on the cluster on the right, corresponding more to an \emph{erosion} based on some concept-related condition. Opposite conditions are again clearly placed in opposite locations of the cluster. Even more strikingly, these two clusters are themselves opposites: one is a dilation cluster, the other an erosion, and these two clusters are placed far apart in the global UMAP projection. All these observations tend to indicate that task embeddings mainly encode actions, operations that need to be performed in order to solve the tasks, and that the conditions or objects on which the operations are conducted are then more finely used within those action concepts. 

\newpage

\begin{table*}[t]
    \centering
    \small
    \renewcommand{\arraystretch}{1.15} 
\caption{\textbf{Visual features used for matching tasks in the controlled action separation analysis.}
Each grid contributes four geometry and ten color features, with two additional features capturing pixel changes between input and output. Concatenating features across three support examples per task produces 42- and 90-dimensional vectors, respectively.}
    \label{tab:visual_matching_features}
    \begin{tabular}{@{}
        >{\raggedright\arraybackslash}p{0.15\linewidth}
        >{\raggedright\arraybackslash}p{0.24\linewidth}
        >{\raggedright\arraybackslash}p{0.53\linewidth}
    @{}}
        \toprule
        Type & Feature & Description and computation \\
        \midrule

        Geometry
        & Height
        & Number of grid rows, $H$. \\

        & Width
        & Number of grid columns, $W$. \\

        & Area
        & Total number of grid cells, $H \times W$. \\

        & Aspect ratio
        & Grid width divided by height, $W/H$. \\

        \addlinespace
        Color
        & Color proportions
        & For each color $k \in \{0,\ldots,9\}$, divide the number of
          cells with that color by the grid area. This gives ten features,
          including the proportion of background cells (color~0). \\

        \addlinespace
        Pixel changes
        & Changed-cell count
        & Number of positions whose color differs between the input and
          output grids. This includes adding, removing, and recoloring
          foreground cells. \\

        & Changed-cell fraction
        & Changed-cell count divided by the grid area, $H \times W$. \\

        \bottomrule
    \end{tabular}
\end{table*}

\begin{table*}[t]
    \centering
    \small
    \renewcommand{\arraystretch}{1.15}
\caption{\textbf{Visual features used to represent tasks in the action separation baseline.}
Each input and output grid contributes 22 features capturing geometry, color, spatial, and object properties. Concatenating features across both grids and three support examples yields a 132-dimensional vector.}
    \label{tab:visual_baseline_features}
    \begin{tabular}{@{}
        >{\raggedright\arraybackslash}p{0.15\linewidth}
        >{\raggedright\arraybackslash}p{0.24\linewidth}
        >{\raggedright\arraybackslash}p{0.53\linewidth}
    @{}}
        \toprule
        Type & Feature & Description and computation \\
        \midrule

        Geometry
        & Height
        & Number of grid rows, $H$. \\

        & Width
        & Number of grid columns, $W$. \\

        & Area
        & Total number of grid cells, $H \times W$. \\

        & Aspect ratio
        & Grid width divided by height, $W/H$. \\

        \addlinespace
        Color
        & Color proportions
        & For each color $k \in \{0,\ldots,9\}$, divide the number of
          cells with that color by the grid area. This gives ten features,
          including the proportion of background cells (color~0). \\

        \addlinespace
        Spatial
        & Foreground height
        & Height of the smallest rectangle containing all foreground
          cells, divided by $H$. \\

        & Foreground width
        & Width of the smallest rectangle containing all foreground
          cells, divided by $W$. \\

        & Horizontal foreground center
        & Mean column index of all foreground cells, divided by $W-1$.
          Column indices start at zero. \\

        & Vertical foreground center
        & Mean row index of all foreground cells, divided by $H-1$.
          Row indices start at zero. \\

        \addlinespace
        Objects
        & Object count
        & Number of foreground objects. \\

        & Largest-object fraction
        & Number of visible cells in the largest object, divided by
          the grid area, $H \times W$. \\

        & Mean object height
        & Average height of the objects' bounding rectangles,
          divided by $H$. Each object receives equal weight. \\

        & Mean object width
        & Average width of the objects' bounding rectangles,
          divided by $W$. Each object receives equal weight. \\

        \bottomrule
    \end{tabular}
\end{table*}

\newpage

\begin{table}[t]
\centering
\scriptsize
\setlength{\tabcolsep}{3pt}
\renewcommand{\arraystretch}{1.08}
\setlength{\extrarowheight}{1pt}

\caption{\textbf{Descriptions of the 16 ConceptARC concepts and their grouping into predicate and action concepts.}}
\label{tab:conceptarc_concepts}

\begin{tabular}{@{}
    >{\centering\arraybackslash}m{0.09\textwidth} |
    >{\centering\arraybackslash}p{0.18\textwidth} |
    >{\raggedright\arraybackslash}p{0.69\textwidth}
@{}}
\hline
\textbf{Type}
& \textbf{Concept}
& \multicolumn{1}{c}{\textbf{Description}} \\
\hline

\multirow{8}{0.09\textwidth}{%
    \centering\raisebox{-3.5cm}{\emph{Predicate}}%
}
& \emph{Above and Below}
& The transformation depends on cells or objects lying above or below a visible reference, such as a line, band, diagonal, or distinguished object. Content on a specified side may be selected, removed, recolored, extended, or moved. Unlike \emph{Top and Bottom 2D}, the relation is defined by a reference rather than solely by vertical rank. \\
\cline{2-3}

& \emph{Center}
& The transformation uses the geometric center or midpoint of a grid, object, line, gap, container, or pair of objects. It may mark, extract, remove, or fill the center, move an object relative to it, or otherwise perform an operation determined by the center. \\
\cline{2-3}

& \emph{Filled and Not Filled}
& The defining distinction is whether a region or object is solid or hollow. The transformation may select, label, fill, empty, or invert objects according to their interior state. \\
\cline{2-3}

& \emph{Horizontal and Vertical}
& The operation depends on horizontal versus vertical orientation or on an object's dominant axis. Horizontal and vertical objects, internal stripes, marker alignments, or directions may be selected, removed, recolored, or extended differently. \\
\cline{2-3}

& \emph{Inside and Outside}
& Containment determines which cells or objects are relevant: for example, whether they lie inside, outside, or between containers, outlines, or boundary lines. The selected content may then be retained, removed, recolored, counted, extracted, or otherwise transformed. \\
\cline{2-3}

& \emph{Top and Bottom 2D}
& The vertical position or rank of whole objects in a two-dimensional grid determines the operation, such as selecting, extracting, marking, rearranging, or otherwise manipulating the topmost, bottommost, higher, lower, or intermediate object. Unlike \emph{Above and Below}, this concept normally compares objects by vertical rank without an explicit separating reference. \\
\cline{2-3}

& \emph{Top and Bottom 3D}
& Overlap and occlusion are interpreted as depth ordering, such that one object appears in front of or behind another. The transformation may select or reconstruct a layer, change a crossover cell, propagate the visible layer's color, swap which object appears on top, or otherwise manipulate objects according to their depth ordering. \\
\cline{2-3}

& \emph{Same and Different}
& Objects, shapes, colors, interiors, or patterns are compared for equivalence, duplication, or difference. Matching or nonmatching items are then selected, removed, retained, or labeled, with the demonstrations determining which attributes define equivalence. \\

\hline

\multirow{8}{0.09\textwidth}{%
    \centering\raisebox{-3.0cm}{\emph{Action}}%
}
& \emph{Clean Up}
& The input contains noise, corruption, misplaced cells, incorrect colors, gaps, or an occluding distractor that disrupts an otherwise regular scene. The transformation restores the clean dominant pattern by correcting or removing these irregularities. \\
\cline{2-3}

& \emph{Complete Shape}
& Partial segments, fragments, reflections, or repeated templates imply a complete geometric shape or pattern. The transformation reconstructs the missing cells, sides, intersections, or counterparts needed to realize the latent complete form. \\
\cline{2-3}

& \emph{Copy}
& A complete source object or pattern is duplicated into another location, indicated region, incomplete counterpart, or expanded canvas. The copied structure is preserved, with replication rather than repair of a damaged shape constituting the central operation. \\
\cline{2-3}

& \emph{Count}
& A set of objects, cells, colors, lines, or other selected features is counted. The resulting quantity is encoded through the output dimensions, a colored run, a bar, or another compact representation. \\
\cline{2-3}

& \emph{Extend To Boundary}
& A marker, ray, line, or existing pattern grows in one or more indicated directions until reaching a grid edge, divider, enclosing boundary, or target object. New cells extend the structure rather than translating it intact. \\
\cline{2-3}

& \emph{Extract Objects}
& One or more relevant objects or recurring motifs are isolated from their surrounding context. The selected content is cropped or packed into a minimal output grid while preserving its geometry, colors, and required ordering. \\
\cline{2-3}

& \emph{Move To Boundary}
& One or more objects are translated as rigid units until they touch or align with a grid edge, reference line, another object, or another boundary. Their size and orientation remain intact, distinguishing this operation from growing a line as in \emph{Extend To Boundary}. \\
\cline{2-3}

& \emph{Order}
& Objects, regions, or colors are ranked according to properties such as size, count, height, marker position, or nesting order. The transformation rearranges them or serializes their values according to an inferred ascending, descending, outer-to-inner, or top-to-bottom sequence. \\

\hline
\end{tabular}
\end{table}

\begin{figure}[tbp]
\centering
\includegraphics[width=\linewidth]{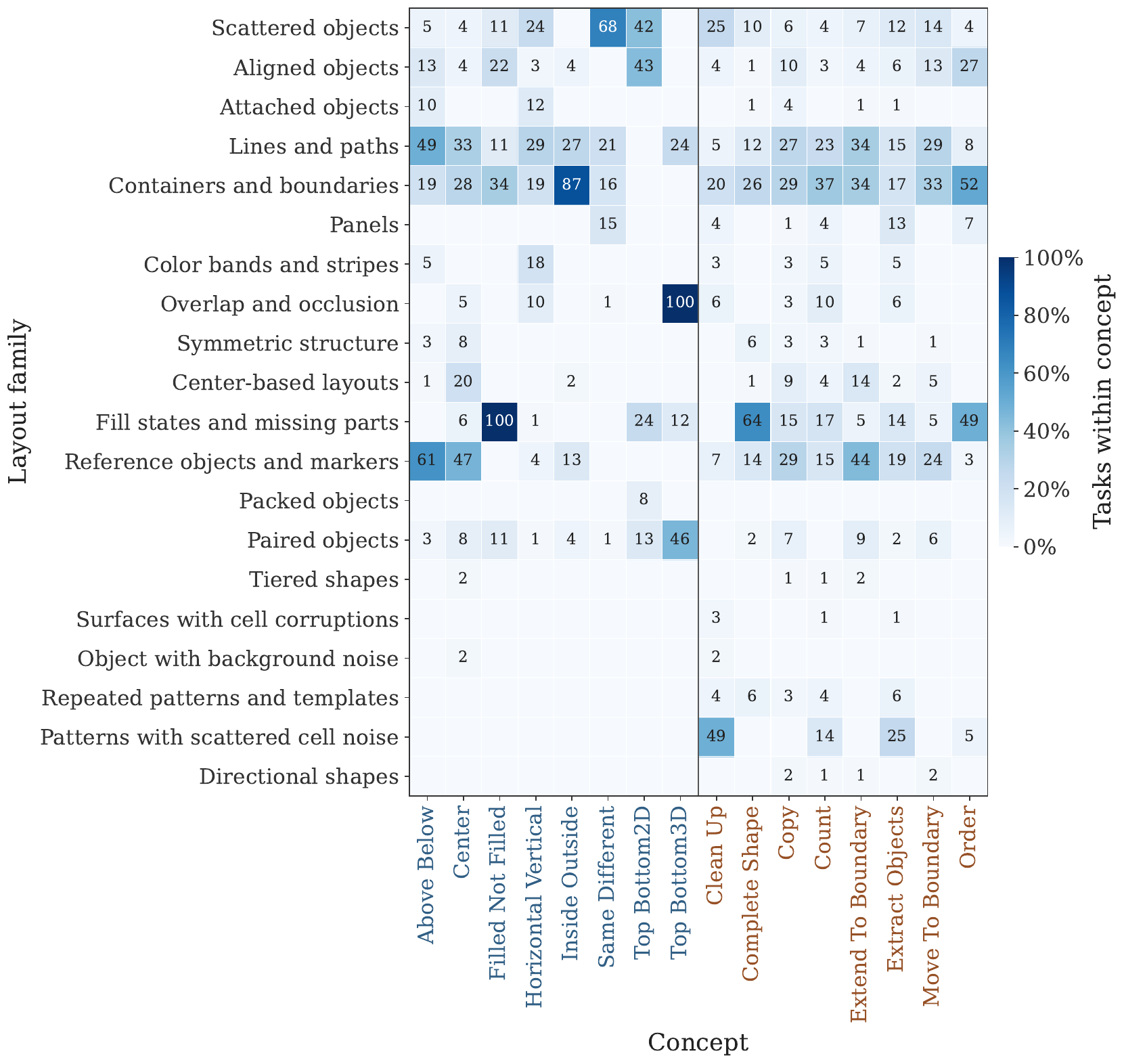}
\caption{\textbf{Distribution of layout families across concepts.} Each cell shows the percentage of tasks within a concept that use the corresponding layout family. A task may involve multiple layout families, so percentages within a concept need not sum to 100\%. Descriptions of each layout family can be found in Table~\ref{tab:layout-families}.}
\label{fig:layouts_concept_heatmap}
\end{figure}

\begin{figure}[tbp]
\centering
\includegraphics[width=\linewidth]{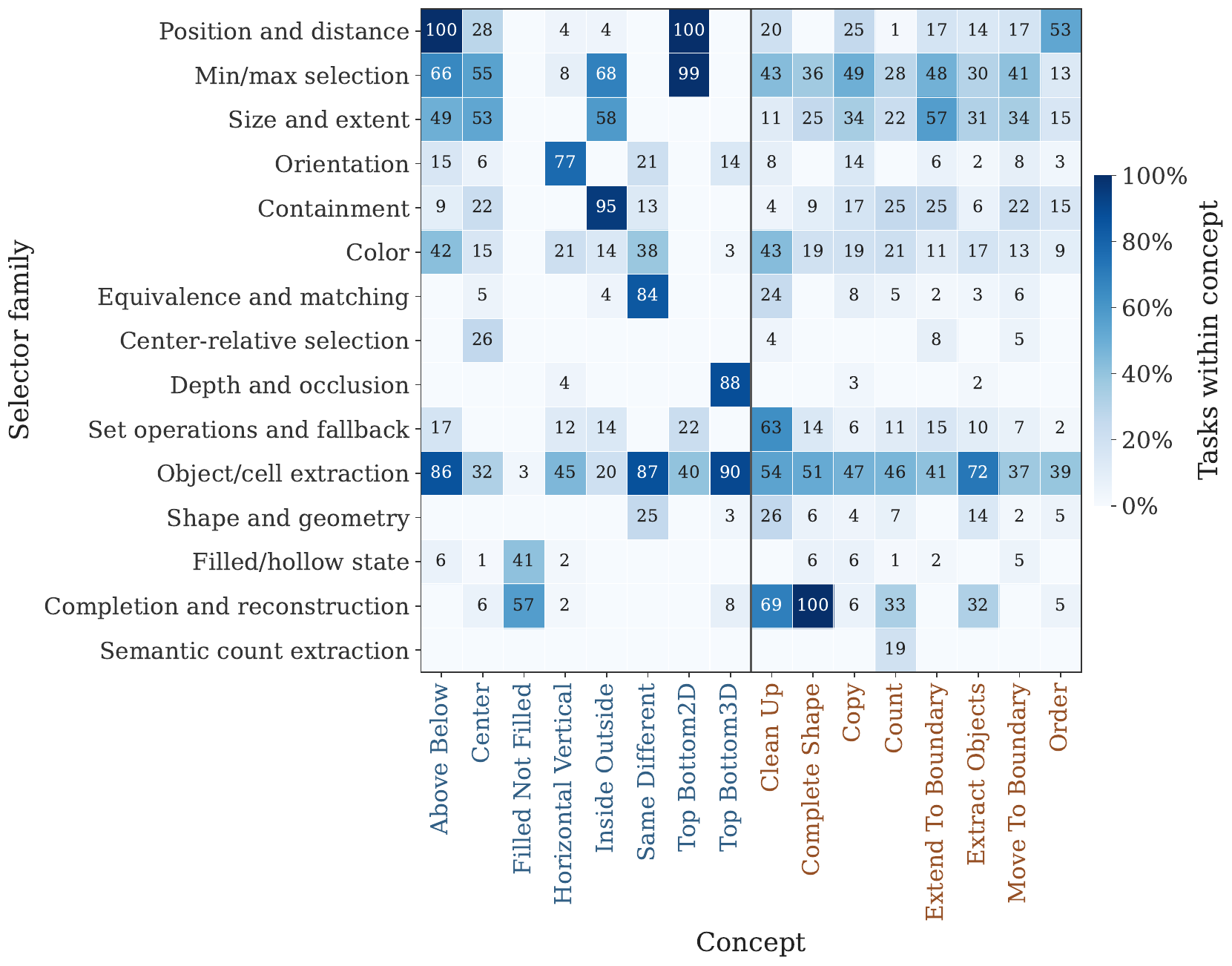}
\caption{\textbf{Distribution of selector families across concepts.} Each cell shows the percentage of tasks within a concept that use the corresponding selector family. A task may involve multiple selector families, so percentages within a concept need not sum to 100\%. Descriptions of each selector family can be found in Table~\ref{tab:selector-families}.}
\label{fig:selector_concept_heatmap}
\end{figure}

\begin{figure}[tbp]
\centering
\includegraphics[width=\linewidth]{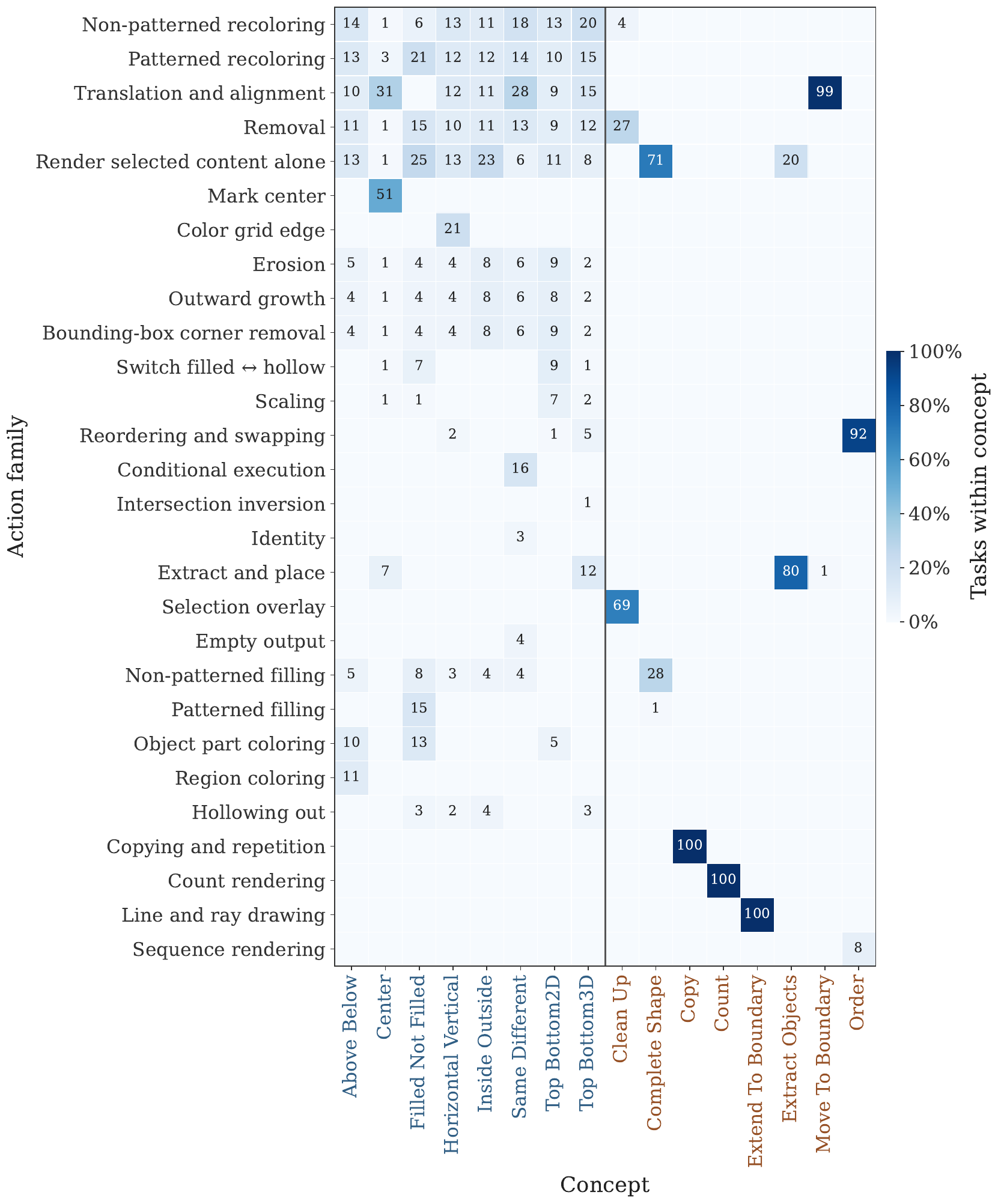}
\caption{\textbf{Distribution of action families across concepts.} Each cell shows the percentage of tasks within a concept that use the corresponding action family. A task may involve multiple action families, so percentages within a concept need not sum to 100\%. Descriptions of each action family can be found in Table~\ref{tab:action-families}.}
\label{fig:action_concept_heatmap}
\end{figure}

\begin{table*}[t]
\centering
\small
\renewcommand{\arraystretch}{1.15}
\caption{\textbf{Textual descriptions of layout families.} Each layout family describes a type of input-grid structure used by the task generator.}
\label{tab:layout-families}

\begin{tabular}{@{}
  >{\raggedright\arraybackslash}p{0.29\linewidth}
  >{\raggedright\arraybackslash}p{0.66\linewidth}
@{}}
\toprule
\textbf{Layout family} & \textbf{Description} \\
\midrule
Scattered objects &
Several distinct objects or lines are placed at separate locations across the grid. \\
Aligned objects &
Objects are placed in aligned sequences or stacks, for example along a shared row or column. \\
Attached objects &
Objects are physically connected to reference objects, for example by touching or extending from their sides. \\
Lines and paths &
The scene contains lines, bars, strips, or paths, including continuous, dashed, and interrupted forms. \\
Containers and boundaries &
Frames, outlines, or dividing lines define enclosed regions, distinguish inside from outside, or separate areas of the grid. \\
Panels &
The grid is divided into distinct sections by separators or panel boundaries, with separate content in each section. \\
Color bands and stripes &
Colors are arranged in bands or repeating stripes across the grid or within individual regions. \\
Overlap and occlusion &
Shapes or lines may overlap or cross, hiding parts of one another. Overlap is optional in some layouts. \\
Symmetric structure &
Objects or patterns have an underlying symmetry. Added markers or missing cells may make the visible grid asymmetric. \\
Center-based layouts &
An object or marker is placed at, or deliberately away from, the center of the grid, another object, or a frame. \\
Fill states and missing parts &
Shapes have a specified fill or completion state: solid, hollow, partly filled, or missing cells from an outline or pattern. \\
Reference objects and markers &
The scene includes a designated reference object or marker, such as a guide line, an endpoint dot, or a template. \\
Packed objects &
Objects with the same general kind of shape are placed close together in a compact area. \\
Paired objects &
Two objects or shapes form an explicitly defined pair; they may be separated, touching, or overlapping. \\
Tiered shapes &
Shapes are built from distinct horizontal levels that can be counted, as in a stepped shape or a pyramid. \\
Surfaces with cell corruptions &
A uniform field or a scene with a solid rectangle contains scattered color errors, such as isolated cells of another color. \\
Object with background noise &
Extra colored cells are scattered in the background around an object, while the object's own cells are left unchanged. \\
Repeated patterns and templates &
The same shape or color pattern appears repeatedly, either across the grid or as complete or partial copies of a template. \\
Patterns with scattered cell noise &
Geometric shapes or patterns are accompanied by scattered noise cells, which may lie within the shapes or in the surrounding background. \\
Directional shapes &
Connected shapes have a geometry that indicates a direction: up, down, left, or right. \\
\bottomrule
\end{tabular}
\end{table*}

\begin{table*}[t]
\centering
\small
\renewcommand{\arraystretch}{1.15}
\caption{\textbf{Textual descriptions of selector families.} Each selector family describes how objects, cells, or regions are identified for a transformation.}
\label{tab:selector-families}

\begin{tabular}{@{}
  >{\raggedright\arraybackslash}p{0.29\linewidth}
  >{\raggedright\arraybackslash}p{0.66\linewidth}
@{}}
\toprule
\textbf{Selector family} & \textbf{Description}\\
\midrule
Position and distance &
Select objects or cells by position or distance, such as above a reference or nearest to it. \\

Min/max selection &
Select items with a minimum or maximum value, such as size, position, distance, or color frequency. \\

Size and extent &
Select objects or containers by size, dimensions, or line length. \\

Orientation &
Select objects or cells by orientation or aspect ratio, such as horizontal versus vertical. \\

Containment &
Select containers, enclosed regions, or objects and cells inside or outside a region. \\

Color &
Select by color, color frequency, or agreement with a reference color. \\

Equivalence and matching &
Select objects according to shared properties or matches to a reference object or pattern. \\

Center-relative selection &
Select objects or cells at, near, or away from a specified center. \\

Depth and occlusion &
Select shapes or layers by front/back order, visibility, or occlusion. \\

Set operations and fallback &
Combine selections using union, intersection, or difference, or use an alternative selection when needed. \\

Object/cell extraction &
Collect objects, cells, regions, or pairs for subsequent selection or transformation. \\

Shape and geometry &
Select objects or cells by shape, geometric dimensions, or agreement with a shape or pattern template. \\

Filled/hollow state &
Select objects according to whether their interiors are filled or hollow. \\

Completion and reconstruction &
Select by completeness, or infer object geometry to identify present, missing, or reconstructed parts. \\

Semantic count extraction &
Measure a quantity, such as the number of regions, shape levels, boundary sides, enclosed cells, or distinct colors.\\
\bottomrule
\end{tabular}
\end{table*}

\begin{table*}[t]
\centering
\small
\renewcommand{\arraystretch}{1.15}
\caption{\textbf{Textual descriptions of action families.} Each action family describes a type of transformation applied to selected content or the output grid.}
\label{tab:action-families}

\begin{tabular}{@{}
  >{\raggedright\arraybackslash}p{0.29\linewidth}
  >{\raggedright\arraybackslash}p{0.66\linewidth}
@{}}
\toprule
\textbf{Action family} & \textbf{Description} \\
\midrule
Non-patterned recoloring &
Recolor entire objects or layers in a consistent color while preserving their shapes and positions.\\
Patterned recoloring &
Apply an alternating color pattern (e.g., stripes or checkerboard) to existing objects or layers without adding, removing, or moving cells.\\
Translation and alignment &
Move selected content by a specified offset or align it with a reference position, center, or boundary.\\
Removal &
Erase selected objects, layers, or cells, replacing them with background cells.\\
Render selected content alone &
Keep only selected content, usually at its original position on a blank grid, or restricted to a specified grid region.\\
Mark center &
Mark a center or midpoint of the selected content or grid with a point, cross, or line segment. \\
Color grid edge &
Draw a solid or dashed colored stripe along an outer row or column of the grid.\\
Erosion &
Shrink objects or layers by removing their outermost boundary cells.\\
Outward growth &
Enlarge objects or layers by adding cells around their boundaries.\\
Bounding-box corner removal &
Remove cells at the corners of the smallest rectangle enclosing the object or layer.\\
Switch filled \ensuremath{\leftrightarrow} hollow &
Fill hollow or incomplete objects, while turning filled objects into hollow outlines.\\
Scaling &
Resize selected objects or layers by consistently increasing their size in all directions around a specified center.\\
Reordering and swapping &
Rearrange objects or layers, including swapping their positions or changing which layer appears in front.\\
Conditional execution &
Choose which action to perform according to whether a condition is satisfied.\\
Intersection inversion &
Where two shape layers overlap, switch which layer supplies the visible color.\\
Identity &
Return the input grid unchanged.\\
Extract and place &
Place or arrange selected content on a blank grid at specified positions, preserving its colors.\\
Selection overlay &
Draw selected or reconstructed content over the input grid at its original positions, preserving its colors.\\
Empty output &
Return a grid containing only background cells.\\
Non-patterned filling &
Fill gaps in an object's shape or interior using a solid and consistent color or a color inferred from the existing content.\\
Patterned filling &
Fill gaps in an object's shape or interior with an alternating color pattern (e.g., stripes or checkerboard). \\
Object part coloring &
Change the colors of specific existing object parts, such as individual cells, edges, or endpoints.\\
Region coloring &
Recolor a specified area or ray on the grid, which may include background cells.\\
Hollowing out &
Clear an object's or container's interior while leaving its boundary intact.\\
Copying and repetition &
Place selected content at one or more locations on the input grid; existing source content remains unless overwritten.\\
Count rendering &
Display the number of selected items or another measured count using colored cells, for example as a bar.\\
Line and ray drawing &
Draw one or more lines or rays with specified starting points, directions, and stopping conditions.\\
Sequence rendering &
Create an ordered sequence of colored cells, with one cell showing the color of each selected object.\\
\bottomrule
\end{tabular}
\end{table*}

\begin{figure}[tpb]
\centering
\resizebox{0.975\linewidth}{!}{\input{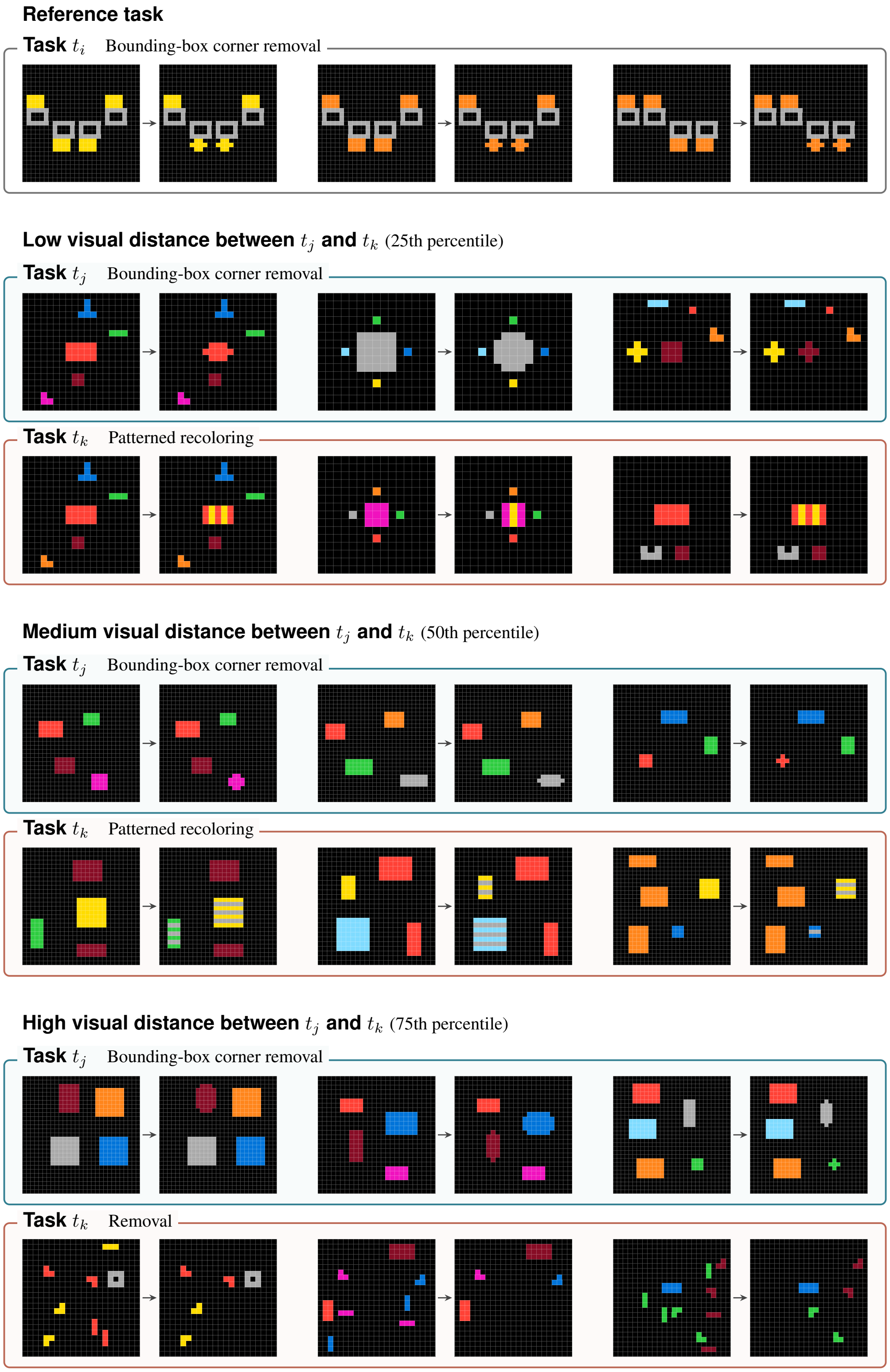}}
\caption{\textbf{ARC task comparisons across visual-distance
quantiles.}
The reference task $t_i$ illustrates corner removal.
For each visual-distance quantile, task $t_j$ shares
the same action label as $t_i$ (teal), whereas task
$t_k$ has a different action label (terracotta).
Each task is illustrated by three support
input--output pairs.}
\label{fig:visual_distance_corner_removal}
\end{figure}

\begin{figure}[tpb]
\centering
\resizebox{0.975\linewidth}{!}{\input{figures/representation_analysis/visual_control/visual_distance_example_mark_center}}
\caption{\textbf{ARC task comparisons across visual-distance
quantiles.}
The reference task $t_i$ illustrates mark center.
For each visual-distance quantile, task $t_j$ shares
the same action label as $t_i$ (teal), whereas task
$t_k$ has a different action label (terracotta).
Each task is illustrated by three support
input--output pairs.}
\label{fig:visual_distance_mark_center}
\end{figure}

\begin{figure}[tpb]
\centering
\resizebox{0.975\textwidth}{!}{%
    \input{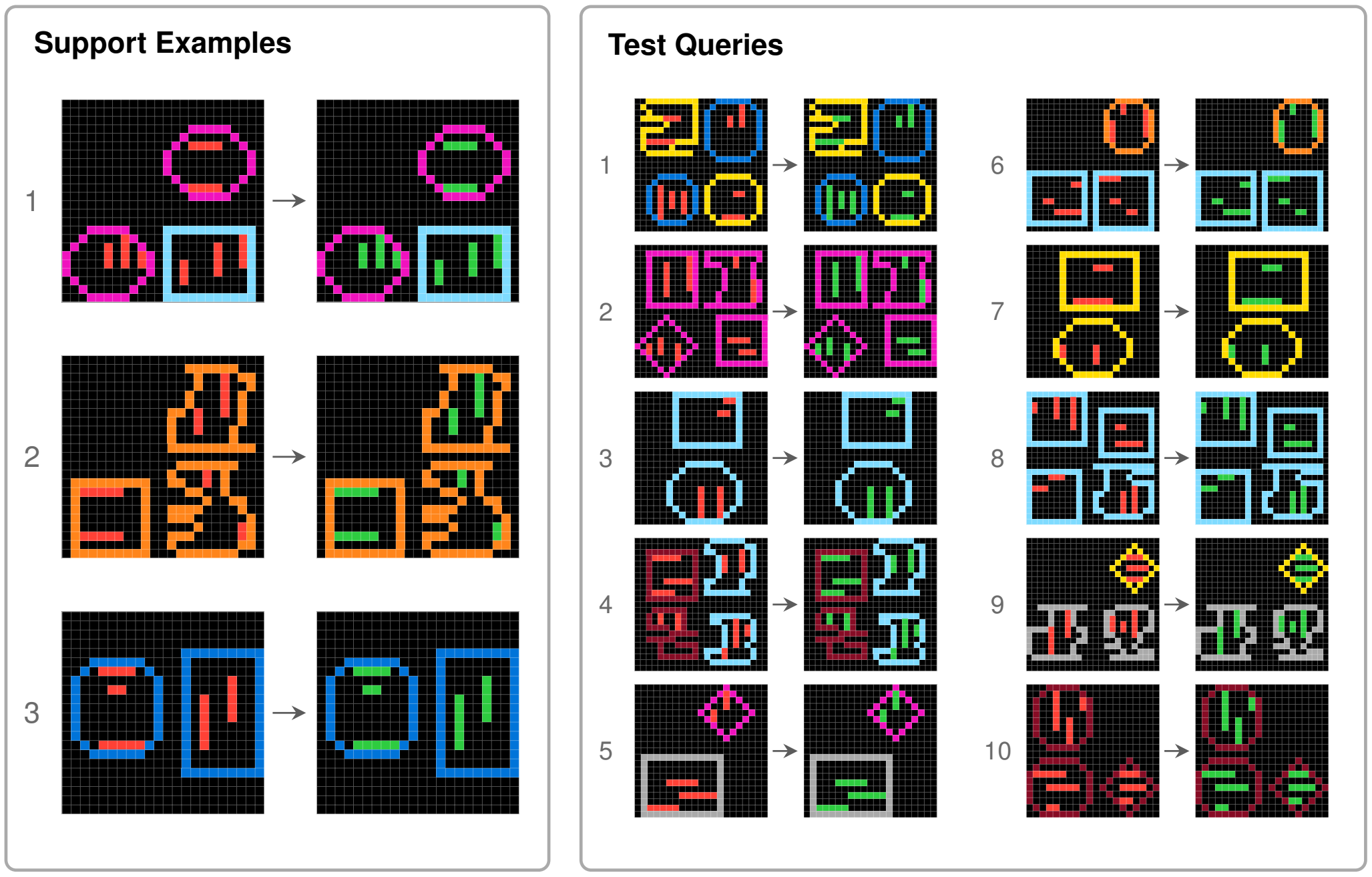}
}
\caption{\textbf{An example task from \textsc{ConceptARC-100} illustrating the ``\emph{Inside and Outside}'' concept.}
Each support example consists of an input grid and its corresponding output grid, shown to its right, which together illustrate the task-specific transformation. The goal is to infer the underlying transformation rule (here, to recolor every line object enclosed by another object green) from the three support examples and apply it to the input grid of each of the ten test queries to predict the corresponding output grid.}
\label{fig:inside_outside_example}
\end{figure}

\newcommand{\conceptARCPrompt}{
\begin{tikzpicture}
\node[
    draw=black!35,
    rounded corners=3pt,
    fill=black!2,
    inner sep=10pt,
    text width=14cm,
    align=left
] {
\ttfamily\footnotesize

\textbf{System:}\\
"You are a helpful assistant that solves analogy making puzzles.\\
Only give the answer, no other words or text."\\[8pt]

\textbf{User:}\\
"Let's try to complete the pattern:"\\[8pt]

input 1: 
[$\ldots$] [$\ldots$] $\ldots$ [$\ldots$] [$\ldots$]\\

output 1:
[$\ldots$] [$\ldots$] $\ldots$ [$\ldots$] [$\ldots$]\\[6pt]

input 2:
[$\ldots$] [$\ldots$] $\ldots$ [$\ldots$] [$\ldots$]\\

output 2:
[$\ldots$] [$\ldots$] $\ldots$ [$\ldots$] [$\ldots$]\\[6pt]

input 3:
[$\ldots$] [$\ldots$] $\ldots$ [$\ldots$] [$\ldots$]\\

output 3:
};
\end{tikzpicture}
}
\begin{figure}[tbp]
\centering
\resizebox{\linewidth}{!}{\conceptARCPrompt}%
\caption{\textbf{Prompt template used for ConceptARC evaluation.}
The system prompt specifies the required response format, while the user prompt provides task-specific demonstrations and a query grid for each ARC instance. The grids shown here are an illustrative example; the number, dimensions, and cell values of the input/output grids are replaced with the corresponding data for each puzzle. Colors are encoded as distinct integers from 0--9.}
\label{fig:conceptarc_prompt}
\end{figure}

\begin{figure}[h!]
\centering
\resizebox{\textwidth}{!}{%
    \input{figures/dataset/AboveBelow}
}
\caption{\textbf{Examples from 48 different generated tasks for the ``\emph{Above and Below}'' concept.} Each example consists of an input grid and its corresponding output grid to the right, illustrating the task-specific transformation. To illustrate task diversity, we show only \emph{one} input--output example per task. For more examples, we invite the reader to use our \href{https://mainlp.github.io/conceptarc-100-project-page/data_viewer.html}{data viewer}.}
\label{fig:above_below_examples}
\end{figure}

\begin{figure}[h!]
\centering
\resizebox{\textwidth}{!}{%
    \input{figures/dataset/TopBottom3D}
}
\caption{\textbf{Examples from 48 different generated tasks for the ``\emph{Top / Bottom (3D)}'' concept.} Each example consists of an input grid and its corresponding output grid to the right, illustrating the task-specific transformation. To illustrate task diversity, we show only \emph{one} input--output example per task. For more examples, we invite the reader to use our \href{https://mainlp.github.io/conceptarc-100-project-page/data_viewer.html}{data viewer}.}
\label{fig:top_bottom_3d_examples}
\end{figure}

\begin{figure}[h!]
\centering
\resizebox{\textwidth}{!}{%
    \input{figures/dataset/InsideOutside}
}
\caption{\textbf{Examples from 48 different generated tasks for the ``\emph{Inside and Outside}'' concept.} Each example consists of an input grid and its corresponding output grid to the right, illustrating the task-specific transformation. To illustrate task diversity, we show only \emph{one} input--output example per task. For more examples, we invite the reader to use our \href{https://mainlp.github.io/conceptarc-100-project-page/data_viewer.html}{data viewer}.}
\label{fig:inside_outside_examples}
\end{figure}

\begin{figure}[h!]
\centering
\resizebox{0.965\textwidth}{!}{%
    \input{figures/opener/extendtoboundary_overview}
}
\caption{\textbf{Examples from 42 different generated tasks for the ``\emph{Extend To Boundary}'' concept}. Each example consists of an input grid and its corresponding output grid to the right, illustrating the task-specific transformation rule. To illustrate task diversity, we show only \emph{one} input--output example per task. For more examples, we invite the reader to use our \href{https://mainlp.github.io/conceptarc-100-project-page/data_viewer.html}{data viewer}.}
\label{fig:extend_to_boundary_examples}
\end{figure}

\end{document}